\documentclass{article} 
\usepackage{iclr2027_conference,times}

\usepackage{graphicx}
\usepackage{booktabs}
\usepackage{tabularx}
\usepackage{hyperref}
\usepackage{url}
\usepackage{listings}
\lstdefinestyle{appendixjson}{
  basicstyle=\small\rmfamily,
  columns=fullflexible,
  keepspaces=true,
  breaklines=true,
  showstringspaces=false,
  frame=single,
  framerule=0.4pt,
  framesep=6pt,
  aboveskip=0.8em,
  belowskip=0.8em
}

\title{From 1D Benchmarks to Complex 3D Geometries:
Can Agents Reliably Specify Problems for Automated PDE Simulation?}

\author{%
  \parbox[t]{\dimexpr\textwidth-2\tabcolsep\relax}{%
    \centering
    Akshat Jani\textsuperscript{\ensuremath{\dagger}} \qquad
    Sakhinana Sagar Srinivas\textsuperscript{\ensuremath{\dagger}}\thanks{%
      Correspondence to: Sakhinana Sagar Srinivas
      (\href{mailto:sagar.sakhinana@tcs.com}{sagar.sakhinana@tcs.com}).
    }
    \qquad Venkataramana Runkana\textsuperscript{\ensuremath{\dagger}}
    \\[0.4em]
    \normalfont
    \textsuperscript{\ensuremath{\dagger}}Tata Research Development
    and Design Center, Pune, India 411057
  }%
}

\iclrfinalcopy

\newcommand{\contentspace}[1][0.7cm]{\par\vspace{#1}}

\begin{document}
\maketitle
\lhead{}


\begin{abstract}
Turning an engineering request into a simulation requires consistent
geometry, physics, boundary conditions, and numerical settings.
Recent agent-assisted systems automate simulation setup, but successful
execution alone does not establish whether the specification preserves
the intended problem or correctly handles missing information.
We study whether agents can reliably construct complete executable JSON
specifications for automated geometry creation, meshing, PDE solving,
and result export. Our workflow combines model extraction with
deterministic scientific and capability checks, explicit clarification,
and validated local patches that preserve unrelated accepted fields.
We evaluate specification correctness, execution completion, and numerical
verification separately. On 120 template-related requests, the full
workflow achieves 98.3\% correct first-turn specification and decision
outcomes, compared with 92.5\% for schema-assisted single shot and
82.5\% for model-only: gains of 5.8 and 15.8 percentage points,
respectively. It makes no false-ready decisions among the 42 requests
requiring clarification or rejection. Component ablations identify
contributions from scientific checks and conditional repair.
Separate downstream studies cover 1D, 2D, and 3D steady and unsteady,
linear and nonlinear problems, including flow and elasticity with
complex three-dimensional wing geometries. A 30-case gallery demonstrates
automated execution from retained JSON specifications, while 9 of 12
selected live user-to-results tests achieve joint specification,
execution, export, and numerical-reference success (75.0\%).
These results support traceable, engineer-reviewed simulation assistance
within the tested scope, while fresh-response regressions identify
consistency as a priority for further improvement.
\end{abstract}


\section{Introduction}
\label{sec:introduction}

An engineer designing a battery cooling system, a water pump, or a drone
wing faces a common task: turn a physical question into a simulation that
can support a design decision. This requires consistent geometry,
governing equations, material properties, boundary conditions, mesh, and
solver settings. Boundary conditions must refer to the correct surfaces,
and numerical settings must match the equation and time regime.
Three-dimensional geometries, nonlinear equations, and time-dependent
conditions make this coordination more demanding.

Language-model agents can assist by interpreting requests, collecting
missing information, preparing configurations, and coordinating simulation
tools. ReAct demonstrates how reasoning can be combined with actions and
external feedback \citep{yao2023react}. The aim is to accelerate routine
setup and iteration so that engineers can focus on unresolved physical
questions, analysis, and design decisions. Engineers retain responsibility
for choosing appropriate models and assessing the results. Useful
assistance therefore requires preserving the intended physics and keeping
unresolved information visible.

Recent systems demonstrate progress in simulation automation.
MetaOpenFOAM uses collaborating agents and retrieved examples to configure
and execute OpenFOAM cases \citep{chen2024metaopenfoam}, while
Foam-Agent 2.0 covers meshing, execution, and visualization
\citep{yue2025foamagent2}. ChatCFD combines domain knowledge with error
diagnosis and evaluates physical fidelity beyond successful execution
\citep{fan2026chatcfd}. For finite-element simulation, MechAgents explores
collaborating agents for elasticity problems \citep{ni2024mechagents},
and ALL-FEM combines specialized agents with fine-tuned models for
FEniCS code generation \citep{deotale2026allfem}. JSONSchemaBench examines
structured-output generation through schema coverage, efficiency, and
output quality \citep{geng2025jsonschemabench}. Together, these studies
motivate examining whether an accepted specification represents the
request, contains sufficient information, and remains consistent after
corrections.

PDEFlow directly motivates this study \citep{jani2026pdeflow}.
It connects conversational specification with finite-element data
generation, neural-operator training, and checkpoint-based inference.
Its stateful workflow supports local updates, with a formulation covering
linear one- and two-dimensional problems. An incorrect specification can
affect subsequent simulations, training data, and predictions.
Here, we examine specification reliability while extending the supported
simulation scope to nonlinear equations and complex three-dimensional
geometries.

We therefore ask:
\begin{quote}
Can agents reliably construct complete JSON specifications that preserve
the intended physics, identify missing or unsupported information, and
trigger automated geometry creation, meshing, solving, and result export?
\end{quote}

Our study builds on these advances by examining the reliability of the
scientific specification and the decision to permit execution.
We address three connected research gaps.

First, successful execution alone cannot establish whether a system
correctly handles incomplete, conflicting, or unsupported requests.
We test whether it should produce a complete specification, ask for
clarification, or reject the request. Scientific fields are checked
against fixed references, and incorrect decisions to proceed are measured
separately. Clarification completion is evaluated separately from the
initial decision to ask a question.

Second, repairing a configuration must preserve the intended problem
beyond the corrected field. We use a retained JSON specification as the
interface between the agent and reviewed numerical routines. The model
proposes content; deterministic code checks its structure, request
evidence, and supported capabilities. Clarifications and repairs use
local patches restricted to authorized paths, with revision checks and
validation before acceptance. This makes changes traceable and protects
unrelated accepted fields. Component ablations examine how these checks
and repair mechanisms affect correctness.

Third, specification correctness, execution completion, and numerical
accuracy require different evidence. We evaluate user-to-JSON outcomes,
numerical verification, and execution from retained specifications
separately. Reference comparisons and convergence studies assess numerical
accuracy; the gallery assesses downstream execution; selected live tests
assess the complete user-to-results path. These studies retain distinct
denominators and cover supported 1D/2D/3D, steady/unsteady, and
linear/nonlinear benchmarks, including flow and elasticity with
three-dimensional wing geometries.

This design provides a practical basis for production integration.
An explicit JSON interface connects agents to engineering software while
keeping assumptions and accepted changes available for review.
Reviewed solver templates, controlled updates, and retained records
support repeatable operation. Commercial deployment would additionally
require application-specific validation, access and computing-budget
controls, and engineer approval for consequential decisions.
Our study evaluates the underlying reliability mechanisms and identifies
where further development is needed.


\section{Methodology}
\label{sec:methodology}

\begin{figure}[htbp]
  \centering
  \includegraphics[width=\linewidth]{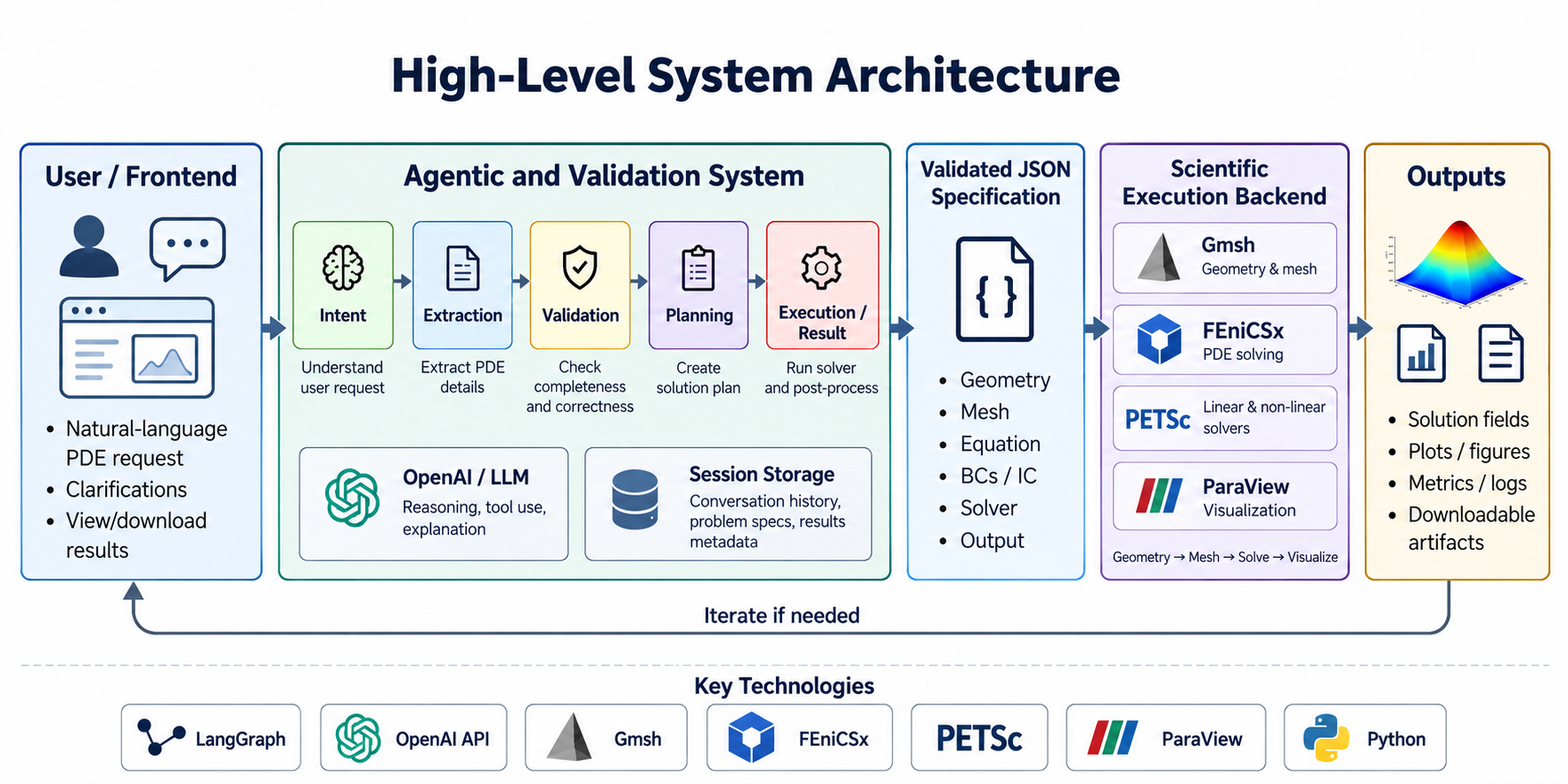}
  \caption{Overall workflow, from left to right.
  (a) User/frontend: receives PDE requests and clarification replies.
  (b) Agentic and validation system: constructs and checks the specification
  while retaining session history.
  (c) Validated JSON: records the problem and numerical settings.
  (d) Execution backend: Gmsh creates geometry and meshes, FEniCSx and
  PETSc solve the PDEs, and ParaView supports visualization.
  (e) Outputs: returns fields, plots, diagnostics, and downloadable
  artifacts. The feedback loop supports validated user revisions.}
  \label{fig:overall-system}
\end{figure}

The framework separates specification construction from numerical execution
(Figure~\ref{fig:overall-system}). A language model interprets the request
and proposes a JSON specification. Deterministic code checks the
specification, manages authorized changes, and controls whether it can
enter the numerical workflow.

\subsection{The executable JSON specification}
\label{sec:json-specification}

The JSON specification describes the physical problem and the numerical
settings required to run it (Table~\ref{tab:json-contract}). Fields follow
explicit types: a scalar value is represented as a constant or an
expression, and a vector contains scalar components. Boundary conditions
identify both their target boundaries and the required values. This
structure preserves spatially varying expressions rather than reducing
them to constants.

\begin{table}[htbp]
  \caption{Main fields of the executable JSON specification. The original
  request, schema version, and descriptive labels are also retained.}
  \label{tab:json-contract}
  \centering
  \small
  \begin{tabularx}{\linewidth}{@{}p{0.40\linewidth}X@{}}
    \toprule
    \textbf{JSON fields} & \textbf{Contents} \\
    \midrule
    geometry, mesh
      & Domain parameters, mesh backend, and resolution. \\
    equation
      & Equation family, coefficients, sources, and nonlinear law. \\
    boundary\_conditions, initial\_condition
      & Boundary selectors, condition types and values, and initial fields. \\
    finite\_element, time
      & Element family and degree; steady/transient regime, time step,
        final time, and integration scheme. \\
    solver, output
      & Linear/nonlinear solver settings, tolerances, and requested exports. \\
    \bottomrule
  \end{tabularx}
\end{table}

The initial draft also records ambiguities and the source of extracted
values. Provenance records link values to quoted request text and its
character positions, or identify a permitted default policy. Required
information that is missing remains unresolved. Defaults are
distinguishable from user-supplied facts, and a partial draft becomes
executable only after it satisfies the complete problem contract.

\subsection{Constructing the specification}

The initial model call receives the user request, extraction instructions,
and a strict JSON schema. A vocabulary-based router narrows the geometry
and equation choices when it recognizes the request; otherwise, the full
set of relevant choices remains available. Reviewed case aliases, when
present, help identify supported problem families without supplying
unrequested physical values or benchmark answers. The same model is used
for extraction and any subsequent model repair.

\begin{figure}[htbp]
  \centering
  \includegraphics[width=\linewidth]{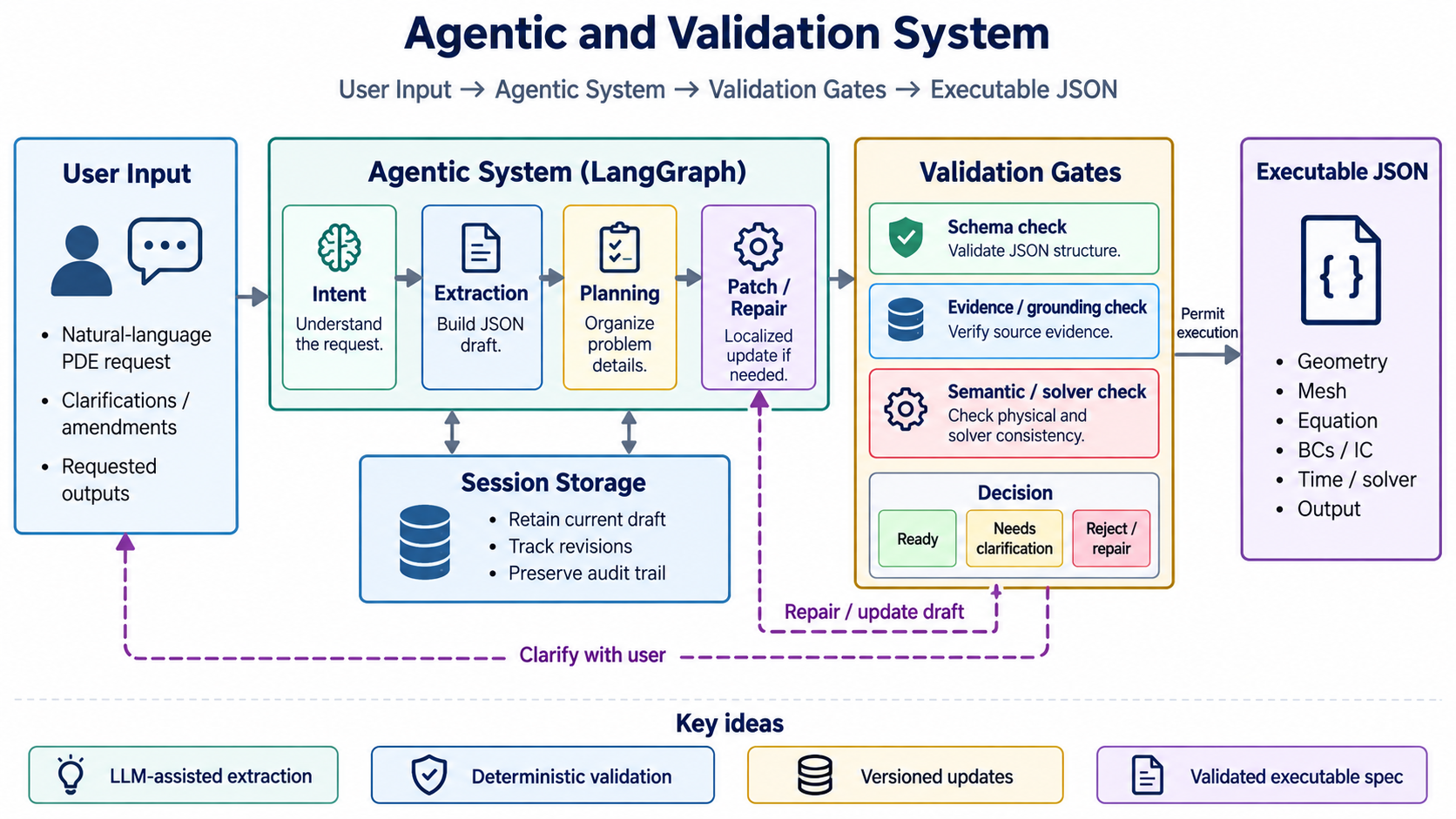}
  \caption{Specification construction and validation. The model proposes
  a draft or local repair, while deterministic checks assess structure,
  request evidence, and solver compatibility. Missing information returns
  to the user for clarification. Session storage retains the specification,
  revisions, and audit records. Execution requires a complete specification
  with no blocking findings.}
  \label{fig:agentic-system}
\end{figure}

The response is parsed into a retained draft. Deterministic normalization
handles supported equivalent representations, such as alternative enum
spellings and evidence-path notation. Evidence positions are checked
against the unchanged request, and raw responses and normalization records
are retained for inspection. Schema-constrained generation controls the
response structure; the checks in Figure~\ref{fig:agentic-system} assess
its scientific consistency and suitability for execution.

\subsection{Validation, clarification, and patch repair}

Validation checks field types, relationships between sections, agreement
with request evidence, and compatibility with supported numerical
capabilities. These checks cover dimensions, mesh and boundary conventions,
scalar/vector fields, time settings, and solver requirements. The decision
is \textit{ready}, \textit{needs\_clarification}, or \textit{rejected}.
Readiness requires a complete specification with no blocking findings.
Missing information produces a clarification question, ordered by field
dependencies and a fixed priority. Unsupported requests and explicit
contradictions are rejected. These checks have finite coverage and do not
prove scientific correctness.

Clarifications, amendments, and repairs update the retained specification
through versioned add, replace, or remove operations tied to a source turn
and base revision. Application code determines the permitted paths;
changing an accepted value requires user authorization. A model repair
receives the findings, permitted paths, and nearby context, with at most
one attempt per accepted user turn. Fully determined boundary corrections
can instead be constructed directly from independently matched request
data. Each proposal is applied to a private copy and checked for revision,
scope, and validity before acceptance. Invalid or stale proposals leave
the accepted specification unchanged. The original request, prior evidence,
and patch audits are retained. Execution permission is recomputed after
updates and session reloads. Appendix~\ref{app:json-protocol} gives the
detailed contract.

\subsection{Geometry, mesh generation, and PDE solving}

The executor maps the validated specification to a registered solver
template. Reviewed code defines the geometry builders and finite-element
weak forms; the model does not generate executable solver code. Structured
builders or Gmsh create the mesh, and boundary selectors identify the
corresponding mesh facets. Complex geometries distinguish structural solids
from external-flow domains and retain their physical boundary groups.
Mathematical expressions are evaluated through a restricted parser.
FEniCSx/UFL constructs and assembles the finite-element forms, while PETSc
solves the linear or nonlinear systems.

Scalar templates cover supported 1D/2D/3D problems; linear elasticity and
steady incompressible flow cover supported 2D/3D problems. Transient heat
and nonlinear reaction-diffusion use backward Euler or Crank--Nicolson
time integration. Geometry and equation combinations are checked
explicitly rather than assumed to be interchangeable. Each result bundle
retains the accepted specification, mesh and field exports, solver
diagnostics, and execution status. Reference comparisons and convergence
studies assess numerical accuracy separately
(Appendix~\ref{app:numerical}). Specification acceptance, execution
completion, and numerical verification therefore remain distinct outcomes.


\section{Experiments and Results}
\label{sec:experiments}

\subsection{Experimental setup}
\label{sec:setup}

We evaluate three stages separately: constructing JSON from user requests,
verifying numerical solutions, and executing retained JSON specifications.
Tables~\ref{tab:agentic-setup}--\ref{tab:gallery-setup} summarize the studies.
Paper IDs identify experiments and gallery groups; their original records
are mapped in Appendix~\ref{app:reproducibility}.

\paragraph{Agentic studies: user requests to JSON.}
We use gpt-5.4-mini-2026-03-17 with reasoning effort set to
\textit{none}, a 6,400-token output limit, and workflow v1.8 with at most
one model patch repair. The main benchmark contains 120 requests:
48 extraction, 30 complete-specification, 30 clarification, and
12 rejection/conflict tasks. Half use natural wording and half use
schema-explicit wording. The requests are code-authored and share related
templates. Scalar problems cover supported 1D/2D/3D, steady/unsteady,
and linear/nonlinear settings.

\begin{table}[htbp]
  \caption{Agentic studies: user requests to JSON. Studies reuse requests,
  so their counts are not added together.}
  \label{tab:agentic-setup}
  \centering
  \small
  \begin{tabularx}{\linewidth}{@{}l p{0.23\linewidth} X r@{}}
    \toprule
    \textbf{ID} & \textbf{Experiment}
      & \textbf{Evaluation} & \textbf{Requests} \\
    \midrule
    A1 & Specification comparison
       & Compare model-only, schema-assisted single shot,
         and the full workflow
       & 120 \\
    A2 & Component ablation
       & Remove one component at a time; input-changing controls
         use a fixed 40-request subset
       & 120 \\
    A3 & Clarification completion
       & Apply fixed authorized replies and check whether
         a correct complete specification is reached
       & 30 \\
    \bottomrule
  \end{tabularx}
\end{table}

\paragraph{Numerical verification.}
We compare solutions with analytical or manufactured references using
integrated finite-element errors and predefined acceptance rules.
Scalar and elasticity studies use P1 elements; incompressible flow uses
P2/P1 elements. Structured builders or Gmsh generate the meshes, and
FEniCSx/PETSc performs the solves. The latest verification campaign uses
the DOLFINx 0.11.0 series, Gmsh 4.12.1, and PETSc 3.19.6.
Exact build versions and hardware are recorded in
Appendix~\ref{app:reproducibility}. BE and CN denote backward Euler
and Crank--Nicolson.

V6 exceeds the spatial-isolation limit: $10.84\%$ compared with the
allowed $10\%$. The optional finest 3D cantilever study was not attempted.
Complete inputs, refinement sequences, solver settings, and numerical
results are provided in Appendix~\ref{app:numerical}.

\begin{table}[h]
  \caption{Numerical verification studies. Counts include 56 completed
  study solves, including studies with failed checks. Development smoke
  runs are excluded.}
  \label{tab:numerical-setup}
  \centering
  \small
  \begin{tabularx}{\linewidth}{@{}l p{0.29\linewidth} X r@{}}
    \toprule
    \textbf{ID} & \textbf{Experiment}
      & \textbf{Verification setup} & \textbf{Solves} \\
    \midrule
    V1 & 2D elasticity
       & Four mesh levels; displacement and energy checks
       & 4 \\
    V3 & 2D steady flow
       & Three mesh levels and one pressure patch test
       & 4 \\
    V4 & 2D unsteady heat
       & Five step sizes for each BE/CN scheme;
         one equilibrium control
       & 11 \\
    V6 & 2D annular heat
       & BE/CN refinement and spatial-isolation check
       & 8 \\
    V7 & 2D rectangular heat
       & BE/CN refinement and spatial-isolation check
       & 8 \\
    \midrule
    V2 & 3D elasticity
       & Four mesh levels; displacement and energy checks
       & 4 \\
    V5 & 3D nonlinear reaction-diffusion
       & Five step sizes for each BE/CN scheme;
         one equilibrium control
       & 11 \\
    V8 & 3D parallel agreement
       & Same cantilever problem with 4, 8, 16,
         and 32 MPI processes
       & 4 \\
    \midrule
    V9 & Solver tolerance checks
       & Tighter tolerances for 2D flow
         and 3D reaction-diffusion
       & 2 \\
    \midrule
    \multicolumn{3}{r}{\textbf{Total}} & \textbf{56} \\
    \bottomrule
  \end{tabularx}
\end{table}

\paragraph{Solution gallery: JSON to results.}
We execute 30 retained JSON specifications through geometry creation,
meshing, solving, and export: three cases in 1D, sixteen in 2D, and
eleven in 3D. The gallery includes steady and unsteady, linear and
nonlinear problems, with flow and elasticity cases involving wing
geometries. Table~\ref{tab:gallery-setup} groups the cases by physics
and dimension. Completion establishes execution; numerical qualifications
are reported separately in Appendix~\ref{app:gallery}.

Gallery inputs are retained JSON specifications, so these cases do not
measure user-to-JSON accuracy. Separate live user-to-results tests
evaluate the complete path (Appendix~\ref{app:e2e}).

\begin{table}[h]
  \caption{Solution gallery: JSON to results. Each case belongs
  to one group.}
  \label{tab:gallery-setup}
  \centering
  \small
  \begin{tabularx}{\linewidth}{@{}l X p{0.30\linewidth} r@{}}
    \toprule
    \textbf{ID} & \textbf{Experiment group}
      & \textbf{Scope} & \textbf{Cases} \\
    \midrule
    G1 & Rod and polyline diffusion
       & 1D; steady/unsteady; linear
       & 3 \\
    \midrule
    G2 & Linear scalar equations
       & 2D; steady; linear
       & 6 \\
    G3 & Nonlinear diffusion
       & 2D; steady; nonlinear
       & 2 \\
    G4 & Heat and reaction-diffusion
       & 2D; unsteady; linear/nonlinear
       & 4 \\
    G7 & Airfoil and cylinder flow
       & 2D; steady; nonlinear
       & 2 \\
    G9 & Cantilever and perforated plate
       & 2D; steady; linear elasticity
       & 2 \\
    \midrule
    G5 & Diffusion and conduction
       & 3D; steady; linear/nonlinear
       & 5 \\
    G6 & Heat and reaction-diffusion
       & 3D; unsteady; linear/nonlinear
       & 2 \\
    G8 & Wing and sphere flow
       & 3D; steady; nonlinear
       & 2 \\
    G10 & Wing and solid cantilevers
        & 3D; steady; linear elasticity
        & 2 \\
    \midrule
    \multicolumn{3}{r}{\textbf{Total}} & \textbf{30} \\
    \bottomrule
  \end{tabularx}
\end{table}

\subsection{Tasks, baselines, and evaluation measures}
\label{sec:evaluation}

\paragraph{Agentic studies (A1--A3).}
A1 measures whether the first response contains the expected scientific
fields and correctly decides to proceed, request clarification, or reject.
All systems use the same fixed references and scorer. The 120 requests
include 78 complete requests and 42 requiring clarification or rejection.
We report first-turn correctness over all 120 requests, complete-specification
correctness over the 78 complete requests, and false-ready decisions over
the 42 non-ready requests. A false-ready decision occurs when the system
permits execution although clarification or rejection is required.
Valid JSON alone does not establish scientific correctness.

We compare three systems. Model-only produces JSON in one response.
Schema-assisted single shot adds schema handling and basic deterministic
checks. The full workflow adds request grounding, scientific and capability
checks, a readiness decision, and conditional patch repair.
Single shot and the full workflow share the initial model response.
Model-only uses a different prompt and an independent response, so its
comparison also reflects those differences.

A2 compares each component removal with its own full-workflow reference.
Deterministic removals reuse saved responses for 120 requests.
Input-changing controls use new model responses on a fixed 40-request
subset. We distinguish the number evaluated from the number that actually
used the removed component. A3 applies fixed authorized replies to
30 clarification requests and checks whether a correct complete
specification is reached. This measures completed interaction separately
from first-turn clarification decisions. API costs and model-call times
are recorded separately from solver time; shared calls are counted once.

\paragraph{Numerical verification (V1--V9).}
V1--V3 test spatial convergence against reference fields, with pressure
and energy checks where applicable. V4--V7 test temporal convergence
under BE and CN; V4--V5 include equilibrium controls, and V6--V7 check
whether spatial error is sufficiently small to assess temporal accuracy.
We report integrated $L^2$ errors, $H^1$ seminorm errors where applicable,
and observed convergence orders. The spatial-contamination limit is
$10\%$. V8 compares parallel solutions with a retained serial solution,
using a relative coefficient and quantity-of-interest limit of $10^{-7}$.
V9 compares baseline and tighter-tolerance solutions using a relative
field and quantity-of-interest limit of $10^{-6}$. Passing verification
requires every declared check; completed solves and unattempted studies
are not automatically counted as passes
(Appendix~\ref{app:numerical}).

\paragraph{Solution gallery and live workflow tests.}
For G1--G10, execution succeeds when geometry creation, meshing, solving,
and the required exports complete. Numerical accuracy is assessed
separately using available references and refinement checks; verified,
failed, and preliminary qualifications are retained.
For the 12 selected live user-to-results attempts, joint success requires
the intended specification, a completed solve, the required artifacts,
and a passed numerical reference check. Blocked attempts, wrong-intent
executions, and reference failures remain in the denominator.
Detailed outcomes appear in Appendices~\ref{app:gallery}
and~\ref{app:e2e}.

\subsection{Specification construction results}
\label{sec:a1-results}

Table~\ref{tab:a1-results} reports the v1.8 comparison on all 120 requests in A1.
The full workflow achieves $118/120$ correct first-turn outcomes
($98.3\%$), compared with $111/120$ ($92.5\%$) for schema-assisted
single shot and $99/120$ ($82.5\%$) for model-only.
It produces correct complete specifications for $76/78$ complete
requests ($97.4\%$) and makes no false-ready decisions on the
42 requests requiring clarification or rejection.
In A3, fixed authorized replies lead to correct complete specifications
for $28/30$ clarification requests ($93.3\%$).
These results describe v1.8 on the main template-related benchmark.
Numerical accuracy is assessed separately. Appendix~\ref{app:additional-agentic}
provides detailed task results, clarification studies, and later v1.9
checks. The latter distinguish re-evaluation of saved contexts from
construction using fresh model responses. Replay improvements do not
establish a uniform gain in fresh construction or replace the A1 scores.

\begin{table}[h]
  \caption{Specification and decision correctness for A1 (v1.8).
  Denominators are 120 total requests, 78 complete requests,
  and 42 requests requiring clarification or rejection.
  Failed cases remain in the denominators.}
  \label{tab:a1-results}
  \centering
  \small
  \begin{tabularx}{\linewidth}{@{}Xccc@{}}
    \toprule
    \textbf{System}
      & \shortstack{Correct first-turn\\outcomes}
      & \shortstack{Correct complete\\specifications}
      & \shortstack{False-ready\\decisions} \\
    \midrule
    Model-only
      & 99/120 (82.5\%)
      & 70/78 (89.7\%)
      & 6/42 \\
    Schema-assisted single shot
      & 111/120 (92.5\%)
      & 75/78 (96.2\%)
      & 1/42 \\
    Full workflow
      & 118/120 (98.3\%)
      & 76/78 (97.4\%)
      & 0/42 \\
    \bottomrule
  \end{tabularx}
\end{table}

\subsection{Component ablation results}
\label{sec:ablations}

\begin{figure}[htbp]
  \centering
  \includegraphics[width=\linewidth]{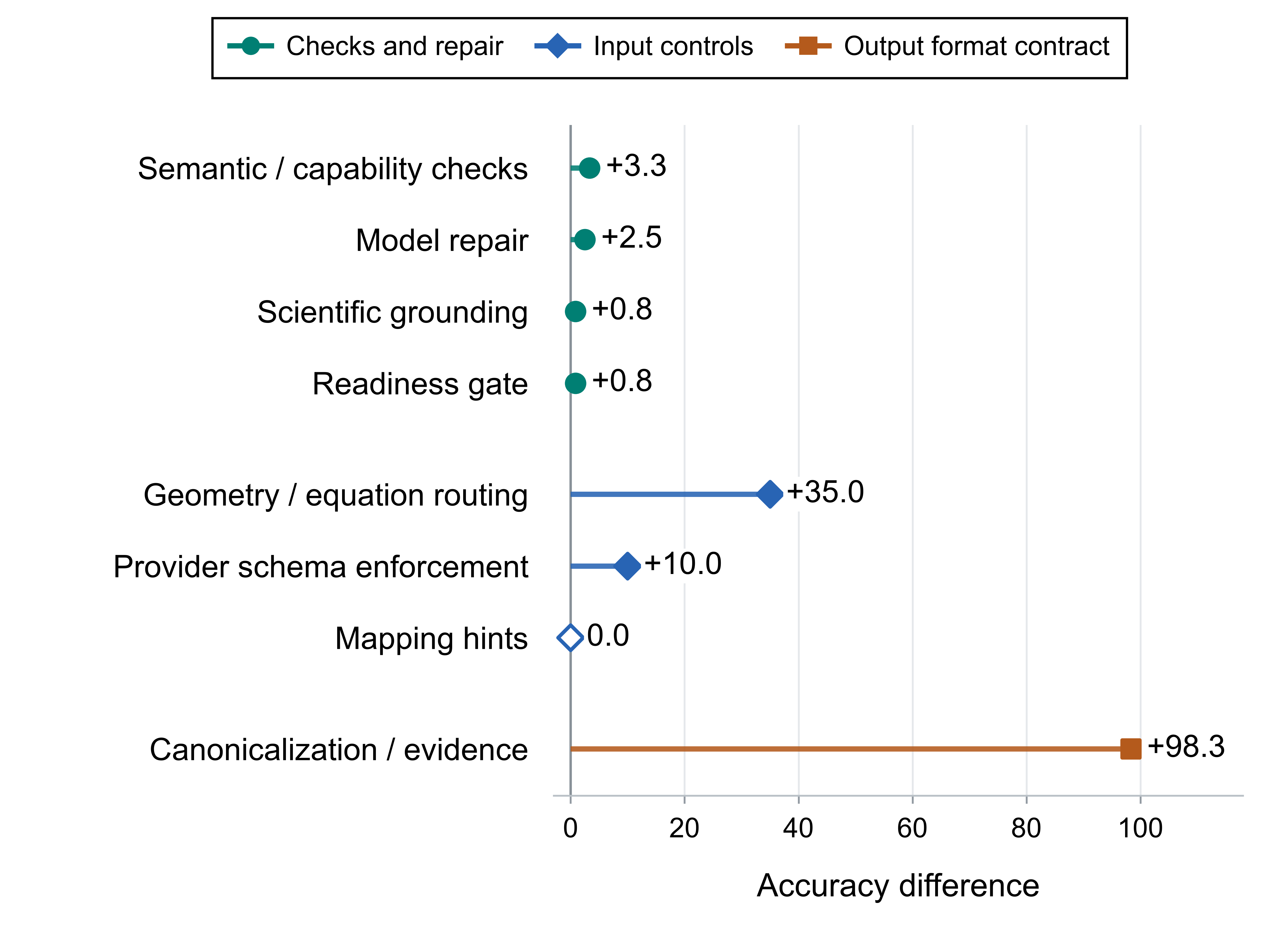}
  \caption{Component ablation results. Values are full-workflow accuracy
  minus ablated accuracy, in percentage points, on matched requests.
  Positive values favour the full workflow.
  The 120-request comparisons use a reference score of $118/120$;
  the 40-request input controls use $40/40$.
  All groups share one horizontal scale.
  Output-format removal causes parsing failures on all 120 requests.
  The hollow marker identifies unexercised mapping hints.
  Schema and routing controls use new model draws, so their differences
  also include variation between responses.}
  \label{fig:component-ablations}
\end{figure}

A2 removes one component at a time from v1.8 and compares each variant
with its full-workflow reference on the same requests
(Figure~\ref{fig:component-ablations} and
Table~\ref{tab:component-ablations}).
Deterministic removals reuse saved model responses, and the model-repair
comparison uses recorded outcomes before repair.
Input-changing controls use fresh model responses on a fixed
40-request subset. Component use is reported separately because an
evaluated request may never reach the component being removed.

Removing semantic/capability reconciliation reduces correct outcomes
by four, and removing model repair reduces them by three.
Removing scientific grounding or the readiness gate reduces them
by one each; mandatory structural checks remain active when grounding
is removed. The output-format result measures dependence on the
required representation. Mapping hints are not exercised in the
40-request input-control subset, so their zero difference does not show
that they are unnecessary. Component effects overlap and must not be added together.

\begin{table}[htbp]
  \caption{Component ablation results for A2 (v1.8).
  Entries show counts and percentages.
  Component exercised counts requests that used the removed component
  in the reference workflow. Each block states its full-workflow
  reference; all evaluated requests remain in the denominators.}
  \label{tab:component-ablations}
  \centering
  \small
  \begin{tabularx}{\linewidth}{@{}Xcc@{}}
    \toprule
    \textbf{Condition}
      & \shortstack{Correct /\\evaluated}
      & \shortstack{Component\\exercised} \\
    \midrule
    \multicolumn{3}{@{}l}{
      \emph{120-request comparisons; reference: 118/120 (98.3\%)}
    } \\
    Full workflow
      & 118/120 (98.3\%)
      & --- \\
    Without model repair
      & 115/120 (95.8\%)
      & 5/120 (4.2\%) \\
    Without scientific grounding
      & 117/120 (97.5\%)
      & 117/120 (97.5\%) \\
    Without semantic/capability reconciliation
      & 114/120 (95.0\%)
      & 12/120 (10.0\%) \\
    Without readiness gate
      & 117/120 (97.5\%)
      & 4/120 (3.3\%) \\
    Without canonicalization/evidence formatting
      & 0/120 (0.0\%)
      & 120/120 (100.0\%) \\
    \midrule
    \multicolumn{3}{@{}l}{
      \emph{40-request input controls; reference: 40/40 (100.0\%)}
    } \\
    Without provider schema enforcement
      & 36/40 (90.0\%)
      & 40/40 (100.0\%) \\
    Without geometry/equation routing
      & 26/40 (65.0\%)
      & 40/40 (100.0\%) \\
    Without mapping hints
      & 40/40 (100.0\%)
      & 0/40 (0.0\%) \\
    \bottomrule
  \end{tabularx}
\end{table}

\subsection{Failure cases and limits of reliability}
\label{sec:failure-limits}

Failures include omitted or altered expressions, incorrect boundary data,
and repairs that do not recover the intended specification. A correct
clarification question does not guarantee a correct completed specification,
and readiness checks can miss scientific errors. Later regressions recover
some retained errors but show mixed results when specifications are rebuilt
from fresh responses. Workflow changes and new responses occur together,
so the fresh-response tests do not isolate the cause of each difference. Related request
templates, a fixed model, and the tested versions limit conclusions about
unfamiliar engineering requests (Appendix~\ref{app:additional-agentic}).

Specification correctness, execution completion, and numerical accuracy
remain separate requirements. Gallery cases start from retained JSON
and therefore assess downstream execution rather than agent construction.
The selected live tests use simple reference fields, while some
numerical studies fail or remain unattempted. These qualifications
are retained in Appendices~\ref{app:numerical},
\ref{app:gallery}, and~\ref{app:e2e}.


\subsection{Conclusion}
\label{sec:conclusion}

We studied whether agents can construct executable JSON specifications
for automated geometry creation, meshing, PDE solving, and result export.
On the main benchmark, the full workflow achieves $118/120$ correct
first-turn outcomes ($98.3\%$), compared with $111/120$ ($92.5\%$)
for schema-assisted single shot and $99/120$ ($82.5\%$) for model-only.
It produces correct complete specifications for $76/78$ complete
requests ($97.4\%$). Ablations identify contributions from scientific
checks and conditional repair. Separately, the gallery demonstrates
execution from 30 retained JSON specifications across supported
steady/unsteady and linear/nonlinear problems, including complex
wing geometries. The v1.9 live workflow test achieves $9/12$ joint
successes ($75.0\%$) on selected known requests.

These results support simulation assistance within the tested scope,
with engineers reviewing assumptions and results. Broader-input failures
show that general reliability remains unresolved. Future work should
improve preservation of expressions and boundary conditions, strengthen
repair coverage, and evaluate independently authored requests with
numerical reference checks.

\clearpage

\bibliographystyle{iclr2027_conference}
\bibliography{references}

\contentspace[1.5cm]

\clearpage
\appendix


\section{JSON Specification, Prompts, and Patch Protocol}
\label{app:json-protocol}

The workflow uses three objects: an extraction draft, a complete
solver-facing JSON specification, and a patch that updates the retained
draft. The model proposes content; application code validates it and
determines whether execution is permitted.

\subsection{Complete schema and field requirements}
\label{app:complete-schema}

The extraction draft contains ``problem'', ``ambiguities'', and
``provenance''. Unresolved fields may be null. Parsing removes null
placeholders and applies documented normalization rules before validation.
A draft becomes executable only when it satisfies the complete problem
contract and the scientific and capability checks.
Table~\ref{tab:appendix-problem-fields} summarizes the solver-facing fields.
The problem schema uses version 2.0; workflow and patch versions are
recorded separately.

\begin{table}[htbp]
  \caption{Fields of the solver-facing JSON specification.
  Requirements depend on the selected geometry and equation.}
  \label{tab:appendix-problem-fields}
  \centering
  \small
  \begin{tabularx}{\linewidth}{@{}p{0.28\linewidth}X@{}}
    \toprule
    \textbf{Field} & \textbf{Contents and requirements} \\
    \midrule
    ``schema\_version''
      & Identifies the problem contract; version 2.0 is used. \\
    ``title''
      & Descriptive label; not scientific evidence. \\
    ``original\_query''
      & Original request, retained unchanged.
        Later user replies are stored separately. \\
    ``geometry''
      & Geometry type, dimension, and required parameters,
        such as extents, radii, or wing dimensions. \\
    ``mesh''
      & Mesh backend, resolution, and cell settings,
        consistent with the geometry and dimension. \\
    ``finite\_element''
      & Element family and degree.
        Lagrange degrees one and two are supported,
        subject to solver requirements. \\
    ``equation''
      & Equation type, coefficients, sources, material properties,
        and nonlinear parameters where required. \\
    ``boundary\_conditions''
      & Condition types, target boundaries, and prescribed data.
        Selectors and field dimensions must match the problem. \\
    ``time''
      & Steady or transient regime.
        Transient settings include a positive time step,
        valid time interval, integration scheme, and step limit. \\
    ``initial\_condition''
      & Initial scalar or vector field for a transient problem. \\
    ``solver''
      & Linear/nonlinear methods, preconditioners, tolerances,
        and iteration limits. \\
    ``output''
      & Requested exports and statistics.
        Supported formats include JSON, XDMF, VTKHDF, and PNG. \\
    \bottomrule
  \end{tabularx}
\end{table}

\paragraph{Supported representations.}
Geometry branches include intervals and polylines in 1D;
rectangles, disks, annuli, L-shaped and perforated domains, and airfoil
flow domains in 2D; and boxes, cylinders, spheres, solid wings,
and external-flow domains in 3D. Equation branches cover Poisson,
Laplace, steady heat, Helmholtz, advection--diffusion, nonlinear Poisson,
transient heat, nonlinear reaction--diffusion, linear elasticity,
and steady incompressible flow. Nonlinear branches use the declared
power, exponential, or hyperbolic-sine law. The capability checks
determine which combinations can run; the catalogues do not imply
support for every combination.

Scalars are represented as constants or expressions, and vectors contain
scalar components. Expressions pass through a restricted mathematical
parser. Boundary selectors use named boundaries for structured meshes
or registered physical groups for Gmsh meshes. Dirichlet conditions
require a value, Neumann conditions require a flux, and Robin conditions
require coefficient and reference fields. Changing a condition type
also requires updating its associated data.

\paragraph{Evidence and defaults.}
Each provenance record identifies the field, extracted value, quoted
request fragment, character positions, and origin. Character positions
start at zero and exclude the end position. The fragment must match
the original request exactly, and its field path must resolve to the
claimed value. List-valued facts are recorded as whole lists.
Generated titles and schema labels are not scientific evidence.

Documented defaults must identify an applicable policy and remain
distinguishable from user-supplied values. Missing critical information,
such as required geometry dimensions or transient settings, remains
unresolved and produces clarification requirements.

\subsection{Prompts and generation settings}
\label{app:prompts-settings}

Assisted extraction combines representation instructions, the extraction
prompt, the field vocabulary, and a request-specific JSON schema.
A vocabulary-based router narrows recognized geometry and equation
choices; otherwise, the relevant catalogue remains available.
Optional reviewed case identifiers describe supported capabilities
without supplying unrequested physical values. The original request
remains the source of scientific evidence.

\begin{table}[htbp]
  \caption{Generation settings for the principal agentic studies.
  Requests and responses are retained locally.}
  \label{tab:appendix-generation-settings}
  \centering
  \small
  \begin{tabularx}{\linewidth}{@{}p{0.30\linewidth}X@{}}
    \toprule
    \textbf{Item} & \textbf{Setting} \\
    \midrule
    Model snapshot
      & gpt-5.4-mini-2026-03-17 \\
    Reasoning effort
      & ``none'' \\
    Maximum output
      & 6,400 tokens per model call \\
    Provider storage
      & Disabled \\
    Assisted extraction
      & Strict JSON-schema output;
        the schema is also supplied in the instructions \\
    Model-only baseline
      & JSON-object output with the generic problem schema
        supplied as text; separate prompt and model response \\
    Model repair
      & At most one attempt per accepted user turn,
        using a strict local-patch schema \\
    Clarification study
      & Fixed authorized replies in A3;
        evaluated separately from first-turn decisions \\
    Versions
      & Workflow 1.8;
        patch contract ``phase2-context-patch-1.0'' \\
    \bottomrule
  \end{tabularx}
\end{table}

The following quoted blocks summarize the instructions.
Exact prompt files and schemas are retained in the source materials.

\paragraph{Extraction instructions.}
\begin{quote}
``Return one extraction draft using the supplied request and schema.
Preserve stated values, expressions, condition types, and boundary targets.
Leave missing information unresolved and record scientific values with
exact request evidence. Report ambiguity and contradictions without
inventing an interpretation. Do not replace unsupported physics or
geometry with an approximation. Do not solve the PDE.''
\end{quote}

\paragraph{Repair instructions.}
The repair request supplies the original request, authorized user turns,
detected findings, permitted paths, current values, and nearby context.

\begin{quote}
``Return local operations against the supplied revision.
Change only permitted paths using the available evidence.
Preserve unrelated fields and correct boundary selectors.
Do not rewrite the original request or evidence.
If the correction cannot be made within the permitted scope,
return no operations.''
\end{quote}

Schema-assisted single shot and the full workflow share their initial
model response. Model-only and input-changing controls use independent
responses, so their comparisons also include variation between model draws.

\subsection{Example requests and specifications}
\label{app:specification-examples}

The following retained request illustrates a scalar problem:

\begin{quote}
``Solve Poisson on a interval on x in [0,1.34], using structured mesh
with cells\_per\_axis [23]. Use diffusivity 1, source 0. Dirichlet
boundary condition with value 0 on x\_min, x\_max.''
\end{quote}

The corresponding scientific and mesh fields are shown below.
This is an excerpt; the complete retained JSON also includes request
identity, finite-element, time, solver, and output settings.

\begin{lstlisting}[style=appendixjson]
{
  "geometry": {
    "kind": "interval",
    "dimension": 1,
    "x_min": 0.0,
    "x_max": 1.34
  },
  "mesh": {
    "backend": "structured",
    "cells_per_axis": [23],
    "cell_type": null,
    "ghost_mode": "shared_facet"
  },
  "equation": {
    "kind": "poisson",
    "diffusivity": {"kind": "constant", "value": 1.0},
    "source": {"kind": "constant", "value": 0.0}
  },
  "boundary_conditions": [
    {
      "kind": "dirichlet",
      "name": "dirichlet_x_min_x_max",
      "boundary": {
        "kind": "named",
        "names": ["x_max", "x_min"],
        "role": null
      },
      "value": {"kind": "constant", "value": 0.0}
    }
  ]
}
\end{lstlisting}

Incomplete requests retain their intended problem type. For example,
missing time settings in a transient heat request trigger clarification;
the workflow does not silently convert the problem to steady heat.

\subsection{Clarification and amendment patches}
\label{app:clarification-patches}

Updates use explicit patches against the retained draft.
Each patch identifies its contract version, base revision, source turn,
and ordered operations (Table~\ref{tab:appendix-patch-contract}).
The base revision is a SHA-256 hash of the retained context, allowing
code to detect changes made since the patch was prepared.
Application code supplies the permitted scope; the model cannot
authorize its own changes.

\begin{table}[htbp]
  \caption{Patch contract used for clarification, amendment, and repair.}
  \label{tab:appendix-patch-contract}
  \centering
  \small
  \begin{tabularx}{\linewidth}{@{}p{0.29\linewidth}X@{}}
    \toprule
    \textbf{Member} & \textbf{Meaning and requirement} \\
    \midrule
    ``contract\_version''
      & Must identify ``phase2-context-patch-1.0''. \\
    ``base\_revision''
      & Must match the current context revision. \\
    ``source\_turn''
      & Identifies the authorizing user turn or repair event. \\
    ``operations''
      & Up to 32 explicit ``add'', ``replace'', or ``remove''
        operations. An empty proposal is recorded as declined. \\
    ``path''
      & A permitted, non-root JSON Pointer.
        Duplicate or overlapping paths are rejected. \\
    ``value\_json''
      & JSON-encoded value for ``add'' and ``replace'';
        null for ``remove''. \\
    \bottomrule
  \end{tabularx}
\end{table}

\paragraph{Clarification.}
A clarification fills pending information. If the time object exists
but its time-step field is absent, the reply
``Use a time step of 0.05'' authorizes:

\begin{lstlisting}[style=appendixjson]
{
  "op": "add",
  "path": "/draft/problem/time/dt",
  "value_json": "0.05"
}
\end{lstlisting}

The operation is included in a patch carrying the current revision
and reply identifier. If the field already exists, ``replace'' is used.
Missing parent objects require explicitly authorized construction.
A clarification reply cannot silently change unrelated accepted values.

\paragraph{Amendment.}
An explicit amendment can change an accepted value.
For example, ``Change the final time to 2'' authorizes:

\begin{lstlisting}[style=appendixjson]
{
  "op": "replace",
  "path": "/draft/problem/time/t_end",
  "value_json": "2.0"
}
\end{lstlisting}

The original request and evidence remain unchanged. The new statement
and authorized updates enter the user-turn journal, and validation uses
the latest authorized value. A user turn may authorize several fields,
with a limit of 24 updates or removals.

\paragraph{Automatic repair.}
Repair corrects an extraction error using existing evidence and a
trusted repair plan. It does not introduce new user requirements.
Fully determined scalar boundary corrections may be constructed directly
from independently matched request data; other eligible corrections
use the model. Both routes pass through the same patch checks.
A failed repair does not authorize replacement of the entire draft.

\subsection{Revision checks and audit records}
\label{app:revision-audit}

Before accepting a patch, application code checks its revision,
source identifier, permitted paths, and authorized values.
It applies the operations to a private copy, then reruns structural,
scientific, and update-scope checks. Replacement and removal targets
must exist. Array insertion or removal requires permission over the
affected array because indices may shift. The candidate is accepted
only if these transition checks pass; existing unresolved fields may
remain, so acceptance does not imply readiness.

Stale, unauthorized, or invalid proposals leave the accepted scientific
context unchanged. Audit records retain the proposal, source, base and
resulting revisions, acceptance decision, rejection reason, and attempted
changes. Original requests, model responses, normalization records,
provenance, and user amendments are also retained. Revision-checked
storage protects audits and repair counters from conflicting saves.
Execution permission is recomputed when the session is saved or reloaded.

\paragraph{Reproduction materials.}
The source bundle retains the schemas, exact prompts, field vocabulary,
and complete example JSON. Saved model requests preserve the exact
request-specific schemas used in each call. Historical code that returned
a complete replacement draft is retained separately for replay and is
not the local patch protocol described here. Revision hashes detect stale
state and changes; they do not authenticate users or establish scientific
correctness.

\section{Numerical Verification}
\label{app:numerical}

We test the numerical workflow separately from the agent's ability to
construct specifications. Each study starts from a prescribed problem
JSON and uses the registered geometry, mesh, solver, and export routines.
Reference solutions and acceptance rules are defined before evaluating
the results. The original campaign contains 56 completed study solves:
34 core solves, two tighter-tolerance solves, 16 heat-refinement solves,
and four MPI solves. Four development smoke runs are excluded.
Additional verification repetitions are reported separately below.

\subsection{Case-by-case experimental setup}
\label{app:numerical-case-setup}

Table~\ref{tab:case-numerical-setup} summarizes the problems and
discretizations. V1--V7 test accuracy or convergence. V8 checks whether
parallel execution preserves the solution, and V9 checks sensitivity to
solver tolerances. Backward Euler (BE) and Crank--Nicolson (CN) are the
two time-integration schemes used in the transient studies.

\begin{table}[htbp]
  \caption{Setup for the original verification campaign.
  For structured meshes, $n$ denotes cells per coordinate direction.
  For Gmsh, $h$ is the requested characteristic mesh length.}
  \label{tab:case-numerical-setup}
  \centering
  \small
  \begin{tabularx}{\linewidth}{@{}l X X r@{}}
    \toprule
    \textbf{ID} & \textbf{Problem and domain}
      & \textbf{Mesh and time sequence} & \textbf{Solves} \\
    \midrule
    V1 & 2D elasticity; unit square; plane stress
       & P1 triangles; $n=8,16,32,64$ & 4 \\
    V3 & 2D steady flow; unit square with a circular hole
       & P2/P1 velocity/pressure;
         $h=0.06,0.03,0.015$;
         pressure patch at $h=0.06$ & 4 \\
    V4 & 2D unsteady heat; unit-radius disk
       & P1; $h=0.3$;
         five step sizes per scheme and one equilibrium control & 11 \\
    V6 & 2D annular heat; radii $0.35$ and $1$
       & P1; $h=0.00175$;
         three BE and five CN step sizes & 8 \\
    V7 & 2D rectangular heat; unit square
       & P1 triangles; $768\times768$ cells;
         three BE and five CN step sizes & 8 \\
    \midrule
    V2 & 3D elasticity; unit cube
       & P1 tetrahedra; $n=4,8,16,24$ & 4 \\
    V5 & 3D nonlinear reaction--diffusion;
         sphere of radius $0.5$
       & P1; $h=0.3$;
         five step sizes per scheme and one equilibrium control & 11 \\
    V8 & 3D elastic cantilever;
         $2\times0.4\times0.4$ box
       & P1 tetrahedra; $32\times16\times16$ cells;
         4, 8, 16, and 32 MPI processes & 4 \\
    \midrule
    V9 & Tighter-tolerance controls for V3 and V5
       & V3 at $h=0.03$;
         V5 with CN at $\Delta t=0.025$;
         tolerances reduced by a factor of 100 & 2 \\
    \midrule
    \multicolumn{3}{r}{\textbf{Total}} & \textbf{56} \\
    \bottomrule
  \end{tabularx}
\end{table}

All transient studies run from $t=0$ to $T=0.4$.
V4 and V5 use time steps
$0.1,0.05,0.025,0.0125,0.00625$ for both schemes.
V6 and V7 use the first three for BE and all five for CN;
the two extra CN runs estimate the remaining spatial error.
The V4 and V5 meshes contain 91 triangles and 108 tetrahedra,
respectively. V6 and V7 contain 2,380,960 and 1,179,648 triangles.
The V8 mesh contains 49,152 tetrahedra and 28,611 displacement
degrees of freedom.

The original campaign runs on a cloud host with 32 logical CPUs,
an Intel Xeon processor at 2.80\,GHz, and approximately 252\,GiB
of visible RAM. Workers use one numerical-library thread.
Core and heat-refinement studies use one MPI process;
V8 varies the process count explicitly. Concurrent jobs were permitted,
so elapsed times are not used to claim parallel speedup.
Table~\ref{tab:numerical-shared-settings} gives the shared settings.
Inputs, measurements, reference definitions, and hashes are retained.
These studies make no language-model calls.

\begin{table}[htbp]
  \caption{Software and requested solver settings for the original
  campaign. Complete case settings are retained in the input JSON files.}
  \label{tab:numerical-shared-settings}
  \centering
  \small
  \begin{tabularx}{\linewidth}{@{}p{0.30\linewidth}X@{}}
    \toprule
    \textbf{Item} & \textbf{Setting} \\
    \midrule
    Software
      & Python 3.12.3; DOLFINx 0.11.0.post0;
        UFL 2026.1.0; Gmsh 4.12.1;
        PETSc 3.19.6; NumPy 1.26.4 \\
    Linear solver
      & PETSc preonly with LU;
        relative tolerance $10^{-8}$,
        absolute tolerance $10^{-10}$;
        maximum 10,000 iterations \\
    Nonlinear solver
      & Newton with backtracking;
        relative tolerance $10^{-8}$,
        absolute tolerance $10^{-10}$;
        maximum 50 iterations;
        nested linear settings specify GMRES with ILU \\
    Tighter-tolerance controls
      & Relative tolerance $10^{-10}$ and
        absolute tolerance $10^{-12}$ in the linear,
        nonlinear, and nested linear settings \\
    Error integration
      & Quadrature degree 12 for V1--V5;
        degree 8 for V6--V7 and cantilever quantities \\
    Execution checks
      & Finite solution coefficients, converged solver steps,
        and no invalid cells \\
    \bottomrule
  \end{tabularx}
\end{table}

\subsection{Analytical and manufactured references}
\label{app:numerical-references}

A manufactured reference is a chosen exact field whose forcing and
boundary data are derived from the governing equation.
We compare the computed field with this reference using the $L^2$ error
for field values and the $H^1$ seminorm error for gradients:
\begin{equation}
  E_0=\left(\int_{\Omega}\|w_h-w_\star\|^2\,dx\right)^{1/2},
  \qquad
  E_1=\left(\int_{\Omega}
  \|\nabla(w_h-w_\star)\|^2\,dx\right)^{1/2}.
  \label{eq:numerical-error-norms}
\end{equation}
Relative errors divide by the corresponding nonzero reference norm;
the zero-velocity pressure patch uses absolute errors.
For elasticity, V1 and V2 prescribe
$\mathbf{u}_\star=(0.01x^2,0)$ and
$\mathbf{u}_\star=(0.01x^2,0,0)$ on the unit square and cube.
Both use Young's modulus $E=1000$ and Poisson's ratio $\nu=0.3$,
with plane stress in 2D. Exact displacement is imposed on the boundary,
and the body force is derived from that field.
Displacement, strain energy, and reaction balance are checked.

For flow, V3 uses a unit square with a circular hole of radius $0.1$
centred at $(0.25,0.25)$, density $1$, and dynamic viscosity $0.1$.
The reference is defined by
$q=\max(y-0.5,0)$,
$\psi=0.1(1-x)^3q^4$,
$\mathbf{u}_\star=(\partial_y\psi,-\partial_x\psi)$,
and $p_\star=0$.
The body force follows from the steady incompressible
Navier--Stokes equations. Exact velocity is imposed at the inlet
and horizontal boundaries, with zero velocity on the hole and
zero outlet pressure. A separate pressure patch uses
$\mathbf{u}_\star=\mathbf{0}$, $p_\star=1-x$,
and body force $(-1,0)$ on the same domain.

For temporal verification, define
$A_2=0.2+0.1x+0.05y$ and
$A_3=0.2+0.1x+0.05y+0.025z$.
V4 uses $u_\star=A_2e^{-t}$ for
$u_t-\Delta u=f$, with $f=-u_\star$.
V5 uses $u_\star=A_3e^{-t}$ for
$u_t-\Delta u+u^3=f$, with $f=-u_\star+u_\star^3$.
Both use exact initial and boundary values.
Their affine spatial fields can be represented by P1 elements,
allowing time-discretization error to be studied separately.
Stationary controls use $u_\star=A_2$ or $A_3$,
with sources zero and $A_3^3$, respectively.
V6 and V7 instead use
$u_\star=0.2(\sin x+\cos y)e^{-t}$ for
$u_t-\Delta u=0$.
These fields have nonzero spatial error, which must be checked
before interpreting temporal convergence.

\subsection{Spatial convergence}
\label{app:spatial-convergence}

\begin{figure}[htbp]
  \centering
  \includegraphics[width=\linewidth]{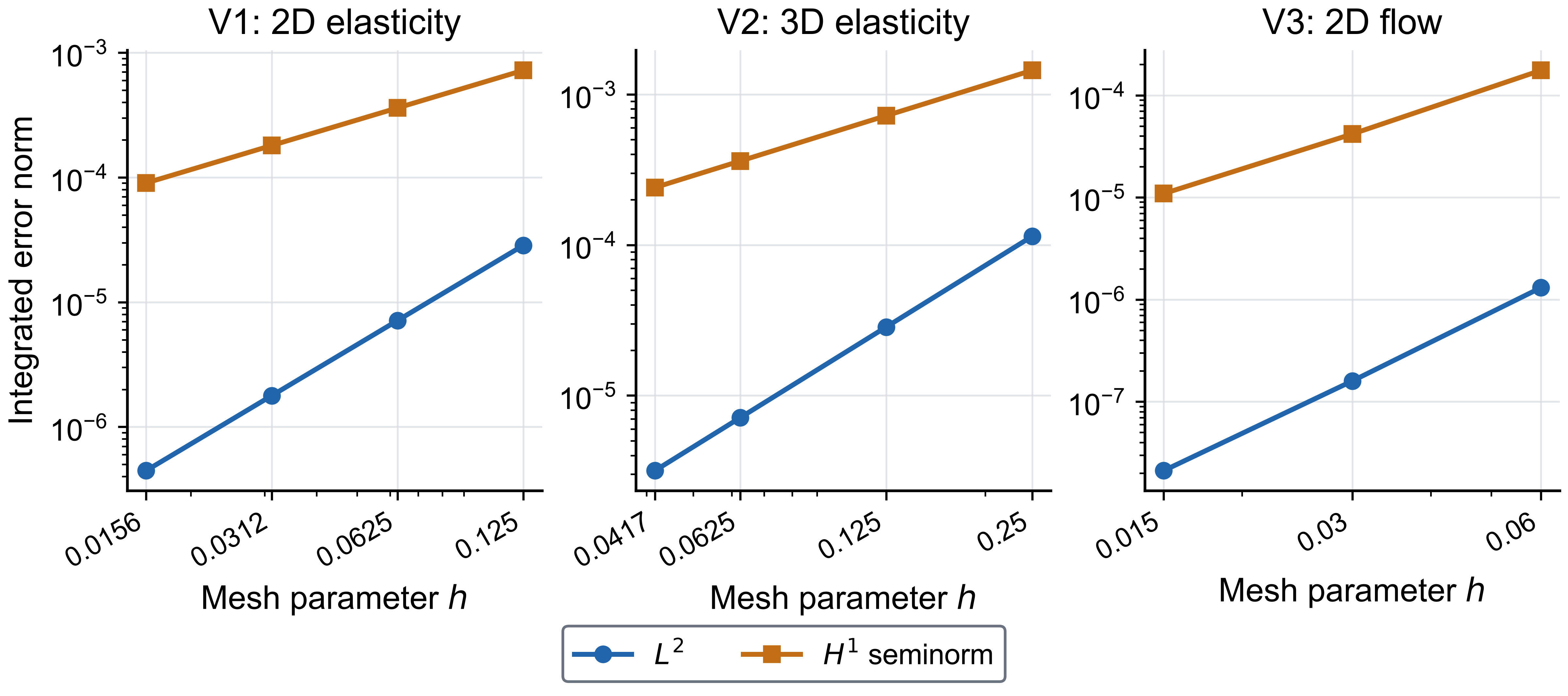}
  \caption{Spatial convergence for V1--V3.
  Orders are fitted using the finest three levels.
  Elasticity uses P1 displacement elements;
  flow uses P2 velocity and P1 pressure.
  The $L^2$ and $H^1$ seminorm errors have different units.}
  \label{fig:spatial-convergence}
\end{figure}

Spatial convergence measures how the error decreases as the mesh
is refined. In the original campaign, observed orders are fitted
from the finest three levels using logarithmic error and mesh-size
values. For structured meshes, $h=1/n$; for flow, $h$ is the requested
Gmsh characteristic length.
Figure~\ref{fig:spatial-convergence} shows all completed levels,
and Table~\ref{tab:numerical-spatial-results} gives the fitted orders
and finest-mesh errors.

\begin{table}[htbp]
  \caption{Spatial verification results.
  Errors are absolute integrated norms on the finest mesh.
  The $H^1$ column reports seminorm errors.}
  \label{tab:numerical-spatial-results}
  \centering
  \small
  \begin{tabularx}{\linewidth}{@{}l X c c c@{}}
    \toprule
    \textbf{ID} & \textbf{Field}
      & \shortstack{$L^2$ order\\and error}
      & \shortstack{$H^1$ order\\and error}
      & \textbf{Outcome} \\
    \midrule
    V1 & 2D displacement
       & \shortstack{2.000\\$4.46\times10^{-7}$}
       & \shortstack{1.000\\$9.02\times10^{-5}$}
       & Pass \\
    V2 & 3D displacement
       & \shortstack{2.000\\$3.17\times10^{-6}$}
       & \shortstack{1.000\\$2.41\times10^{-4}$}
       & Pass \\
    V3 & 2D velocity
       & \shortstack{2.976\\$2.11\times10^{-8}$}
       & \shortstack{2.004\\$1.09\times10^{-5}$}
       & Pass \\
    \bottomrule
  \end{tabularx}
\end{table}

All three studies pass their declared checks.
Accepted $L^2$ and $H^1$ order bands are
$[1.5,2.5]$ and $[0.7,1.4]$ for elasticity,
and $[2.2,3.8]$ and $[1.4,2.6]$ for velocity.
Finest-mesh relative errors must also remain below
$3\%$ in $L^2$ and $15\%$ in $H^1$.
Elasticity additionally requires strain-energy error
at most $2\%$ and reaction imbalance at most $1\%$.

\subsection{Temporal convergence and spatial isolation}
\label{app:temporal-isolation}

\begin{figure}[htbp]
  \centering
  \includegraphics[width=\linewidth]{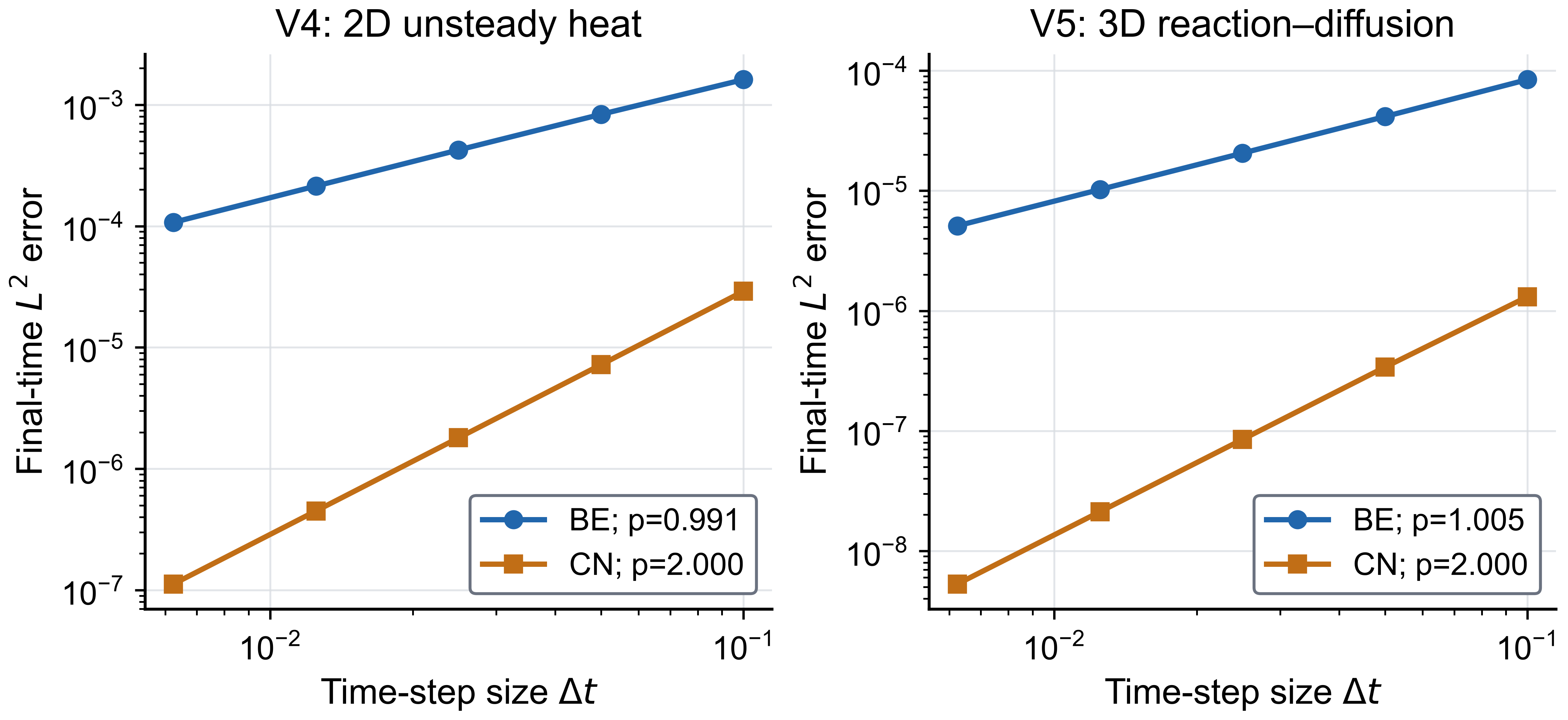}
  \caption{Temporal convergence for V4 and V5 at $T=0.4$.
  All five time steps are shown; fitted orders use the finest three.
  Stationary controls are checked separately.}
  \label{fig:temporal-convergence}
\end{figure}

Temporal convergence measures how the error decreases when the time
step is reduced. V4 and V5 recover approximately first-order BE
and second-order CN behaviour
(Figure~\ref{fig:temporal-convergence}).
Orders use the finest three of the five step sizes.
The stationary controls have $L^2$ errors of
$1.06\times10^{-16}$ and $2.02\times10^{-17}$,
respectively. Both are below the $10^{-8}$ limit and below
$10\%$ of the smallest dynamic error.

For V6 and V7, mesh error must be small enough to distinguish it from
time-integration error. We estimate the remaining spatial error using
two extra CN runs on the same mesh.
With $\delta=0.025$, the extrapolated field is
\begin{equation}
  u_{\mathrm{ext}}=
  \frac{4u_h^{\mathrm{CN}}(\delta/4)
  -u_h^{\mathrm{CN}}(\delta/2)}{3}.
\end{equation}
Its difference from the exact solution estimates the spatial error.
For each scheme $S$, this estimate is compared with the difference
between the primary-step solution and the extrapolated field:
\begin{equation}
  R_S=
  \frac{\|u_{\mathrm{ext}}-u_\star(T)\|_{L^2}}
  {\max(\|u_h^S(\delta)-u_{\mathrm{ext}}\|_{L^2},10^{-30})}.
  \label{eq:numerical-spatial-isolation}
\end{equation}
Both schemes must satisfy $R_S\leq10\%$.
Mesh and coordinate identity are checked before combining fields,
and all fields come from the same campaign.

\begin{table}[htbp]
  \caption{Temporal verification results.
  Accepted order bands are $[0.8,1.2]$ for BE and
  $[1.65,2.35]$ for CN.
  The spatial ratio is the larger of the two scheme ratios;
  CN determines it for both V6 and V7.}
  \label{tab:numerical-temporal-results}
  \centering
  \small
  \begin{tabularx}{\linewidth}{@{}l X c c c l@{}}
    \toprule
    \textbf{ID} & \textbf{Study}
      & \textbf{BE order} & \textbf{CN order}
      & \textbf{Spatial ratio} & \textbf{Outcome} \\
    \midrule
    V4 & Affine heat
       & 0.991 & 2.000 & Control passed & Pass \\
    V5 & Affine reaction--diffusion
       & 1.005 & 2.000 & Control passed & Pass \\
    V6 & Annular heat
       & 1.011 & 1.941 & $10.84\%$ & Fail \\
    V7 & Rectangular heat
       & 1.008 & 1.961 & $6.83\%$ & Pass \\
    \bottomrule
  \end{tabularx}
\end{table}

\begin{figure}[htbp]
  \centering
  \includegraphics[width=\linewidth]{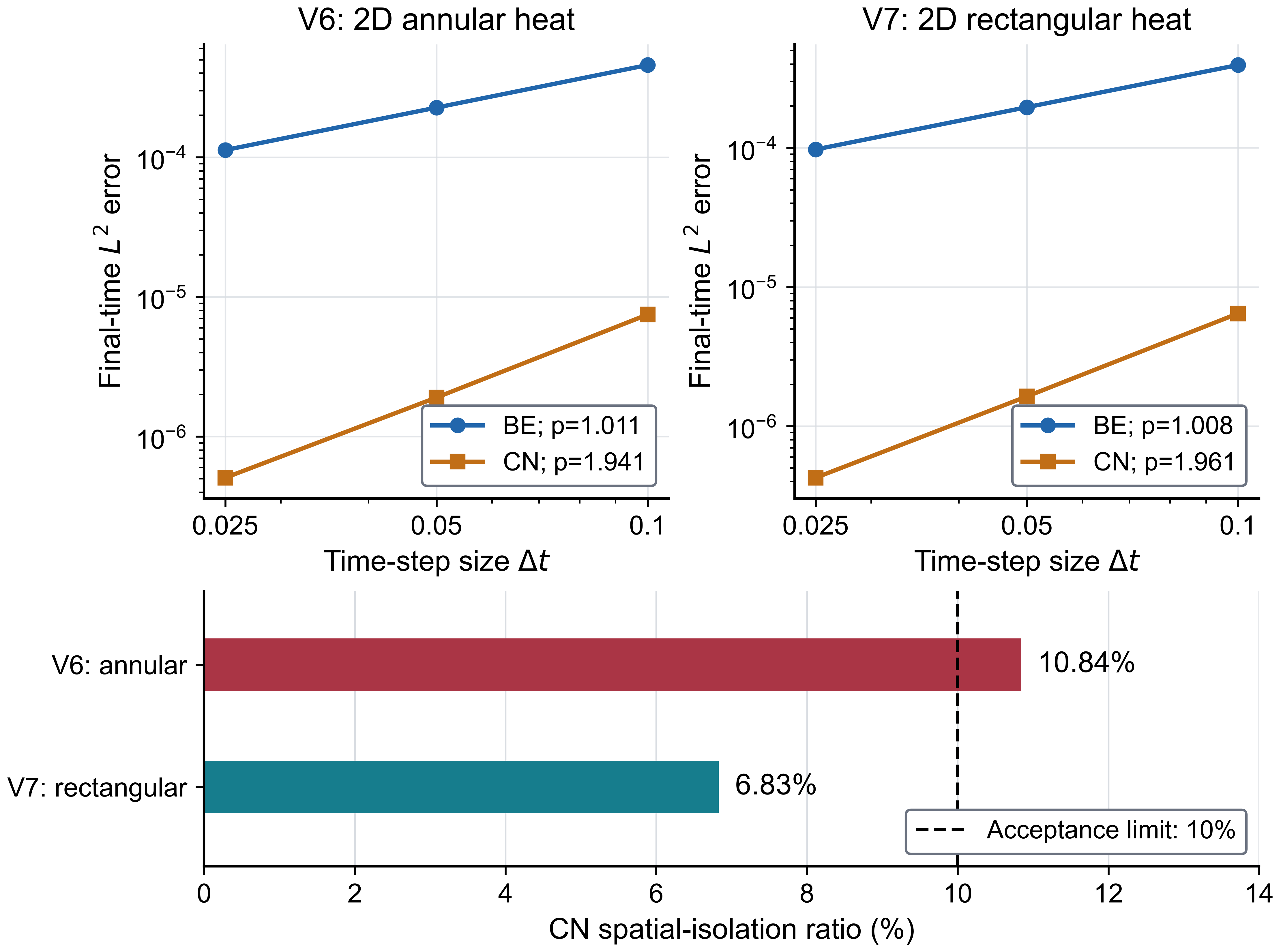}
  \caption{Temporal refinement with nonzero spatial error.
  Upper panels show reference errors at the three primary time steps.
  Two additional CN runs estimate the spatial error.
  The lower panel shows the CN ratio from
  Equation~\ref{eq:numerical-spatial-isolation}.
  V6 exceeds the $10\%$ limit; V7 passes.}
  \label{fig:heat-spatial-isolation}
\end{figure}

V6 and V7 have estimated spatial errors of
$5.04\times10^{-8}$ and $2.75\times10^{-8}$.
Their BE ratios are $0.0448\%$ and $0.0283\%$,
while their CN ratios are $10.84\%$ and $6.83\%$.
V7 passes both checks. V6 passes the temporal-order bands but fails
the spatial-isolation requirement; its failure is retained.

\subsection{Nonlinear, flow, elasticity, and balance checks}
\label{app:numerical-additional-checks}

All completed study solves produce finite fields, converged solver
steps, and valid cells. The core records require at most two nonlinear
iterations per solve or time step, within the limit of 50.
The finest V1 and V2 strain-energy errors are $0.00610\%$ and
$0.0434\%$, and both satisfy the $1\%$ reaction-balance limit.
These results apply to the prescribed verification problems.

For the finest V3 flow mesh, the pressure $L^2$ error is
$1.06\times10^{-7}$ and the velocity-divergence norm is
$5.07\times10^{-6}$, both below their $2\times10^{-3}$ limits.
The pressure patch has velocity and pressure errors of
$1.13\times10^{-13}$ and $1.41\times10^{-12}$,
below the separate $10^{-8}$ limits.
The nonzero-flow case has relative boundary-flux imbalance
$7.60\times10^{-9}$, reported as a diagnostic.
For the zero-flow patch, net boundary flux is approximately
$-6.19\times10^{-14}$; a normalized imbalance is unsuitable
because the reference through-flow is zero.
A separate transient heat-balance check was not measured.

\subsection{MPI and tighter-tolerance agreement}
\label{app:numerical-agreement}

V8 checks whether dividing one simulation among different numbers
of cooperating MPI processes changes the solution.
The same cantilever is solved using 4, 8, 16, and 32 processes
and compared with a retained single-process reference.
Its root is clamped, tip traction is $(0,0,-0.01)$,
and other faces are unloaded, with $E=1000$ and $\nu=0.3$.
Displacement coefficients are matched by physical coordinates;
rounding to $10^{-12}$ is used only for indexing, and original
values and duplicate ghost entries are checked.
We compare the displacement field, mean tip deflection, strain energy,
and integrated stress over $0.2\leq x\leq1.8$,
excluding clamp and load edges.
All four runs satisfy the $10^{-7}$ relative-difference limit.

\begin{figure}[htbp]
  \centering
  \includegraphics[width=\linewidth]{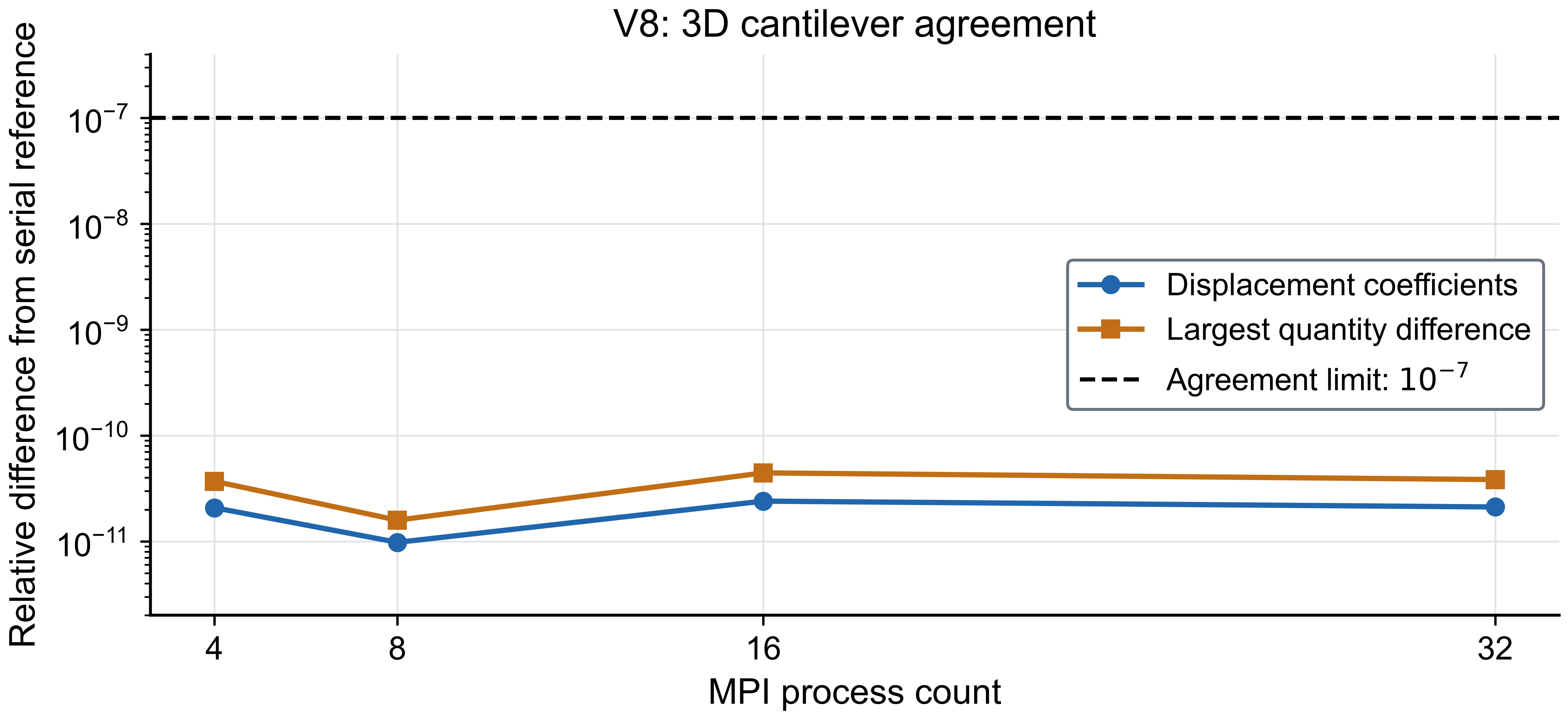}
  \caption{MPI agreement on the same cantilever mesh.
  Displacement-coefficient differences are normalized by the serial
  coefficient-vector norm. The second curve gives the largest relative
  difference in tip deflection, strain energy, or integrated stress,
  normalized by the corresponding absolute serial value.
  The test measures consistency, not parallel speedup.}
  \label{fig:mpi-agreement}
\end{figure}

V9 repeats the flow and reaction--diffusion solves with tighter
tolerances while keeping the problem and discretization unchanged.
Coefficient differences are normalized by the baseline vector norm.
Recorded error-metric differences are normalized by the larger absolute
baseline or tightened value. Both must remain below $10^{-6}$.
Both comparisons pass
(Table~\ref{tab:numerical-agreement-results}).

\begin{table}[htbp]
  \caption{Parallel and tighter-tolerance agreement.
  The reported-quantity column gives the largest relative difference
  among cantilever quantities for V8 or recorded error metrics for V9.
  Zero indicates no difference at saved numerical precision.}
  \label{tab:numerical-agreement-results}
  \centering
  \small
  \begin{tabularx}{\linewidth}{@{}l X c c c@{}}
    \toprule
    \textbf{ID} & \textbf{Comparison}
      & \shortstack{Native-field\\difference}
      & \shortstack{Reported-quantity\\difference}
      & \textbf{Limit} \\
    \midrule
    V8 & 4 MPI processes
       & $2.08\times10^{-11}$ & $3.70\times10^{-11}$ & $10^{-7}$ \\
       & 8 MPI processes
       & $9.79\times10^{-12}$ & $1.59\times10^{-11}$ & $10^{-7}$ \\
       & 16 MPI processes
       & $2.40\times10^{-11}$ & $4.44\times10^{-11}$ & $10^{-7}$ \\
       & 32 MPI processes
       & $2.11\times10^{-11}$ & $3.84\times10^{-11}$ & $10^{-7}$ \\
    \midrule
    V9 & 2D flow, tighter tolerances
       & $8.12\times10^{-10}$ & $3.01\times10^{-7}$ & $10^{-6}$ \\
       & 3D reaction--diffusion, tighter tolerances
       & $0$ & $0$ & $10^{-6}$ \\
    \bottomrule
  \end{tabularx}
\end{table}

\subsection{Additional verification repetitions}
\label{app:numerical-repetitions}

A separate study on 19 October 2026 repeats selected verification
problems using their retained input JSON and reference fields.
All 26 solves complete, and all 12 declared regression checks pass.
The runs use the local WSL environment with one MPI process,
one numerical-library thread, and a 180-second limit per task.
Table~\ref{tab:numerical-repetitions} summarizes the results.
These runs repeat existing problems and are recorded separately
from the original 56-solve campaign.

The repeated transient studies use
$\Delta t=0.1,0.05,0.025$ for each scheme.
The 12 checks cover solver and mesh health, convergence orders,
equilibrium controls, the pressure patch, and coefficient agreement
under tighter tolerances. These repetitions do not repeat the V6--V7
spatial-isolation studies or the V8 MPI comparison, and do not change
their original qualifications.

\begin{table}[htbp]
  \caption{Additional verification repetitions.
  Spatial results give $L^2/H^1$ seminorm orders;
  temporal results give BE/CN orders.
  Orders use the two finest levels, whereas the original campaign
  uses three-level fits.}
  \label{tab:numerical-repetitions}
  \centering
  \small
  \begin{tabularx}{\linewidth}{@{}l X r X c@{}}
    \toprule
    \textbf{Study} & \textbf{Repeated setup}
      & \textbf{Solves} & \textbf{Result} & \textbf{Outcome} \\
    \midrule
    V1 & 2D elasticity;
         $n=8,16,32$
       & 3 & Orders: 2.000 / 1.000 & Pass \\
    V3 & 2D flow;
         three mesh levels and pressure patch
       & 4 & Orders: 2.917 / 1.938;
             pressure patch passed & Pass \\
    V4 & 2D heat;
         three steps per scheme and equilibrium control
       & 7 & Orders: 0.975 / 2.003;
             control passed & Pass \\
    \midrule
    V2 & 3D elasticity;
         $n=4,8,16$
       & 3 & Orders: 2.000 / 1.000 & Pass \\
    V5 & 3D reaction--diffusion;
         three steps per scheme and equilibrium control
       & 7 & Orders: 1.012 / 2.000;
             control passed & Pass \\
    \midrule
    V9 & Flow and reaction--diffusion
         with tighter tolerances
       & 2 & Field differences:
             $8.12\times10^{-10}$ and $0$ & Pass \\
    \midrule
    \multicolumn{2}{r}{\textbf{Total}}
       & \textbf{26} & & \\
    \bottomrule
  \end{tabularx}
\end{table}

\subsection{Scope and remaining limitations}
\label{app:numerical-limitations}

V6 remains a verification failure because its spatial-isolation ratio
exceeds the declared limit. The optional finest cantilever refinement
was not launched because its prerequisite was checked before the MPI
comparison finished. No finest-mesh result was produced, so it is
unattempted rather than a pass or numerical failure.
Agreement across MPI processes establishes consistency on the tested
mesh; it does not establish mesh independence.

Original outcomes, additional repetitions, and development runs retain
their separate records. Gallery qualifications are reported in
Appendix~\ref{app:gallery}. These studies verify the numerical
implementation within the tested scope; they do not establish
the physical validity of every gallery problem or the correctness
of an agent-generated specification.

\section{End-to-End Workflow Tests}
\label{app:e2e}

\subsection{Case selection and success definition}

We evaluate 12 selected requests: three each from the easy, medium,
hard, and advanced-hard groups. They cover three 1D, four 2D, and
five 3D scalar problems, including Poisson, transient heat, and
nonlinear reaction--diffusion. These are known, code-authored,
template-related requests used for regression testing. The study uses
workflow v1.9 with fresh model responses, the model and generation
settings in Section~\ref{sec:setup}, and at most one model patch repair.
The worker executes the revision-checked accepted specification without
an author-written replacement. Flow, elasticity, and wing examples
are covered separately in Appendix~\ref{app:gallery}.

Joint success requires the intended specification, a converged solve
with finite fields, complete required exports, and a passed reference
check. All 12 requests remain in the denominator, including blocked
attempts. Prescribed fields are zero or affine fields of the form
$a+x$. Zero-field cases require absolute finite-element $L^2$ error
at most $10^{-10}$. Affine cases require relative $L^2$ error at most
$10^{-9}$ for linear problems or $10^{-7}$ for nonlinear problems.
References assess the computed field after solving and do not supply
the executor's specification. Transient references are constant in time;
they test the execution path, while temporal accuracy is assessed
separately in Appendix~\ref{app:numerical}.

\subsection{All selected live outcomes}

The workflow achieves joint success on $9/12$ requests ($75.0\%$).
All nine executed cases preserve the intended specification, pass
their reference checks, and produce the required exports.
They include four 3D cases and three nonlinear cases.
Three requests remain blocked before solving.
Table~\ref{tab:e2e-live-outcomes} reports all selected attempts.

\begin{table}[htbp]
  \caption{All selected live requests, ordered by dimension.
  Verified means joint success: correct specification, converged solve,
  complete exports, and a passed reference check.
  Blocked means no solver invocation.}
  \label{tab:e2e-live-outcomes}
  \centering
  \small
  \begin{tabularx}{\linewidth}{@{}l X l@{}}
    \toprule
    \textbf{ID} & \textbf{Requested problem} & \textbf{Outcome} \\
    \midrule
    E1 & 1D Poisson, interval ($L=1.340$) & Verified \\
    E2 & 1D Poisson, interval ($L=2.153$) & Verified \\
    E3 & 1D Poisson, interval ($L=2.257$) & Blocked \\
    \midrule
    E4 & 2D Poisson, rectangle & Verified \\
    E5 & 2D Poisson, annulus & Verified \\
    E6 & 2D unsteady heat, rectangle & Blocked \\
    E7 & 2D unsteady reaction--diffusion, annulus & Verified \\
    \midrule
    E8 & 3D Poisson, box & Verified \\
    E9 & 3D nonlinear Poisson, cylinder & Verified \\
    E10 & 3D Poisson, sphere & Verified \\
    E11 & 3D unsteady reaction--diffusion, box & Verified \\
    E12 & 3D unsteady reaction--diffusion, sphere & Blocked \\
    \bottomrule
  \end{tabularx}
\end{table}

\subsection{Execution timings and saved artifacts}

\begin{figure}[h]
  \centering
  \includegraphics[width=\linewidth]{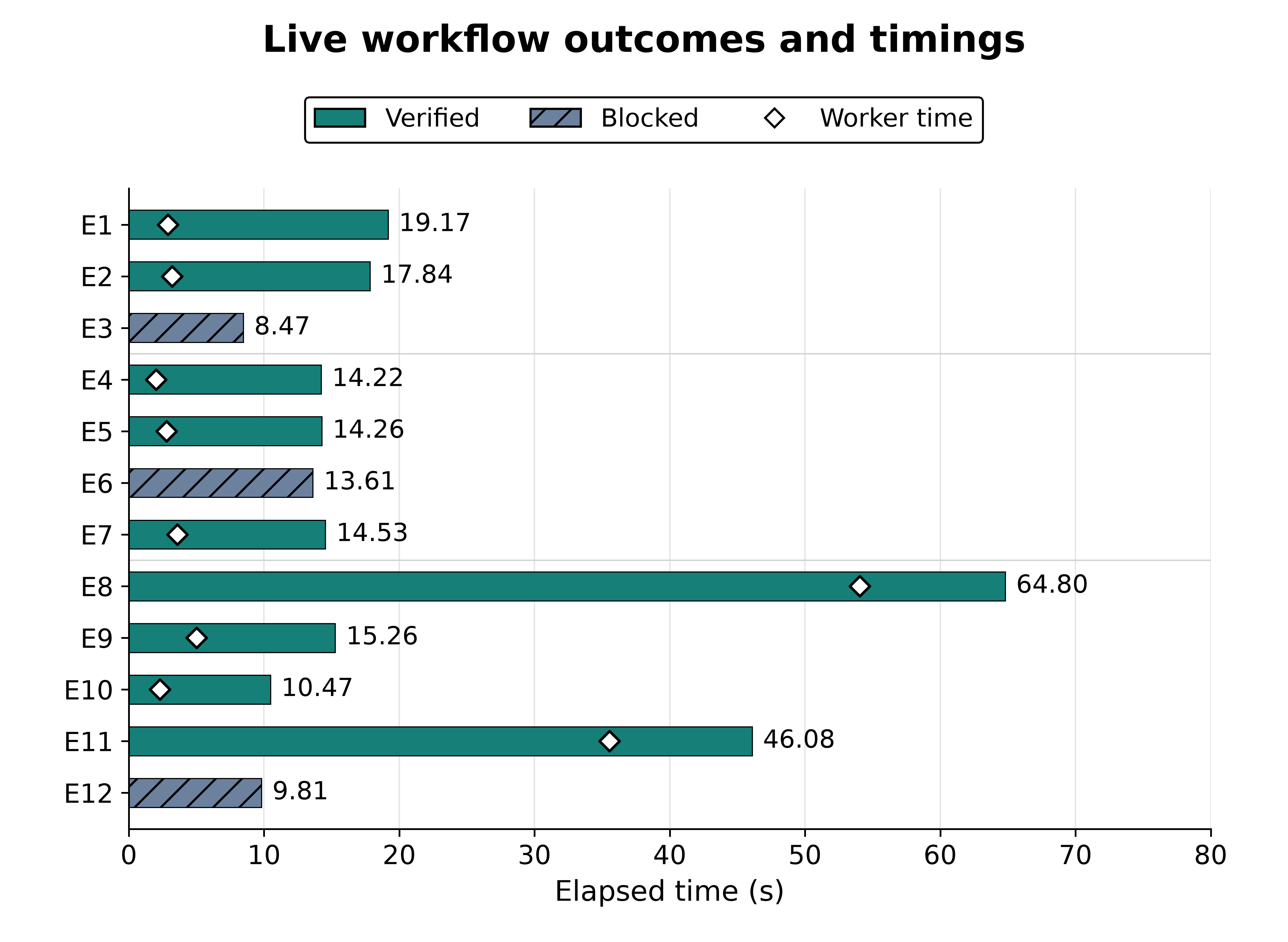}
  \caption{Outcomes and timings for all 12 requests.
  Bars show elapsed time from request receipt to the saved result
  or blocked outcome. Diamonds show worker time for executed cases.
  The clocks are measured separately and are not additive.
  Blocked cases have no numerical-execution time.}
  \label{fig:e2e-outcomes}
\end{figure}

The attempts run locally on WSL with one MPI process,
one numerical-library thread, and a 900-second worker timeout.
For the nine completed cases, request-to-results time ranges from
10.47 to 64.80~s, with a median of 15.26~s.
Recorded worker time ranges from 2.01 to 54.05~s;
its clock starts after imports and includes session checks,
numerical execution, reference assessment, and export
(Figure~\ref{fig:e2e-outcomes}). The model-only baseline is scheduled
after live execution and is excluded from its turnaround time.

Each completed case retains the accepted JSON, patch audit, generated
mesh, native solution field, solver diagnostics, result metadata,
and geometry, mesh, and solution views. Original requests, retained
sessions, outcome records, and artifact hashes link the input to the
executed specification and results. Blocked cases retain their session
and failure records; their terminal times record when blocking was saved.
Some internal stage timers overlap and are not added to reconstruct
turnaround time.

\subsection{Remaining failures and scope}

E3, E6, and E12 remain blocked because extracted boundary, initial,
or source data conflict with the request, lack required expression
values, or fail provenance checks. Their repair proposals do not pass
the update checks. These are failures to complete the requested task;
blocking does not establish that the user request itself is invalid.
The successful cases use zero or affine reference fields, so they
provide limited evidence of complete workflow reliability.


\section{Additional Agentic Studies}
\label{app:additional-agentic}

This section reports development and confirmation studies, detailed
main-benchmark results, component checks, clarification completion,
broader requests, gallery-specification reconstruction, and regression
tests of updated workflows. Unless stated otherwise, a correct first-turn
outcome requires the reference scientific specification and decision to
proceed, clarify, or reject. Completing a clarification dialogue is a
separate measure. These tests assess specification construction; numerical
verification and live simulation are reported in their respective appendices.

\subsection{Development and confirmation studies}

Table~\ref{tab:agentic-history} reports the development set, initial holdout,
four confirmation studies, and the main benchmark (A1). Confirmation
studies 1--4 use workflow versions v1.2, v1.5, v1.6, and v1.7; A1 uses v1.8.
The earlier workflow repairs a complete regenerated draft within checked
limits. From v1.2 onward, repairs use versioned path operations against a
retained specification. Once failures in a cohort were inspected, that
cohort became development evidence for later changes. The table retains
the original results: requests, workflow versions, and some reference and
scoring conditions differ, so the rows are not a controlled improvement curve.

\begin{table}[htbp]
  \caption{Originally recorded first-turn specification and decision
  outcomes. Entries are correct/evaluated requests (percentage).
  SS denotes schema-assisted single shot. A dash means that no model-only
  result was recorded for that study. Model-only uses a separate prompt
  and model draw.}
  \label{tab:agentic-history}
  \centering
  \small
  \begin{tabularx}{\linewidth}{@{}Xccc@{}}
    \toprule
    \textbf{Study} & \textbf{Model-only} & \textbf{SS}
      & \textbf{Full workflow} \\
    \midrule
    Development & --- & 38/60 (63.3\%) & 46/60 (76.7\%) \\
    Initial holdout & --- & 22/40 (55.0\%) & 20/40 (50.0\%) \\
    Confirmation 1 & --- & 40/40 (100.0\%) & 26/40 (65.0\%) \\
    Confirmation 2 & --- & 37/40 (92.5\%) & 38/40 (95.0\%) \\
    Confirmation 3 & --- & 38/40 (95.0\%) & 37/40 (92.5\%) \\
    Confirmation 4 & 32/40 (80.0\%) & 37/40 (92.5\%) & 39/40 (97.5\%) \\
    \midrule
    Main study (A1) & 99/120 (82.5\%) & 111/120 (92.5\%) & 118/120 (98.3\%) \\
    \bottomrule
  \end{tabularx}
\end{table}

The full workflow does not consistently outperform single shot during
development. For example, it falls below single shot in the initial holdout
and first confirmation study, while the last confirmation study obtains
39/40 correct outcomes (97.5\%). In A1, it obtains 118/120 (98.3\%), compared
with 111/120 (92.5\%) for single shot and 99/120 (82.5\%) for model-only.
Single shot and the full workflow share the initial response in A1; their
comparison measures the added processing of that response. The independently
sampled model-only arm is a complete-system comparison. Later regression
results are reported separately below and do not replace these scores.

\subsection{Main-benchmark results by task and request type}
\label{app:repro-breakdowns}

A1 contains 48 extraction, 30 complete-specification, 30 clarification,
and 12 rejection/conflict requests. Extraction requests test the scientific
fields required by the reference problem; complete-specification requests
also ask for the full executable configuration. Neither task involves a
solver run in this agentic benchmark. Each task uses the same scorer across
the three systems. Table~\ref{tab:agentic-breakdowns} shows results by task,
wording, and difficulty; these are separate partitions of the same 120
requests and must not be added together. The full workflow correctly handles
all first-turn clarification and rejection cases, while its two remaining
failures occur on complete requests. Across the 78 complete requests, it
returns 76 correct specifications (97.4\%). False-ready decisions among the
42 requests requiring clarification or rejection are 6/42 for model-only,
1/42 for single shot, and 0/42 for the full workflow.

\begin{table}[htbp]
  \caption{Main-benchmark first-turn outcomes by task, wording, and
  difficulty. Entries are correct/evaluated requests (percentage).
  M is model-only, S is schema-assisted single shot, and F is the full
  workflow. Each block separately accounts for all 120 requests.}
  \label{tab:agentic-breakdowns}
  \centering
  \small
  \begin{tabularx}{\linewidth}{@{}Xccc@{}}
    \toprule
    \textbf{Request group} & \textbf{M} & \textbf{S} & \textbf{F} \\
    \midrule
    \multicolumn{4}{@{}l}{\emph{Task}} \\
    Extraction & 42/48 (87.5\%) & 47/48 (97.9\%) & 47/48 (97.9\%) \\
    Complete specification & 28/30 (93.3\%) & 28/30 (93.3\%) & 29/30 (96.7\%) \\
    Clarification & 17/30 (56.7\%) & 30/30 (100.0\%) & 30/30 (100.0\%) \\
    Rejection/conflict & 12/12 (100.0\%) & 6/12 (50.0\%) & 12/12 (100.0\%) \\
    \midrule
    \multicolumn{4}{@{}l}{\emph{Wording}} \\
    Natural & 44/60 (73.3\%) & 54/60 (90.0\%) & 58/60 (96.7\%) \\
    Schema-explicit & 55/60 (91.7\%) & 57/60 (95.0\%) & 60/60 (100.0\%) \\
    \midrule
    \multicolumn{4}{@{}l}{\emph{Difficulty}} \\
    Easy & 5/5 (100.0\%) & 5/5 (100.0\%) & 5/5 (100.0\%) \\
    Medium & 37/42 (88.1\%) & 36/42 (85.7\%) & 42/42 (100.0\%) \\
    Hard & 50/65 (76.9\%) & 64/65 (98.5\%) & 64/65 (98.5\%) \\
    Advanced hard & 7/8 (87.5\%) & 6/8 (75.0\%) & 7/8 (87.5\%) \\
    \bottomrule
  \end{tabularx}
\end{table}

\begin{figure}[htbp]
  \centering
  \includegraphics[width=\linewidth]{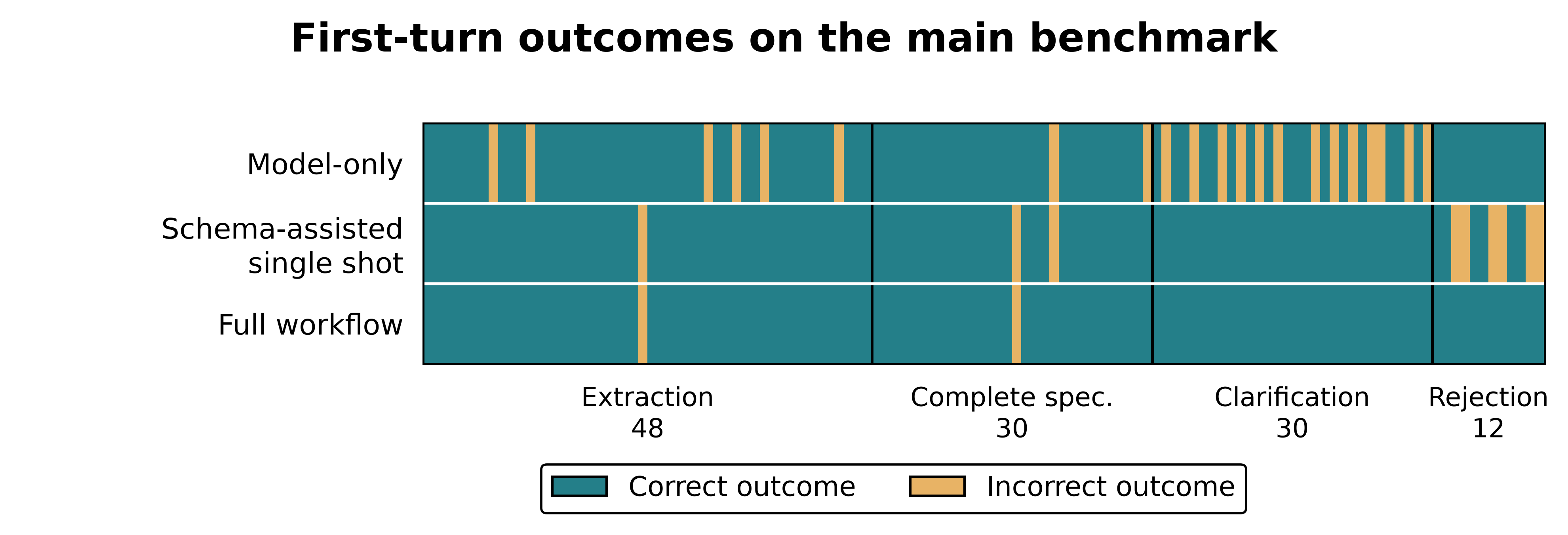}
  \caption{First-turn outcomes for all 120 main-benchmark requests.
  Each column is one request, with aligned cells for all three systems.
  Requests are grouped by task and then by their original record ID;
  labels give task counts. Correct clarification and rejection decisions
  count as correct outcomes. The cells do not indicate solver execution.}
  \label{fig:repro-case-outcomes}
\end{figure}

The wording comparison separates natural requests from requests that name
schema fields explicitly. Difficulty counts five specification features:
3D or non-basic geometry, mixed or varying boundary data, nonlinearity,
time dependence, and expressions.
For non-rejection tasks, zero feature groups is easy, one is medium, two
or three is hard, and four or five is advanced hard. Rejection requests use
a separate rule based on geometry and time wording. The main collection was classified
retrospectively using the existing rule. These categories do not measure
solver runtime or engineering difficulty. Their sizes are uneven, so the
small easy and advanced groups provide limited evidence. The requests also
share construction templates. Figure~\ref{fig:repro-case-outcomes} shows
the individual outcomes rather than only their totals.

\subsection{Component checks and clarification completion}
\label{app:repro-case-records}

Table~\ref{tab:additional-components} compares component removals in the
last confirmation study and A1. Each variant uses its own matched
full-workflow reference. Downstream removals reuse retained responses;
the repair comparison uses recorded outcomes without repair. Provider-schema
and routing controls change the model input and use new responses, so their
differences also include model variation. Removing semantic/capability
checks reduces correct outcomes in both studies. Removing the required
output formatting causes parsing failures throughout, which measures
dependence on the interface rather than scientific reasoning alone.
Scientific grounding and repair do not change the last confirmation
study's total, but affect A1. Their effects overlap and must not be added.

\begin{table}[htbp]
  \caption{Component-removal outcomes in two studies. Entries are
  correct/evaluated requests (percentage). The first block uses the full
  cohorts; the second uses the input-control requests and its own reference.
  No combined accuracy is formed across the two studies.}
  \label{tab:additional-components}
  \centering
  \small
  \begin{tabularx}{\linewidth}{@{}Xcc@{}}
    \toprule
    \textbf{Condition} & \shortstack{\textbf{Confirmation 4}\\\textbf{(v1.7)}}
      & \shortstack{\textbf{Main study}\\\textbf{(v1.8)}} \\
    \midrule
    \multicolumn{3}{@{}l}{\emph{Full-cohort comparisons}} \\
    Full-workflow reference & 39/40 (97.5\%) & 118/120 (98.3\%) \\
    Without repair & 39/40 (97.5\%) & 115/120 (95.8\%) \\
    Without scientific grounding & 39/40 (97.5\%) & 117/120 (97.5\%) \\
    Without semantic/capability checks & 37/40 (92.5\%) & 114/120 (95.0\%) \\
    Without readiness gate & 38/40 (95.0\%) & 117/120 (97.5\%) \\
    Without output canonicalization/evidence formatting & 0/40 (0.0\%) & 0/120 (0.0\%) \\
    \midrule
    \multicolumn{3}{@{}l}{\emph{Input controls: matched requests}} \\
    Matched full-workflow reference & 39/40 (97.5\%) & 40/40 (100.0\%) \\
    Without provider schema enforcement & 36/40 (90.0\%) & 36/40 (90.0\%) \\
    Without geometry/equation routing & 23/40 (57.5\%) & 26/40 (65.0\%) \\
    Without mapping hints & 39/40 (97.5\%) & 40/40 (100.0\%) \\
    \bottomrule
  \end{tabularx}
\end{table}

Evaluated and exercised counts answer different questions. In A1, model
repair is exercised on 5/120 requests, semantic/capability checks on 12/120,
and the readiness gate on 4/120. Mapping hints are not exercised in either
study, so their unchanged scores do not show that they are unnecessary.
Table~\ref{tab:additional-clarification} evaluates a different question:
whether fixed authorized replies lead to a correct complete specification.
Replies update retained sessions through validated patches; this is an
offline interaction test, not a human-user study. Removing clarification
produces no completed specifications in these controls. Removing question
priority leaves completion unchanged; the original two studies have only
one pending field per request, so they do not test competing question orders.

\begin{table}[htbp]
  \caption{Correct complete specifications after fixed authorized
  clarification replies. Each row uses its own retained responses and
  selected clarification requests; these are not first-turn accuracy scores.
  Later rows reuse known requests rather than adding independent cases.}
  \label{tab:additional-clarification}
  \centering
  \small
  \begin{tabularx}{\linewidth}{@{}Xcc@{}}
    \toprule
    \textbf{Study} & \textbf{Correct / evaluated} & \textbf{Percentage} \\
    \midrule
    Confirmation 4 (v1.7) & 10/10 & 100.0\% \\
    Main study: A3 (v1.8) & 28/30 & 93.3\% \\
    Earlier-cohort rerun (v1.8) & 32/50 & 64.0\% \\
    Main-study subset (v1.9) & 14/18 & 77.8\% \\
    \bottomrule
  \end{tabularx}
\end{table}

Patch-scope and revalidation checks are also tested using retained proposals.
In the earlier-cohort regression, 13 of 65 proposals are accepted with all
checks, compared with 16 when scope checking is removed and 36 when
revalidation is removed. These are read-only acceptance tests: no proposal
from a disabled-check condition is committed, and a higher acceptance
count is not an accuracy improvement. Only six of the 50 clarification
interactions in that regression have multiple eligible questions, which
also limits conclusions about question priority.

\subsection{Regression tests of updated workflows}
\label{app:workflow-regression}

Earlier cached-response checks give identical scored outcomes for v1.7
and v1.8 on the development set (55/60) and confirmation studies 1 and 3
(40/40 each). A separate v1.8 rerun with fresh responses to 200 earlier
holdout and confirmation requests obtains 131/200 correct outcomes
(65.5\%) for the full workflow and 126/200 (63.0\%) for single shot.
These tests reuse known requests and retain their original evaluation
conditions; they do not replace the historical scores.

A6 evaluates the v1.9 changes using the retained final JSON contexts
from A1 and A4. The changes address expression handling, request-evidence
matching, duplicate boundaries, and valid patch operations.
Correctness rises from 204/250 (81.6\%) to 230/250 (92.0\%), with
26 recovered outcomes and no losses among previously correct outcomes.
Table~\ref{tab:agentic-regression} gives the two cohorts separately.
This replay starts from contexts that already include earlier processing
and repairs. It tests revised code on retained specifications, without
new model calls or solves; it does not measure fresh construction.

\begin{table}[htbp]
  \caption{Updated-workflow tests. Entries are correct/evaluated
  requests (percentage). A6 scores retained final JSON contexts;
  A7 scores first-turn specification and decision outcomes from new
  responses. Earlier A7 scores use exactly the same 150 selected
  requests. Replay and fresh-response results must not be pooled.}
  \label{tab:agentic-regression}
  \centering
  \small
  \begin{tabularx}{\linewidth}{@{}Xcc@{}}
    \toprule
    \textbf{Test or system} & \textbf{Earlier v1.8}
      & \textbf{Updated v1.9} \\
    \midrule
    \multicolumn{3}{@{}l}{\emph{A6: saved-final-JSON replay, full workflow}} \\
    Main-benchmark contexts & 118/120 (98.3\%) & 119/120 (99.2\%) \\
    Broader-input contexts & 86/130 (66.2\%) & 111/130 (85.4\%) \\
    \midrule
    \multicolumn{3}{@{}l}{\emph{A7: fresh responses, 150 matched requests}} \\
    Model-only & 118/150 (78.7\%) & 127/150 (84.7\%) \\
    Schema-assisted single shot & 112/150 (74.7\%) & 89/150 (59.3\%) \\
    Full workflow & 120/150 (80.0\%) & 121/150 (80.7\%) \\
    \bottomrule
  \end{tabularx}
\end{table}

\begin{figure}[h]
  \centering
  \includegraphics[width=\linewidth]{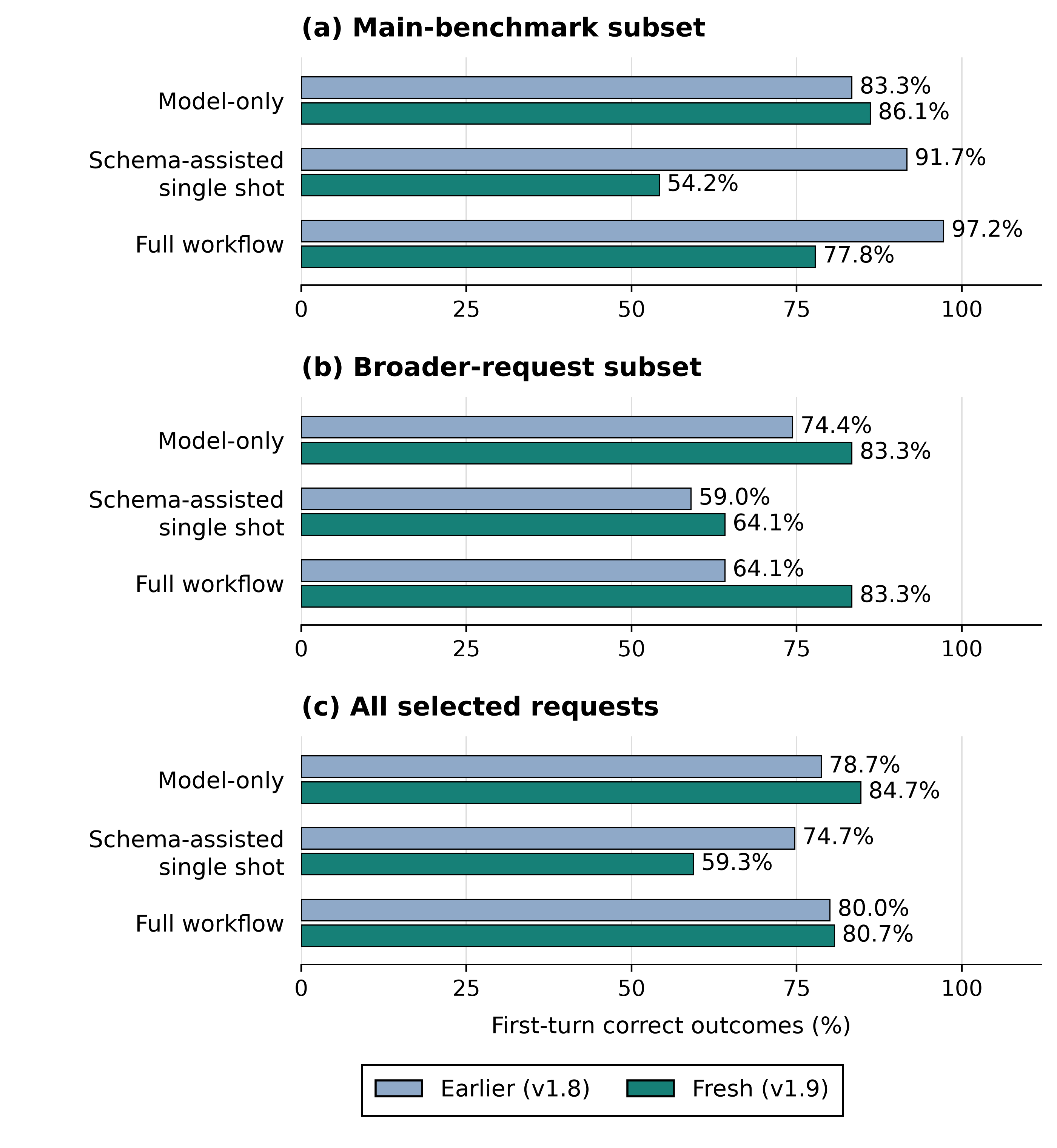}
  \caption{Fresh-response regression A7, with earlier and updated scores
  calculated on the same requests in each panel:
  (a) 72 main-benchmark requests,
  (b) 78 broader-input requests, and
  (c) all 150 selected requests.
  The first two panels partition the third and are not additional tests.
  Values are percentages of correct first-turn specification and decision
  outcomes. Single shot and the full workflow share their initial
  response; model-only uses a separate prompt and draw.
  New responses accompany the workflow update, so changes cannot be
  attributed to the version alone. All panels use the same scale.}
  \label{fig:main-agentic-regression}
\end{figure}

A7 obtains fresh responses to 150 known requests: 72 from A1 and 78
from A4. Overall full-workflow correctness is 121/150 (80.7\%),
compared with 120/150 (80.0\%) in the retained earlier results.
The broader-request subset improves from 50/78 (64.1\%) to
65/78 (83.3\%), while the main-benchmark subset changes from
70/72 (97.2\%) to 56/72 (77.8\%)
(Figure~\ref{fig:main-agentic-regression}).
The updated full workflow returns 70/97 correct complete specifications
(72.2\%) and makes no false-ready decisions on the 53 non-ready requests.
Audit replay reproduces all 150 retained session contexts, including
40 accepted patches and 24 rejected proposals that leave the context
unchanged. This checks state handling, while the outcome scores assess
scientific correctness. Workflow changes and new model responses occur
together, so A7 is a regression test on known requests, not an isolated
version comparison or evidence of generalization to unseen requests.

\section{Reproducibility}
\label{app:reproducibility}

This section summarizes dataset construction and execution settings.
Detailed agentic results are reported in
Appendix~\ref{app:additional-agentic}.

\subsection{Dataset construction and study scope}
\label{app:dataset-construction}

The main study contains 120 code-authored requests, and the broader-input
study adds 130 requests. Each request has expected scientific fields
and a decision to proceed, clarify, or reject. Related templates vary
geometry, coefficients, boundary conditions, and time settings.
Natural and schema-explicit wording are categories within these
collections. Shared templates limit how far the results can be
generalized to unrestricted engineering requests.

The broader collection was balanced by difficulty before model calls,
using geometry, boundary data, nonlinearity, time dependence, and
expressions. The v1.9 regression study uses fresh responses to
150 selected requests from the existing collections. Repeated requests
are not counted as new independent cases. The gallery starts from
retained JSON specifications, and numerical verification uses prescribed
problems; both are evaluated separately from user-to-JSON accuracy.

\subsection{Execution settings}
\label{app:repro-environments}

The model snapshot, generation settings, schema, and patch limits are
specified in Section~\ref{sec:setup} and
Appendix~\ref{app:json-protocol}. Single shot and the full workflow
share the initial model response; model-only uses its own prompt and
response. Original studies use workflow v1.8, while the latest regression
and live workflow tests use v1.9. Their results retain separate
denominators. A fixed request-ordering seed does not make model
responses deterministic.

The cloud verification campaign uses 32 logical CPUs and approximately
252\,GiB of visible RAM. Numerical workers use one library thread;
most studies use one MPI process, while the parallel-agreement study
uses 4, 8, 16, and 32 processes. The additional verification repetitions
and latest live workflow tests run locally on WSL with one MPI process
and one library thread. Case-specific software, meshes, time steps,
and solver tolerances are given in Appendix~\ref{app:numerical};
live execution limits are given in Appendix~\ref{app:e2e}.

The implementation and run instructions will accompany the repository
release. Repeated model calls may produce different responses;
numerical reproduction requires the stated discretization,
solver settings, and acceptance rules.


\begingroup
\raggedbottom

\providecommand{\gallerysingle}[3]{}
\providecommand{\gallerypair}[4]{}

\renewcommand{\gallerysingle}[3]{%
  \par\addvspace{6pt}%
  \noindent
  \begin{minipage}{\linewidth}
    \centering
    \refstepcounter{figure}\label{#3}%
    \includegraphics[width=\linewidth]{solution_gallery/#1}%
  \end{minipage}%
  \par
}

\renewcommand{\gallerypair}[4]{%
  \gallerysingle{#1}{}{#4}%
  \par\addvspace{6pt}%
  \noindent
  \begin{minipage}{\linewidth}
    \centering
    \includegraphics[width=\linewidth]{solution_gallery/#2}%
  \end{minipage}%
  \par
}

\section{Solution Gallery}
\label{app:gallery}

These solutions were generated from retained JSON specifications using
an automated pipeline: JSON $\rightarrow$ geometry $\rightarrow$ mesh
$\rightarrow$ solve $\rightarrow$ results. Each specification defines
the problem and numerical settings; the workflow constructs the domain,
generates the mesh, solves with FEniCSx/PETSc, and exports the fields.

Cases are grouped by dimension. Unsteady snapshots and additional
flow or elasticity fields follow their corresponding cases.
Deformation magnification affects the display only.
The gallery retains verified, failed, and preliminary numerical
qualifications; execution completion does not establish numerical accuracy.
Numerical verification is assessed separately in
Appendix~\ref{app:numerical}.


\subsection{One-dimensional cases}

\gallerypair{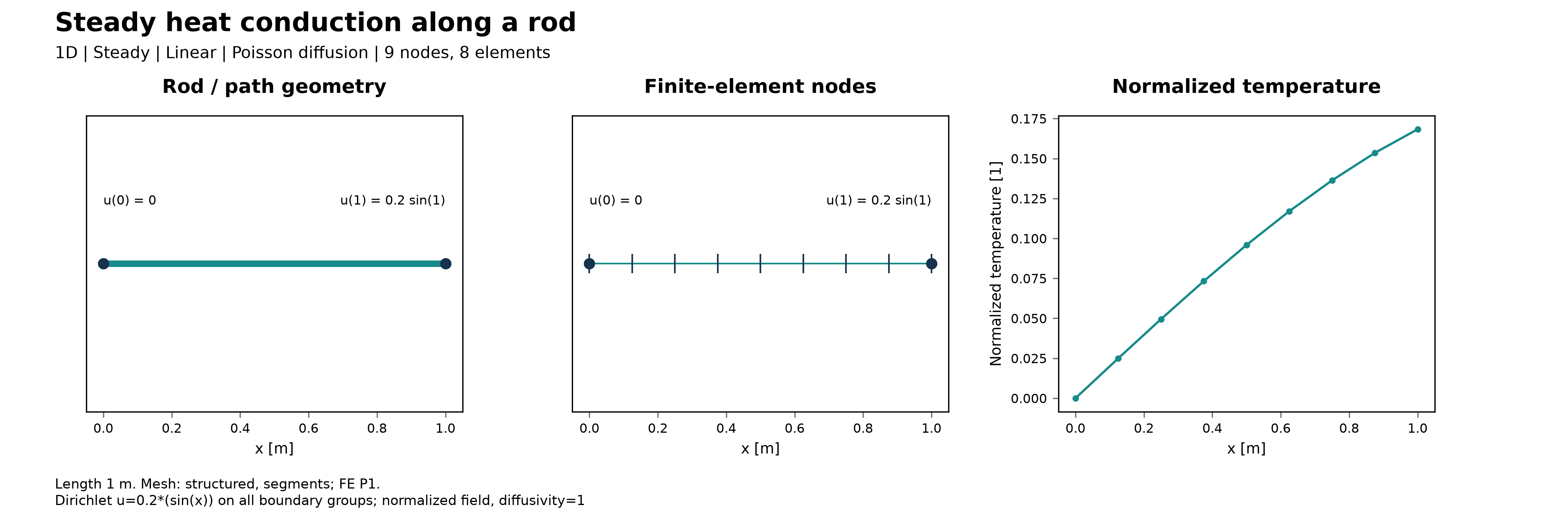}
  {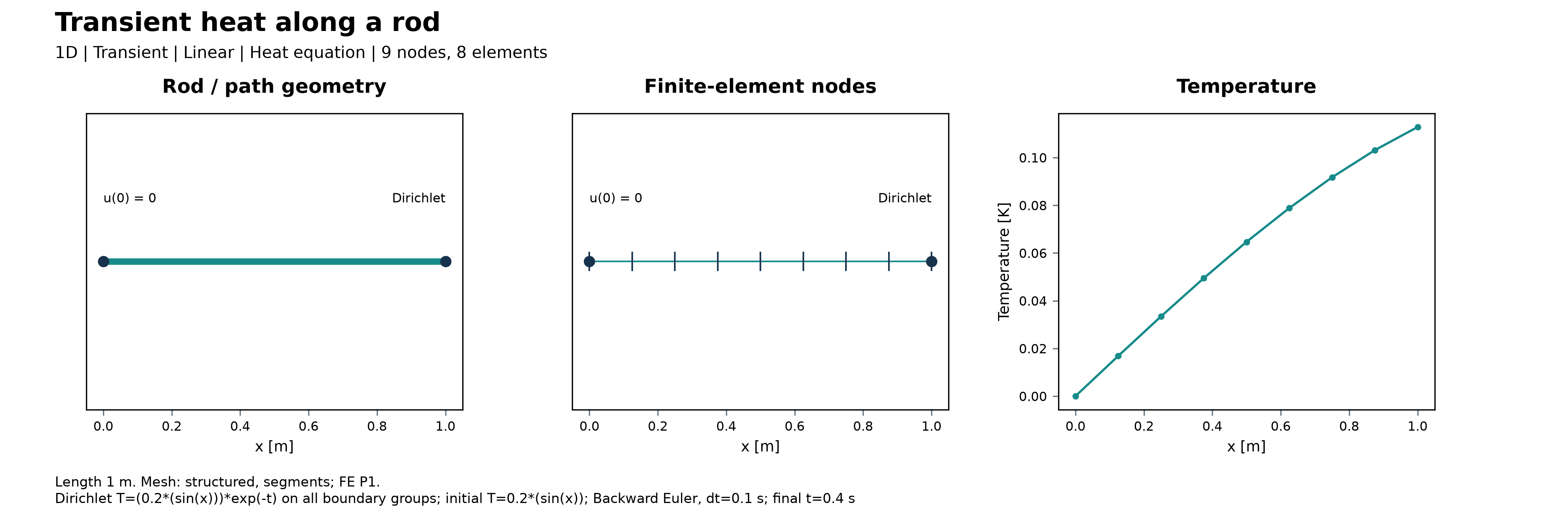}
  {}
  {fig:gallery-1d}

\gallerysingle{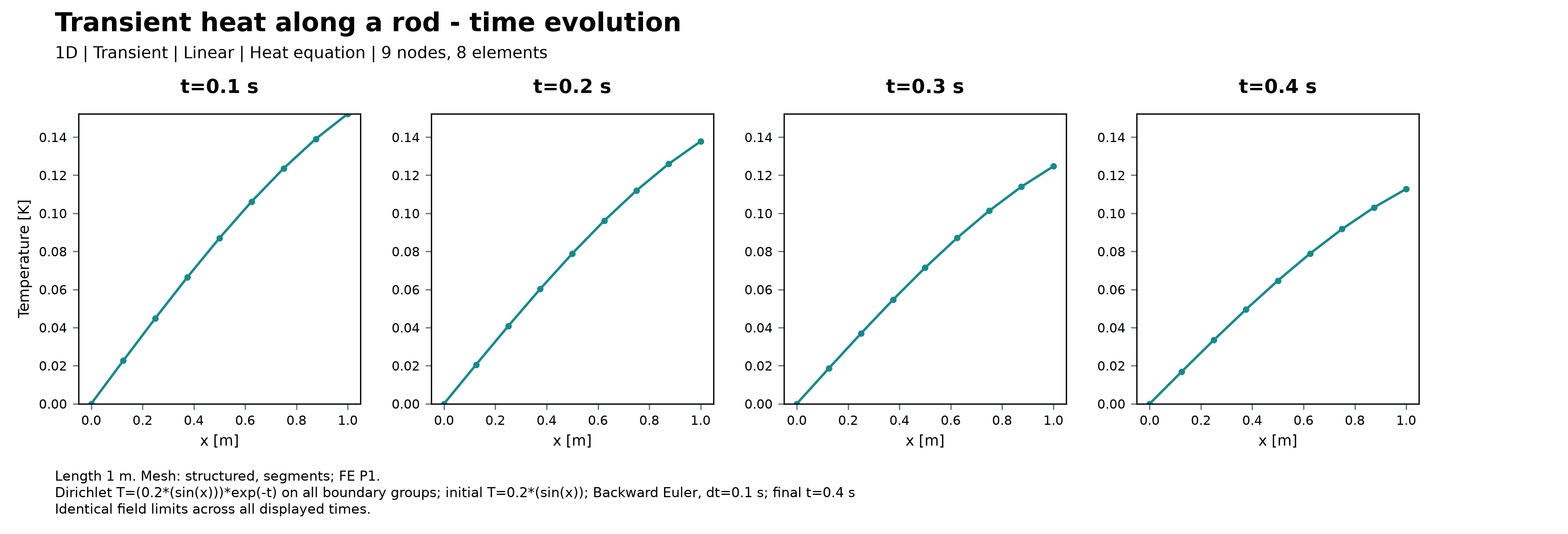}
  {}
  {fig:gallery-unsteady}

\gallerysingle{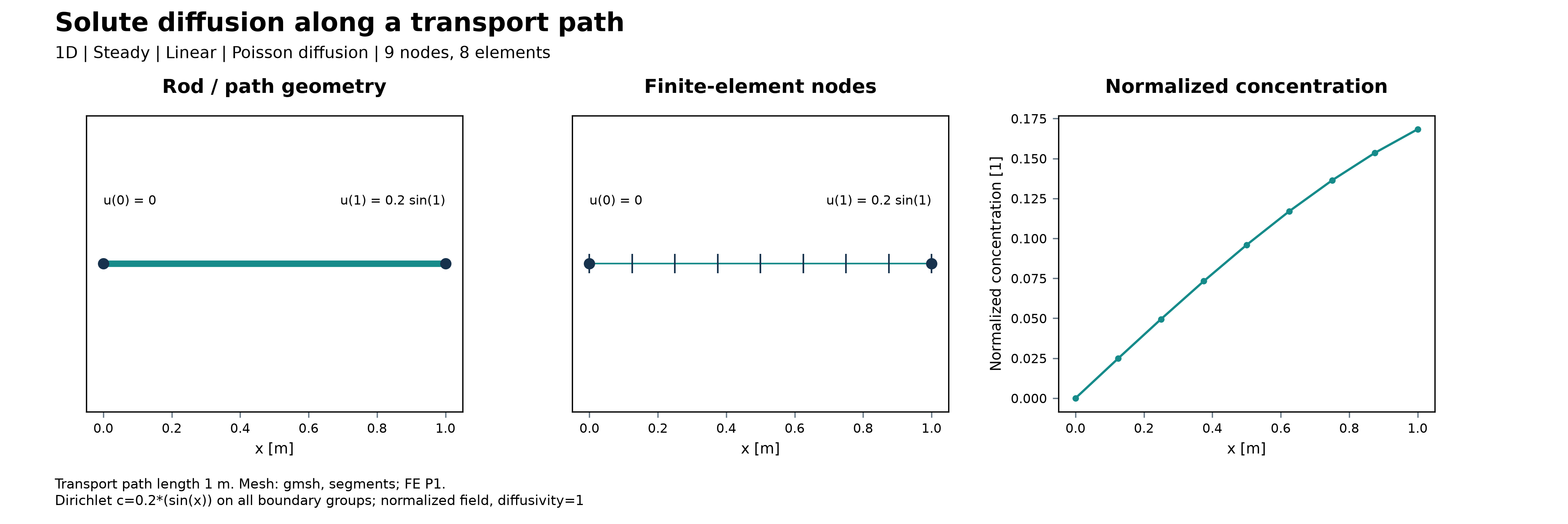}
  {}
  {fig:gallery-polyline}


\subsection{Two-dimensional cases}

\gallerypair{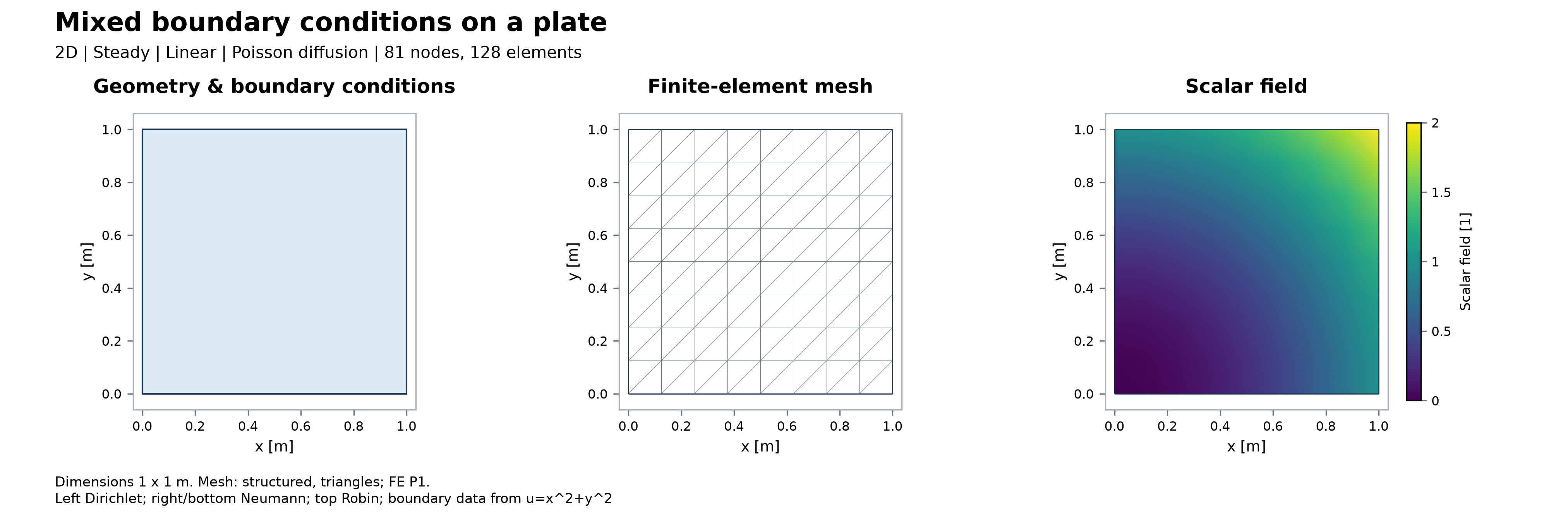}
  {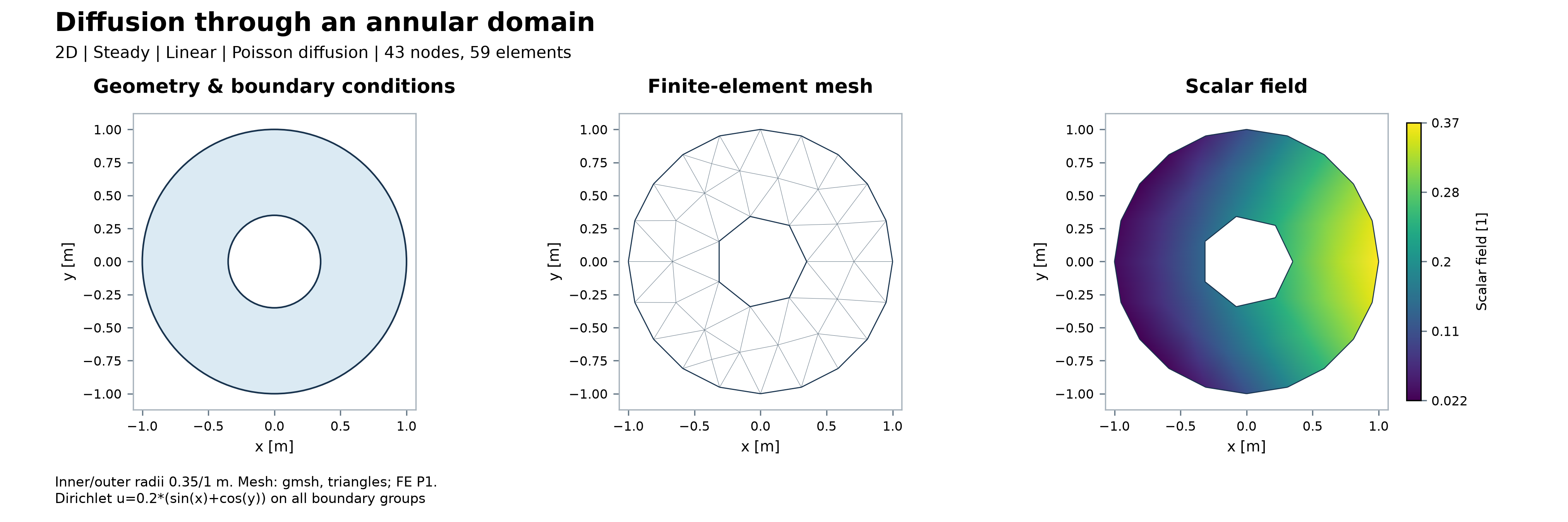}
  {}
  {fig:gallery-2d}

\gallerypair{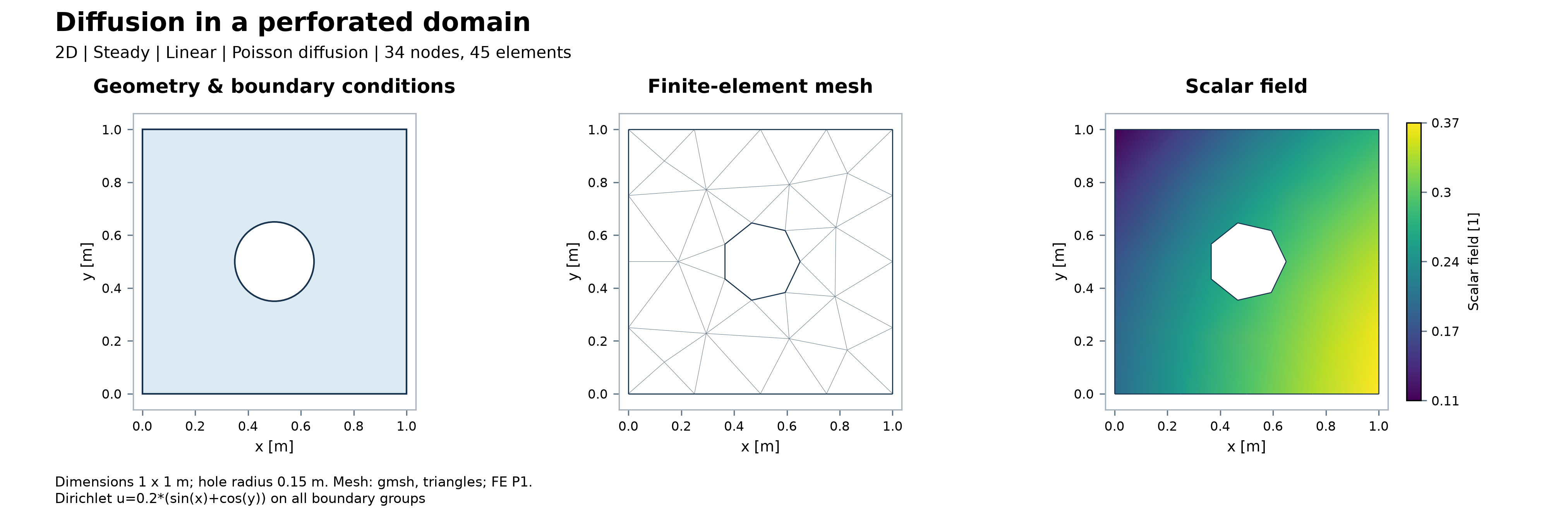}
  {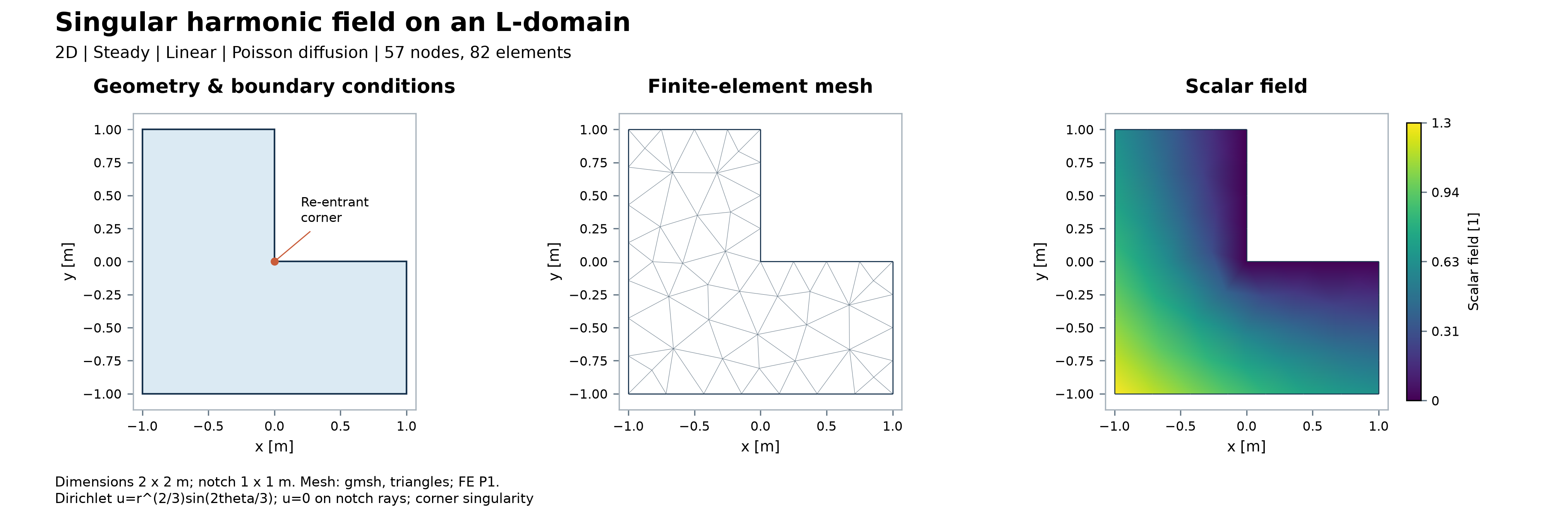}
  {}
  {fig:gallery-2d-perforated-harmonic}

\gallerypair{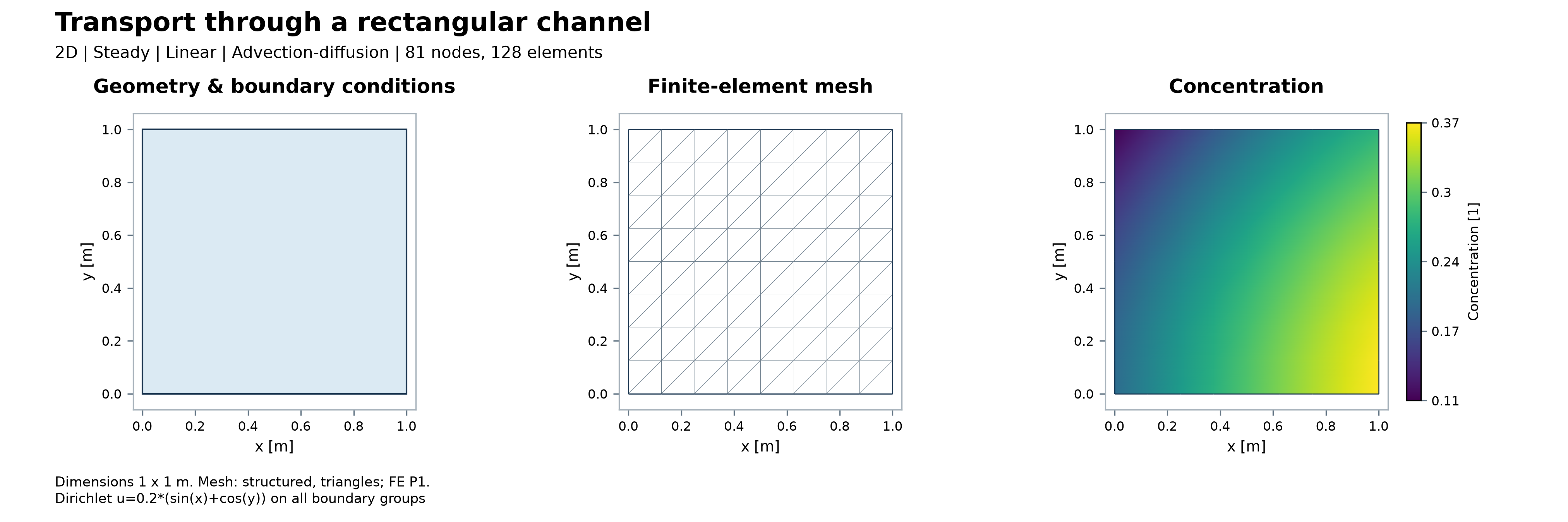}
  {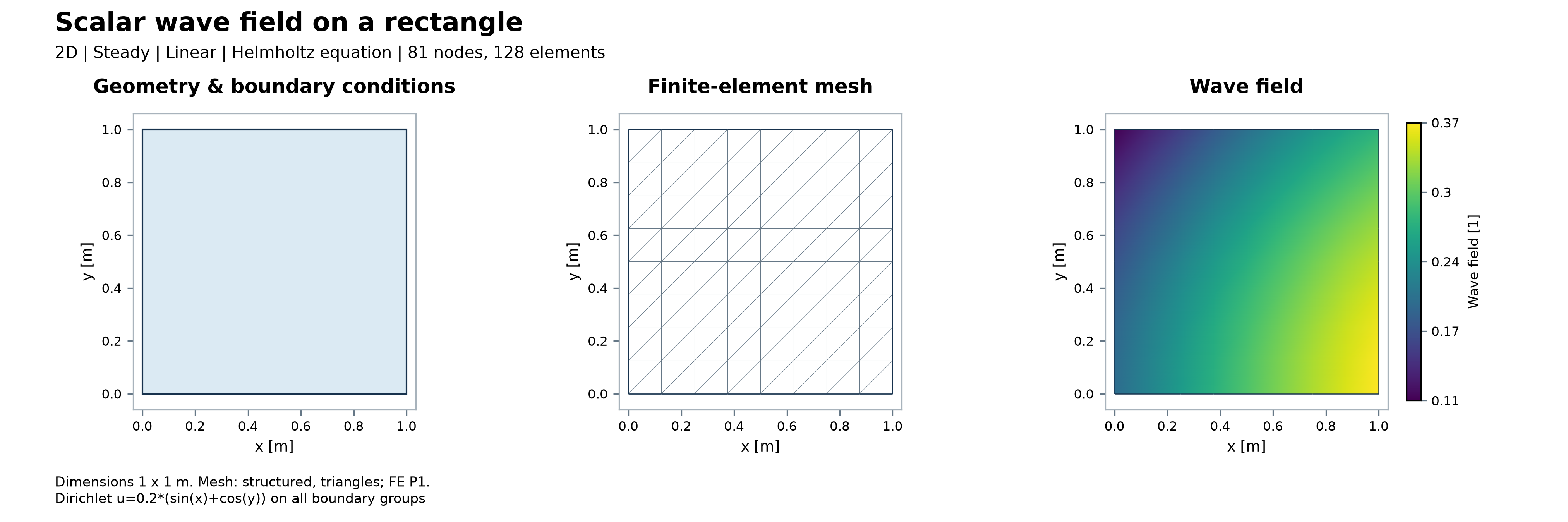}
  {}
  {fig:gallery-2d-transport-helmholtz}

\gallerypair{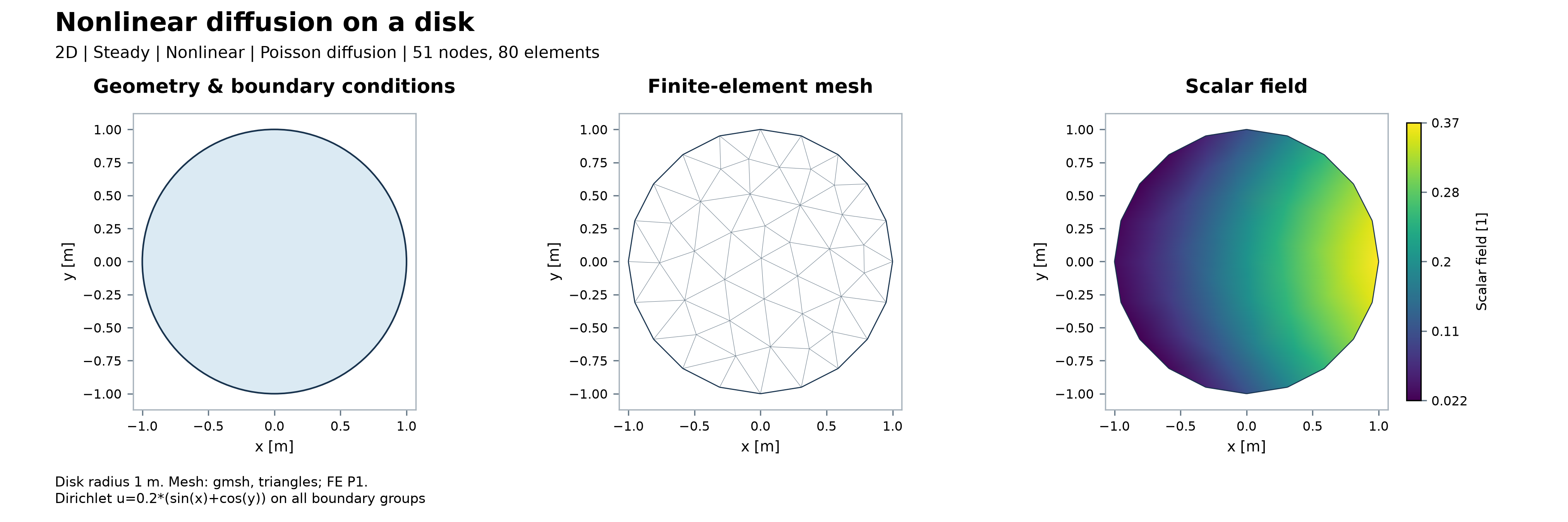}
  {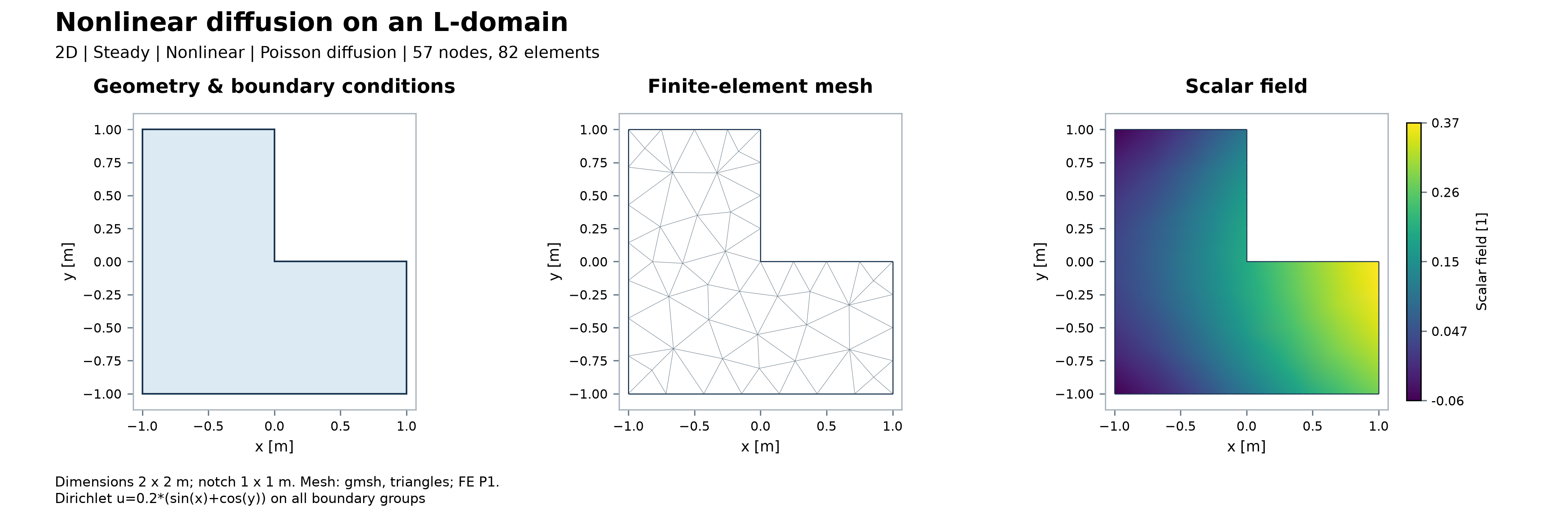}
  {}
  {fig:gallery-2d-nonlinear}

\gallerypair{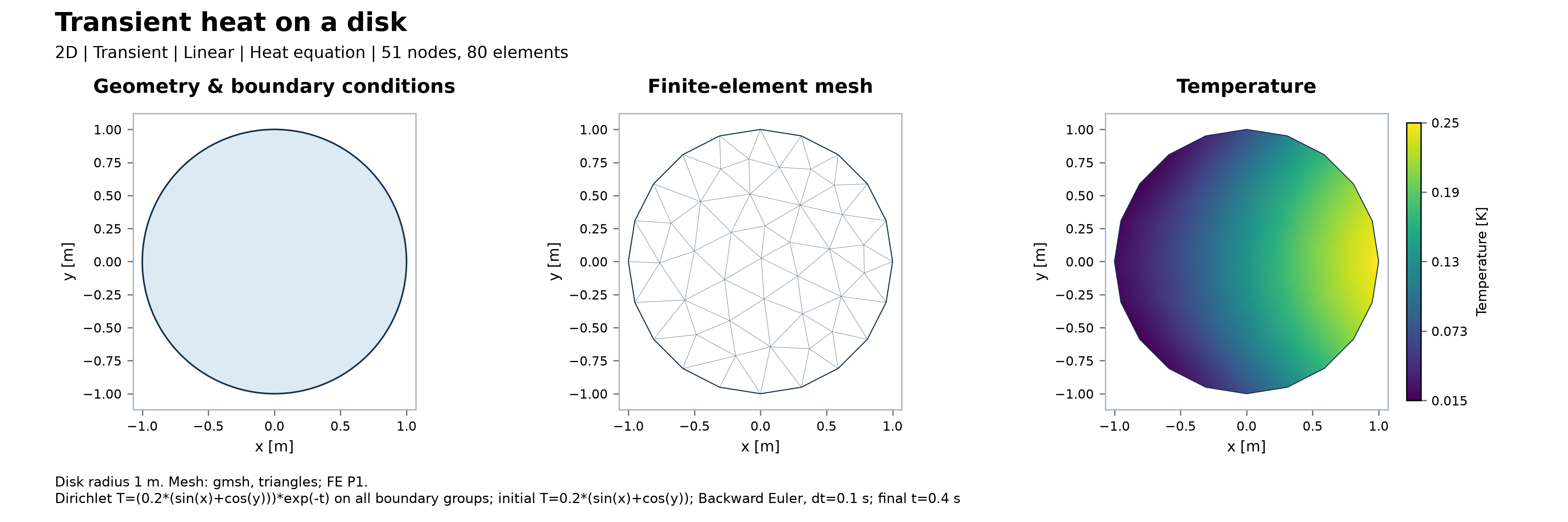}
  {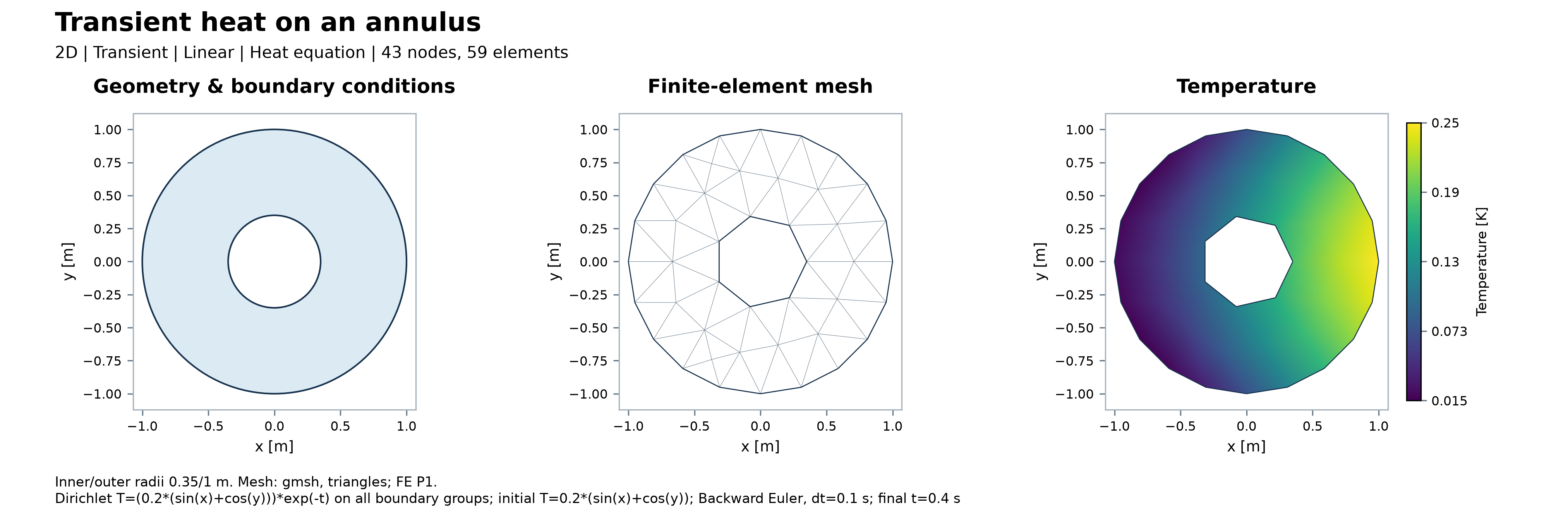}
  {}
  {fig:gallery-2d-disk-annular-heat}

\gallerypair{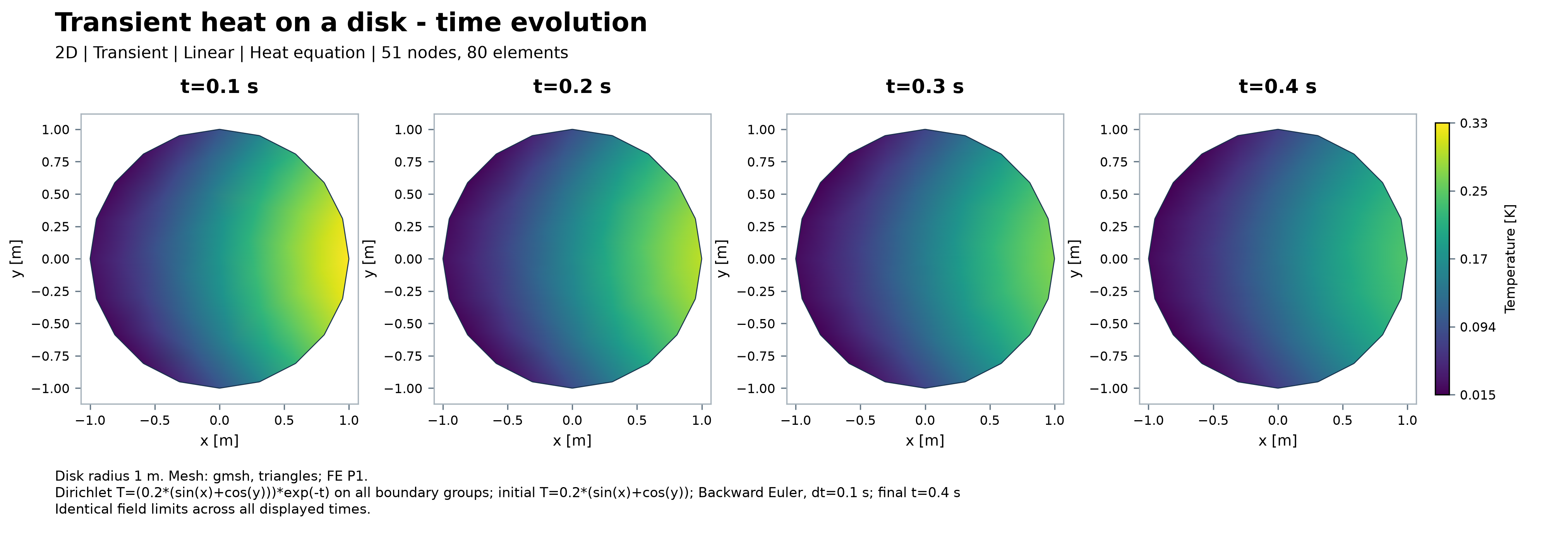}
  {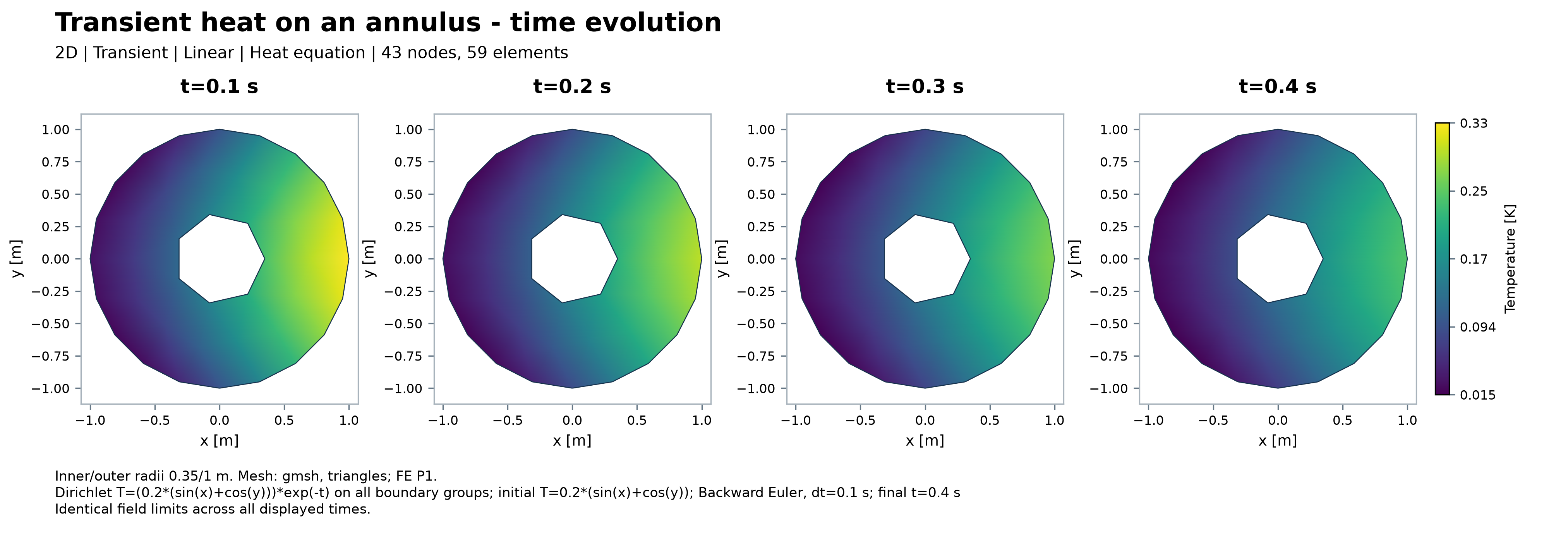}
  {}
  {fig:gallery-unsteady-disk-annulus}

\gallerypair{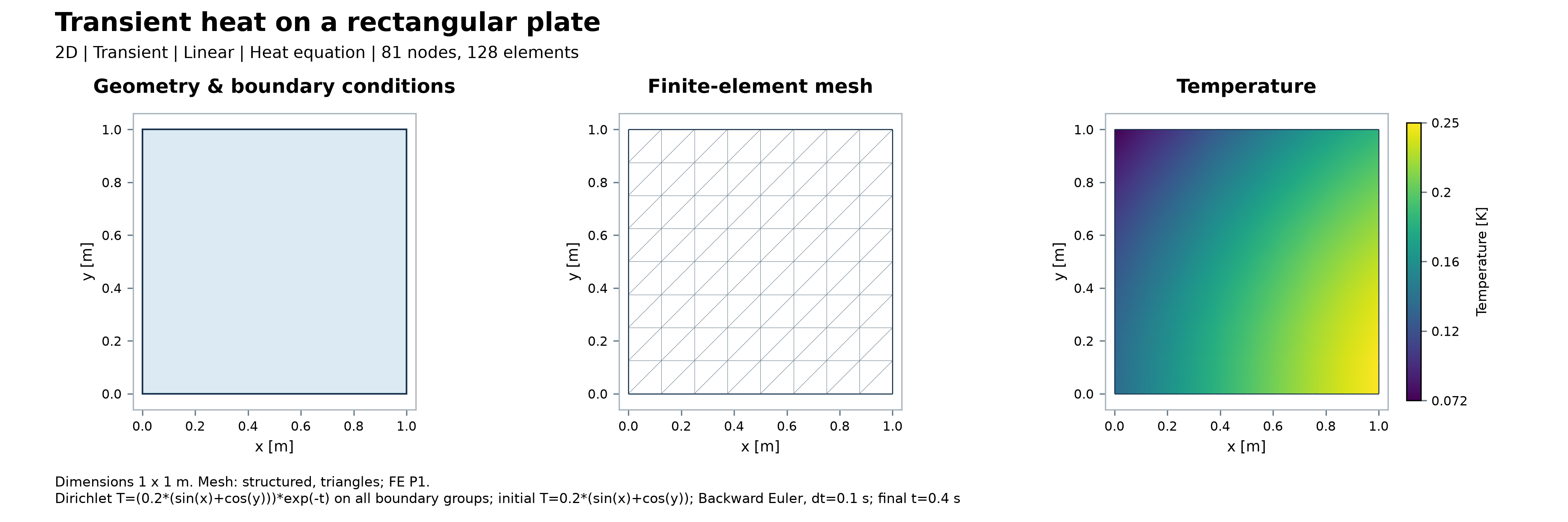}
  {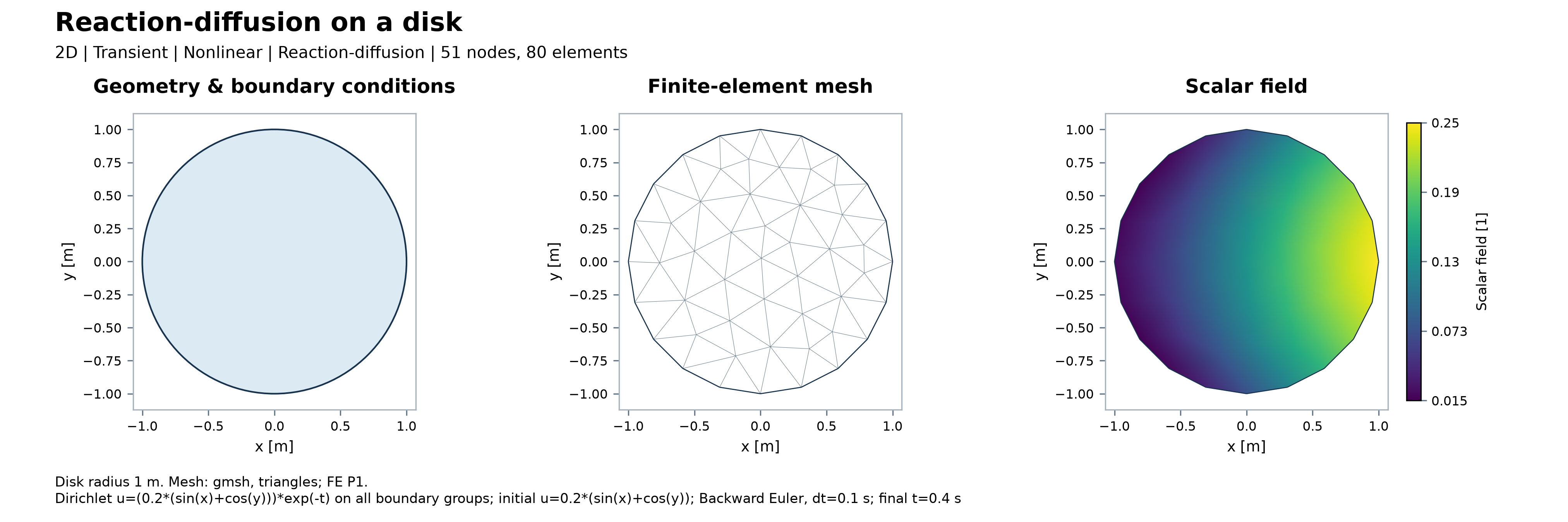}
  {}
  {fig:gallery-2d-heat-reaction}

\gallerypair{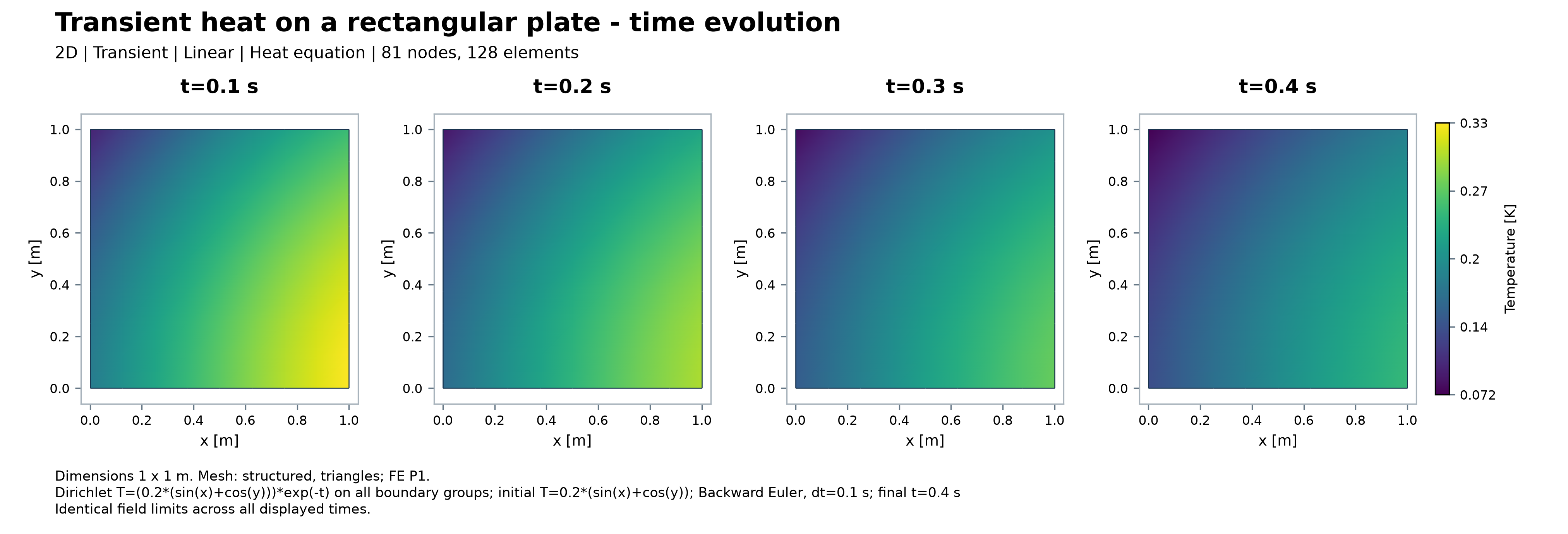}
  {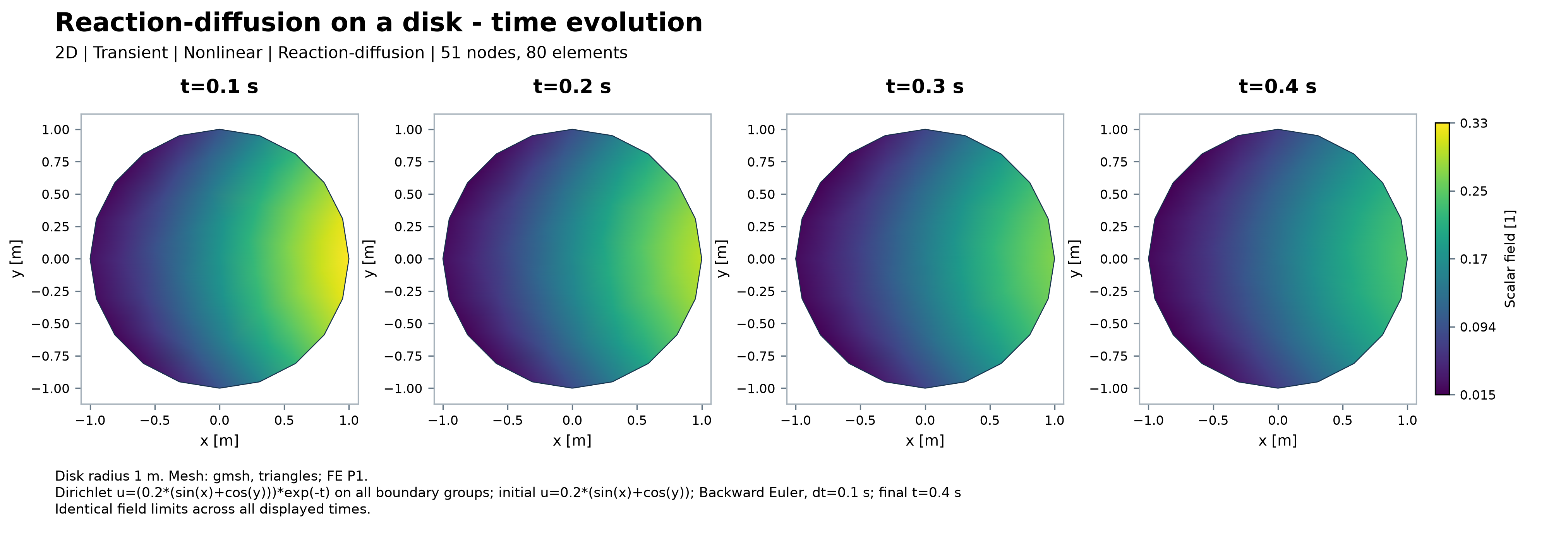}
  {}
  {fig:gallery-unsteady-rectangle-reaction}

\gallerypair{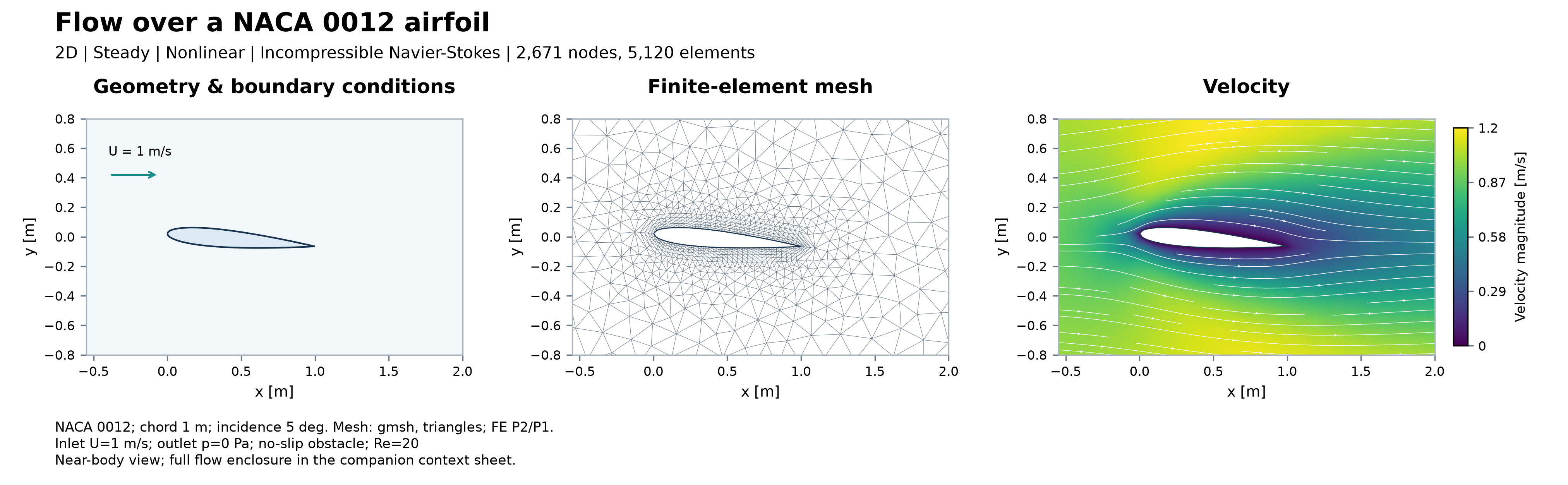}
  {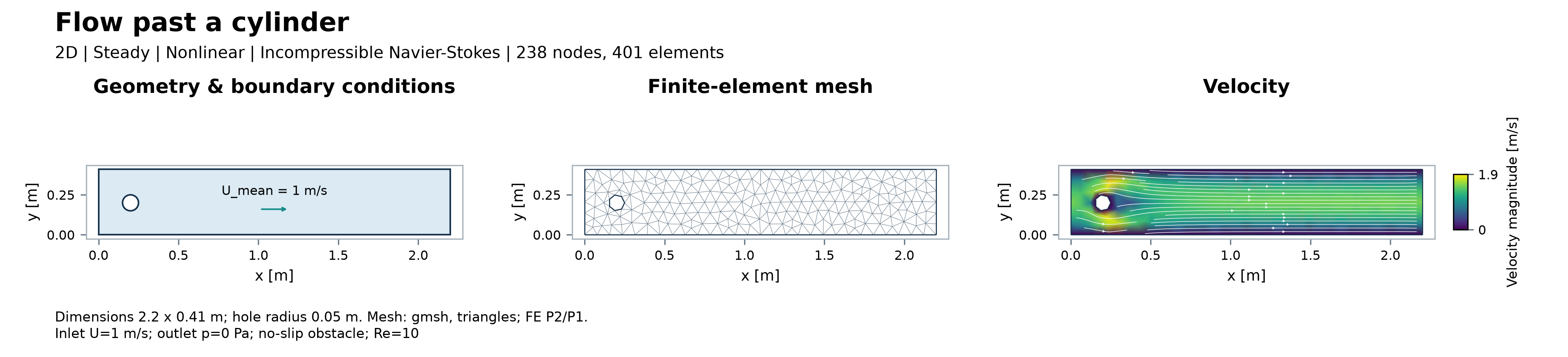}
  {}
  {fig:gallery-2d-flow}

\gallerypair{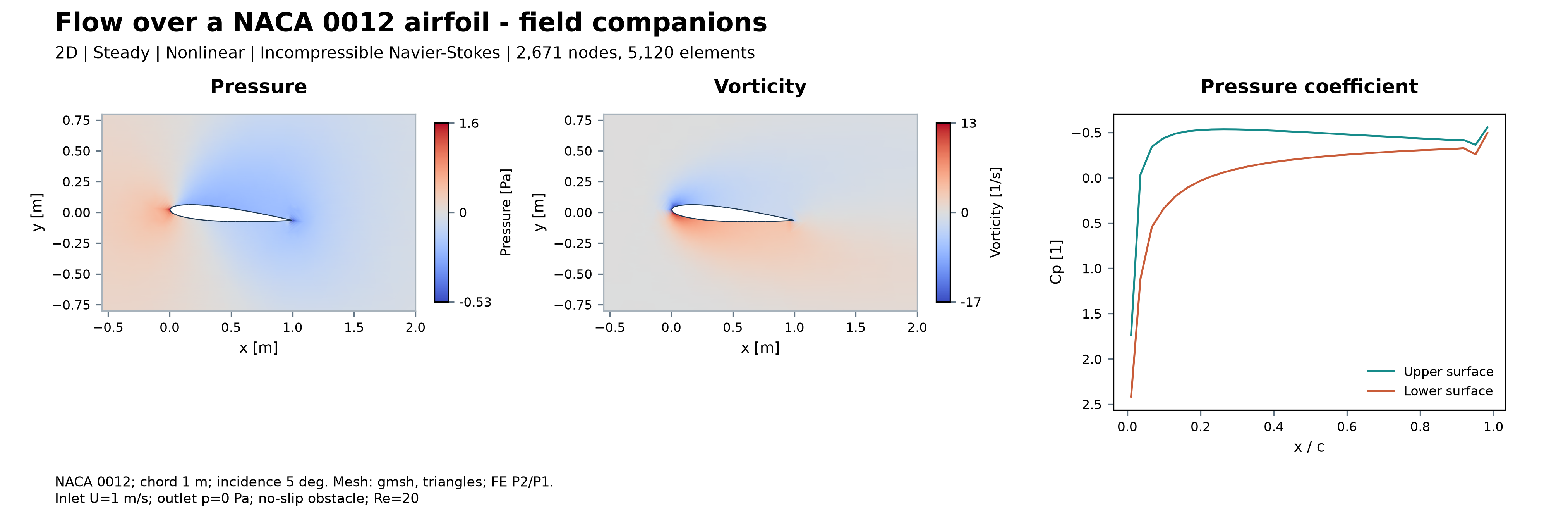}
  {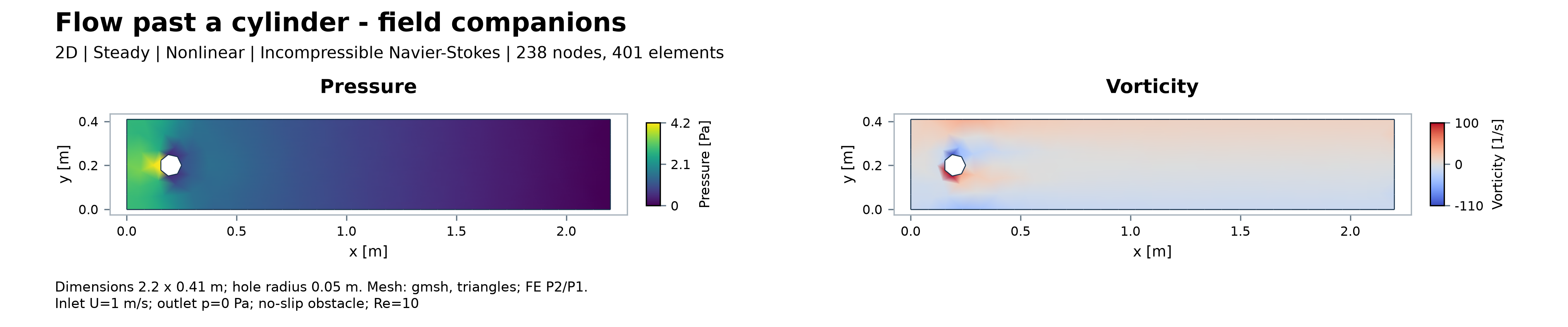}
  {}
  {fig:gallery-companion-2d-flow}

\gallerypair{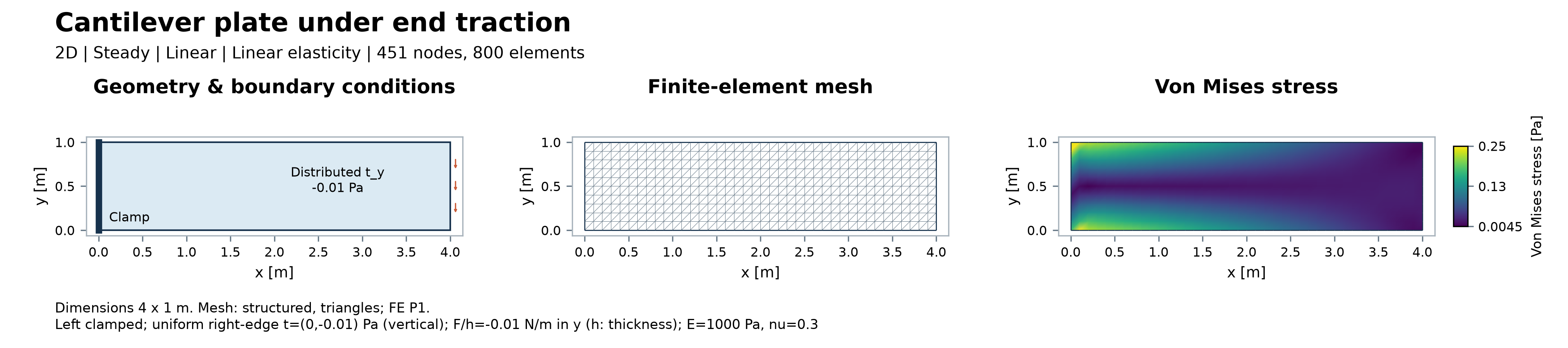}
  {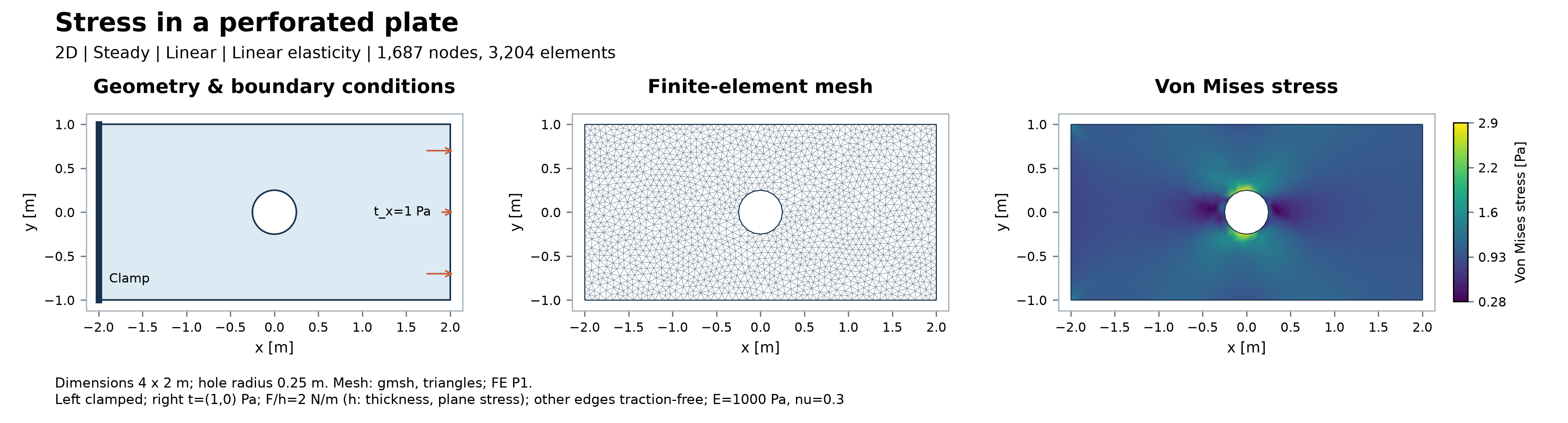}
  {}
  {fig:gallery-2d-elasticity}

\gallerypair{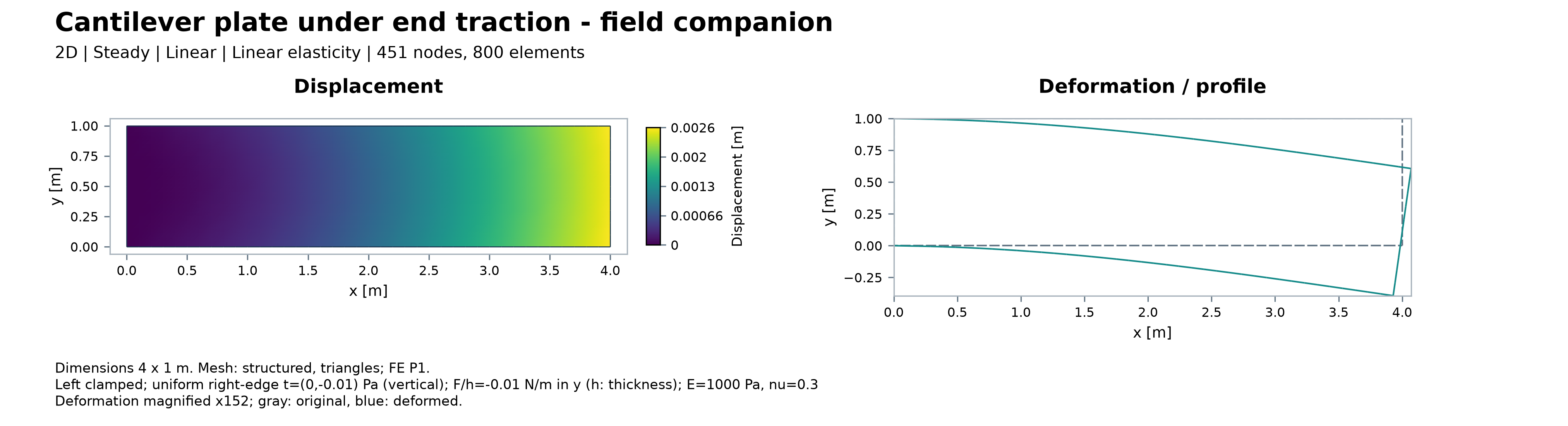}
  {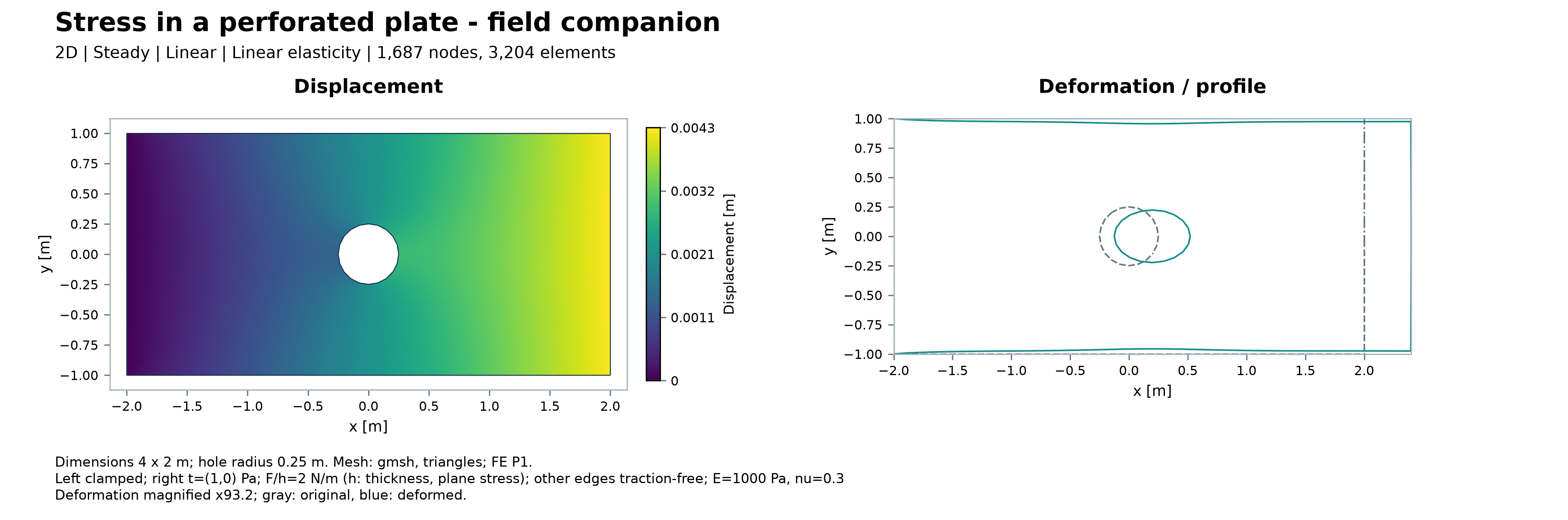}
  {}
  {fig:gallery-companion-2d-elasticity}


\subsection{Three-dimensional cases and complex geometries}

\gallerypair{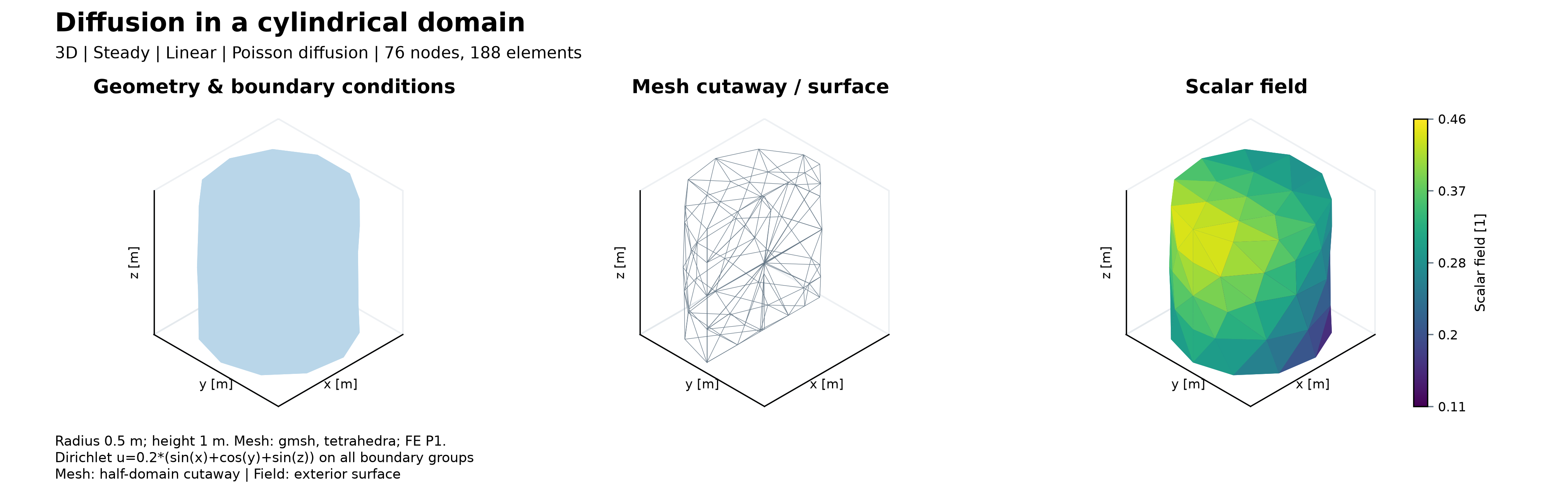}
  {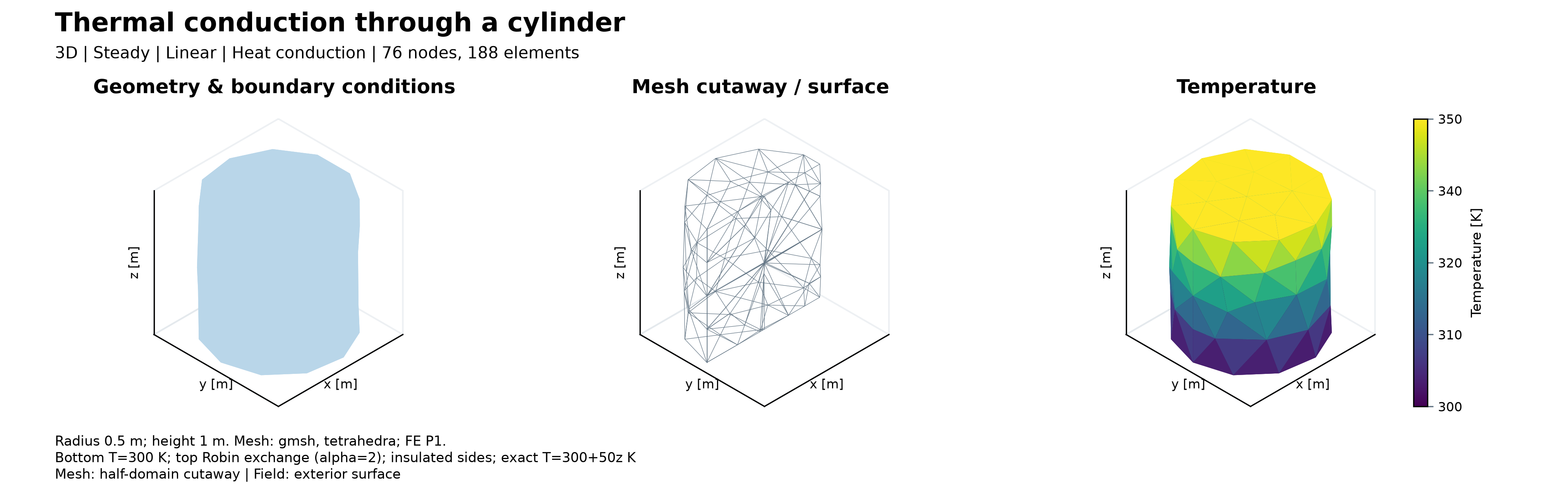}
  {}
  {fig:gallery-3d}

\gallerypair{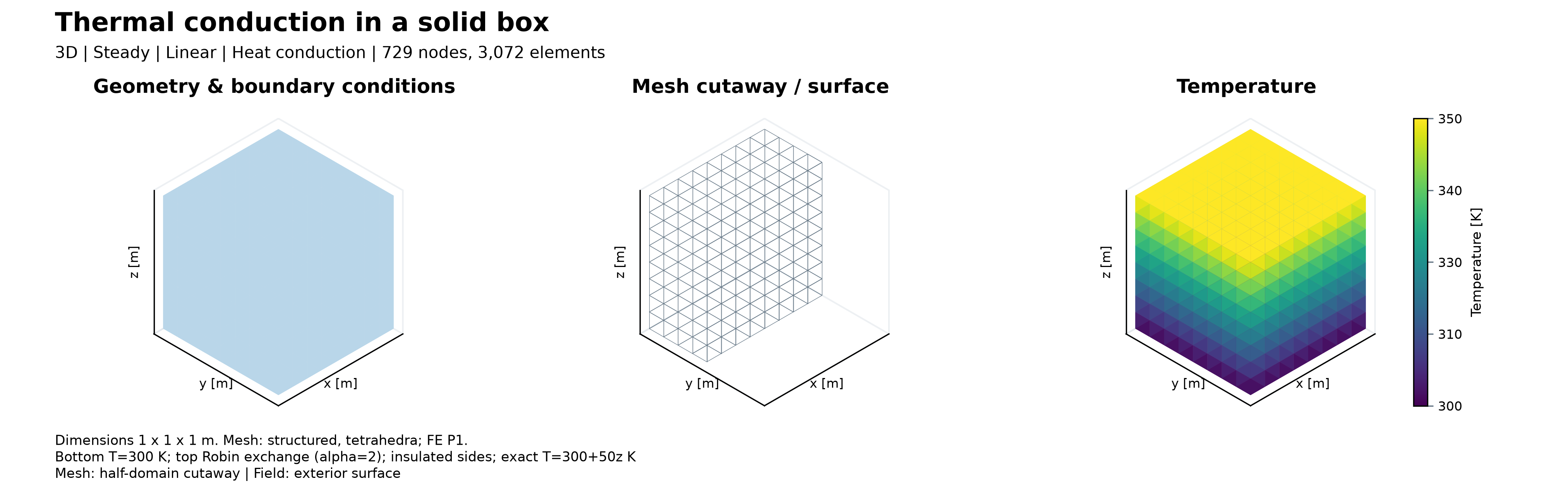}
  {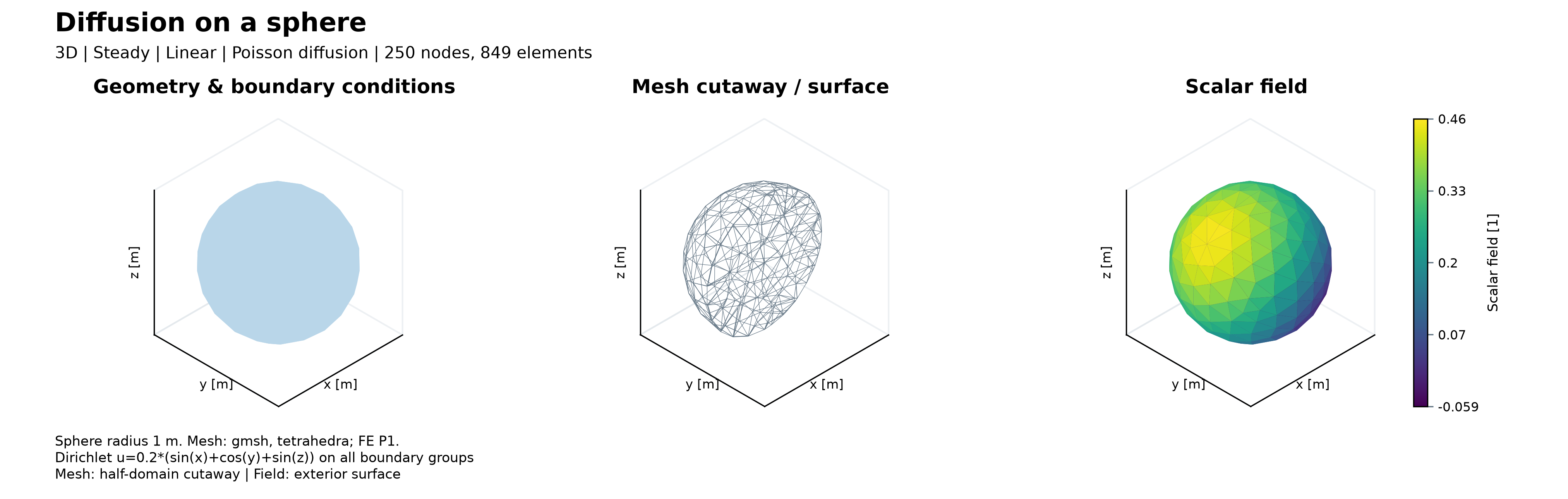}
  {}
  {fig:gallery-3d-box-sphere}

\gallerysingle{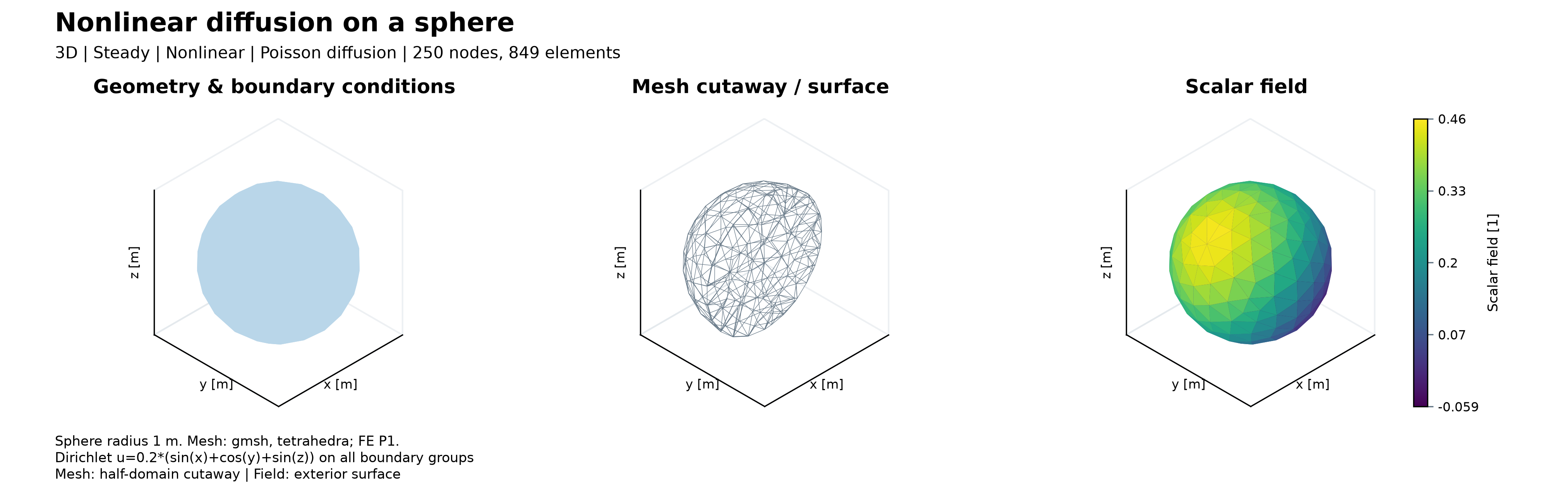}
  {}
  {fig:gallery-3d-nonlinear-sphere}

\gallerypair{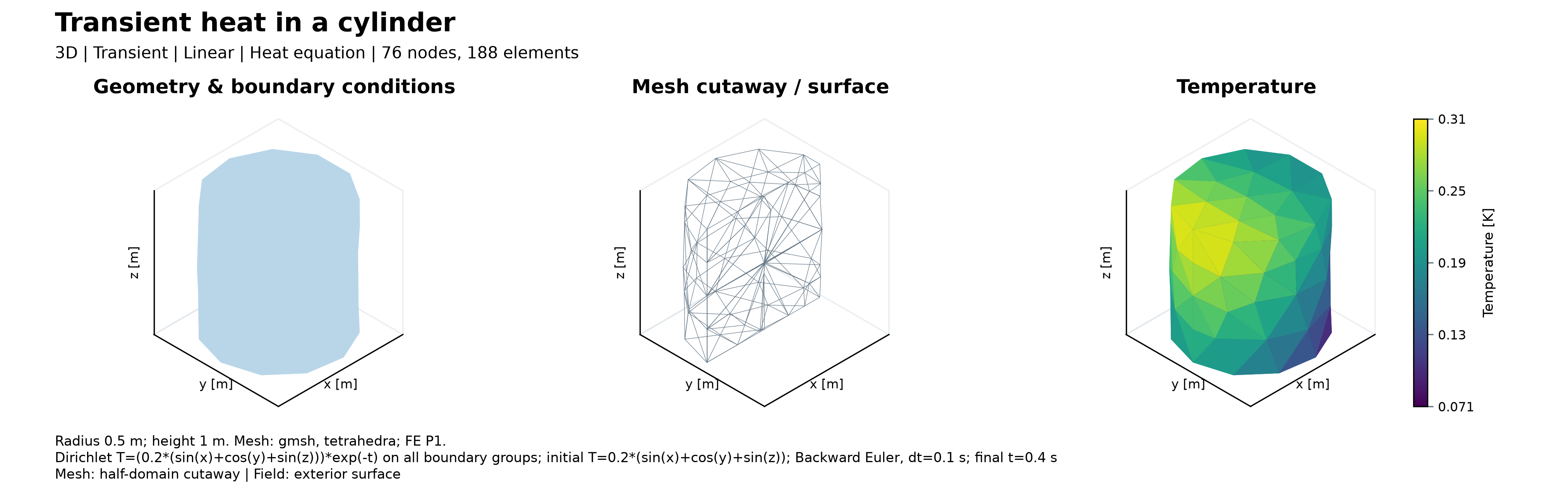}
  {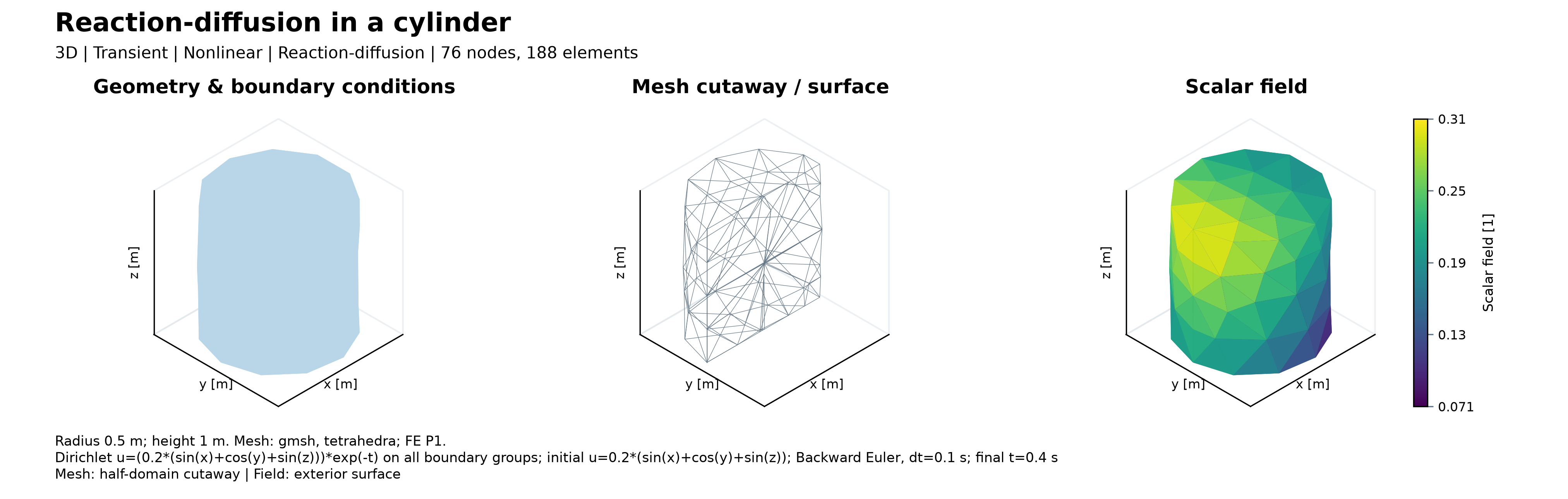}
  {}
  {fig:gallery-3d-unsteady}

\gallerypair{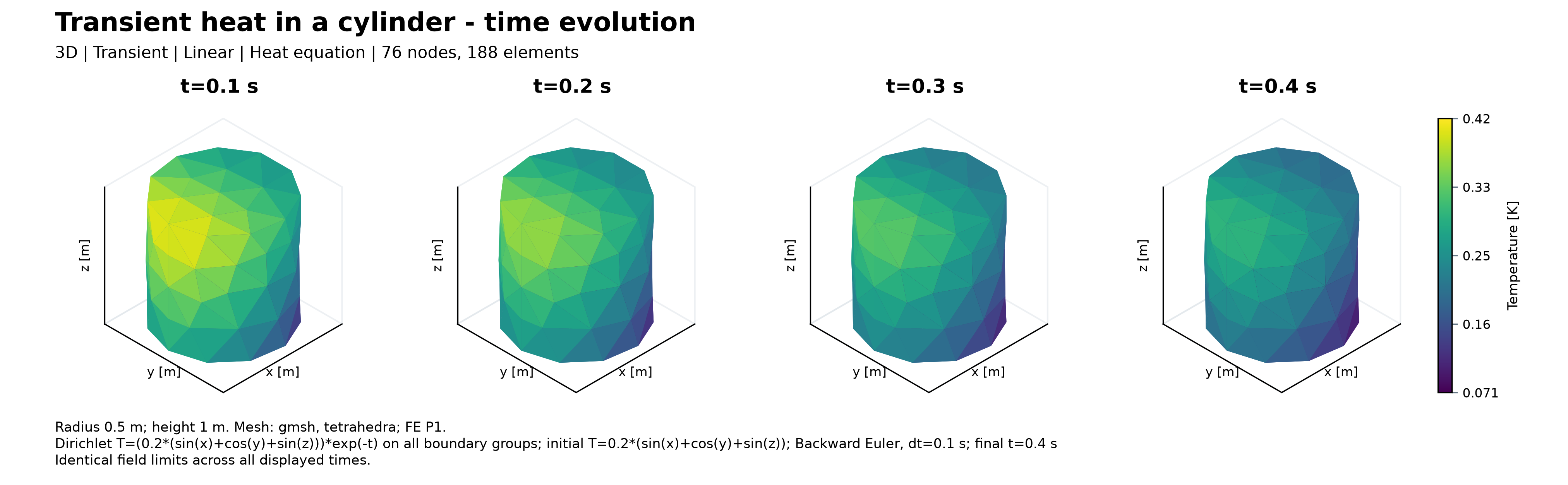}
  {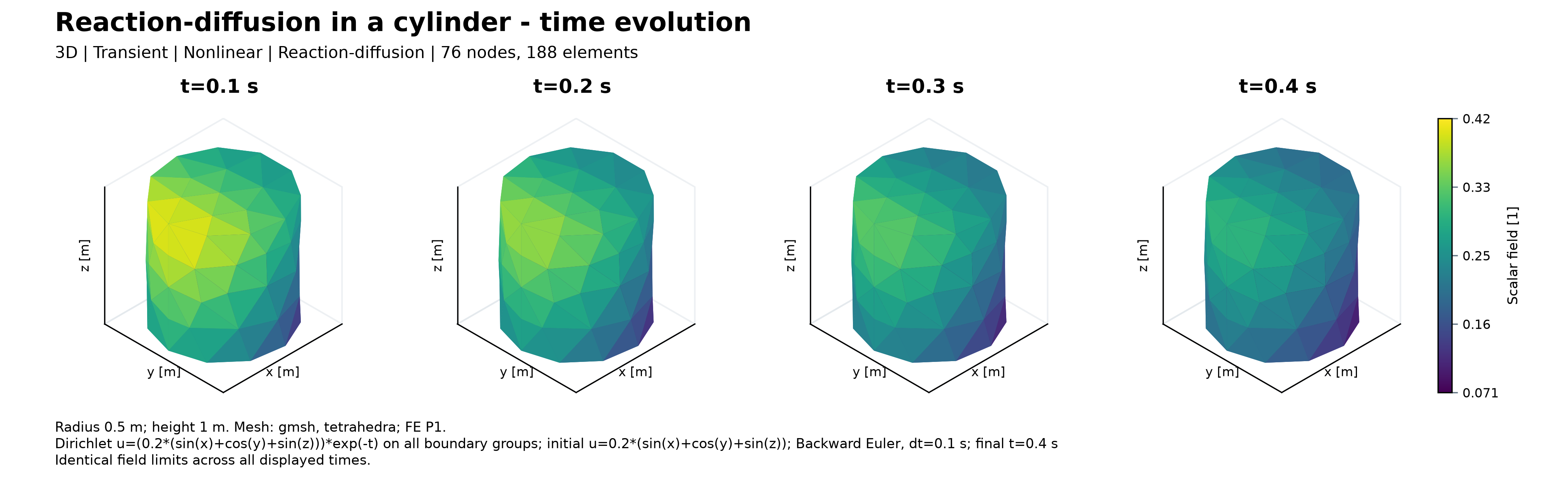}
  {}
  {fig:gallery-unsteady-3d}

\gallerypair{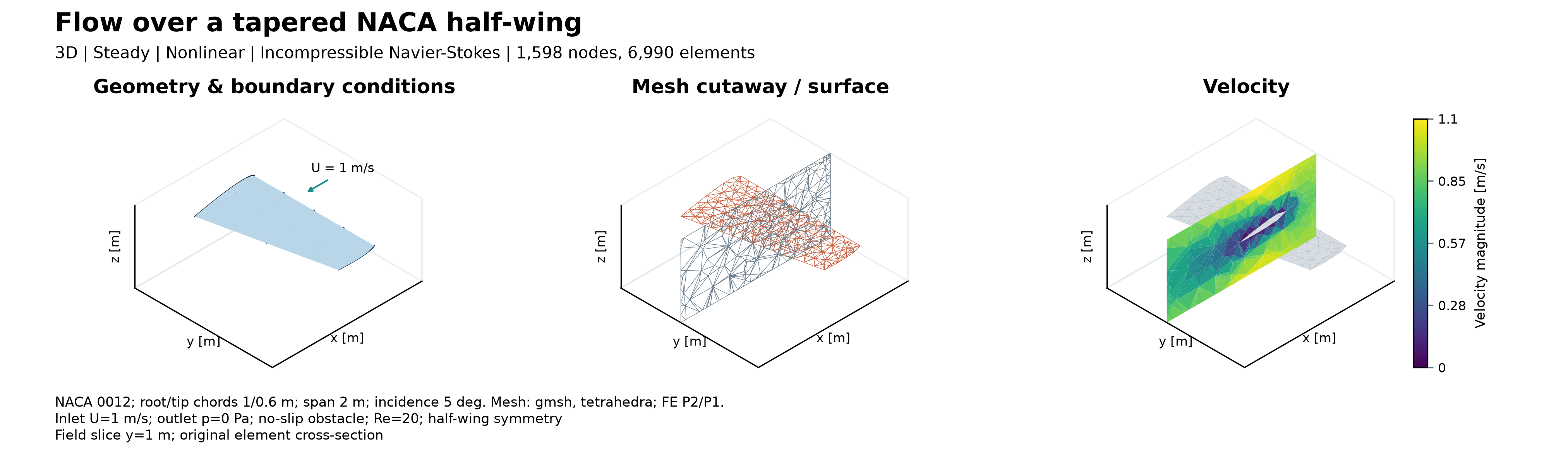}
  {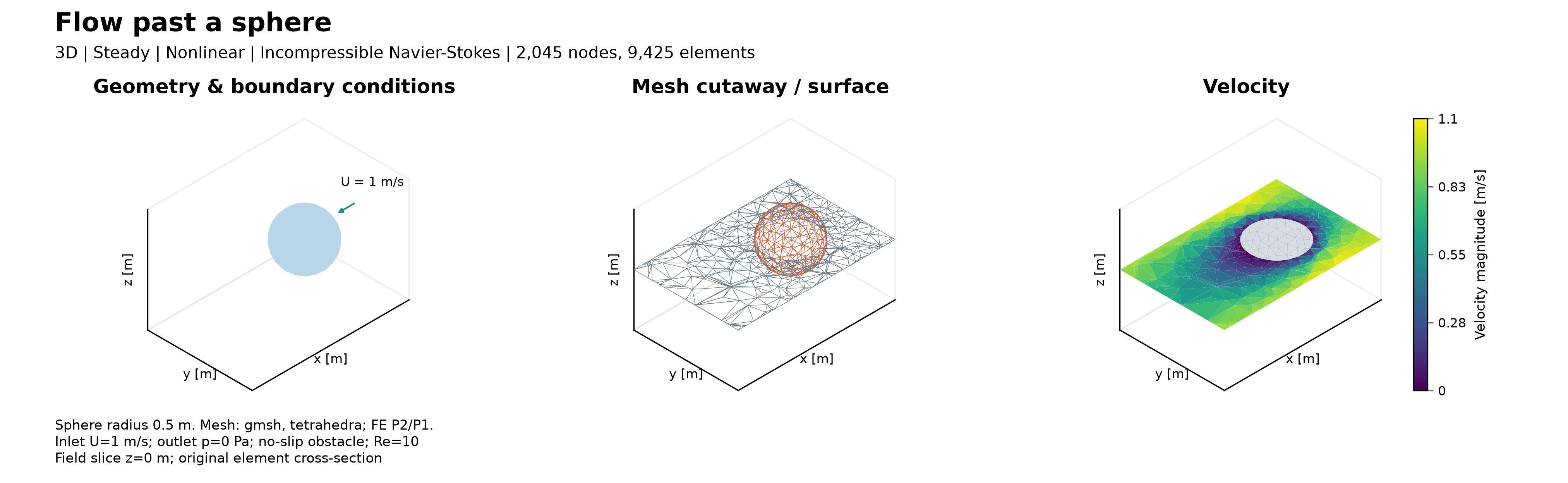}
  {}
  {fig:gallery-3d-flow}

\gallerypair{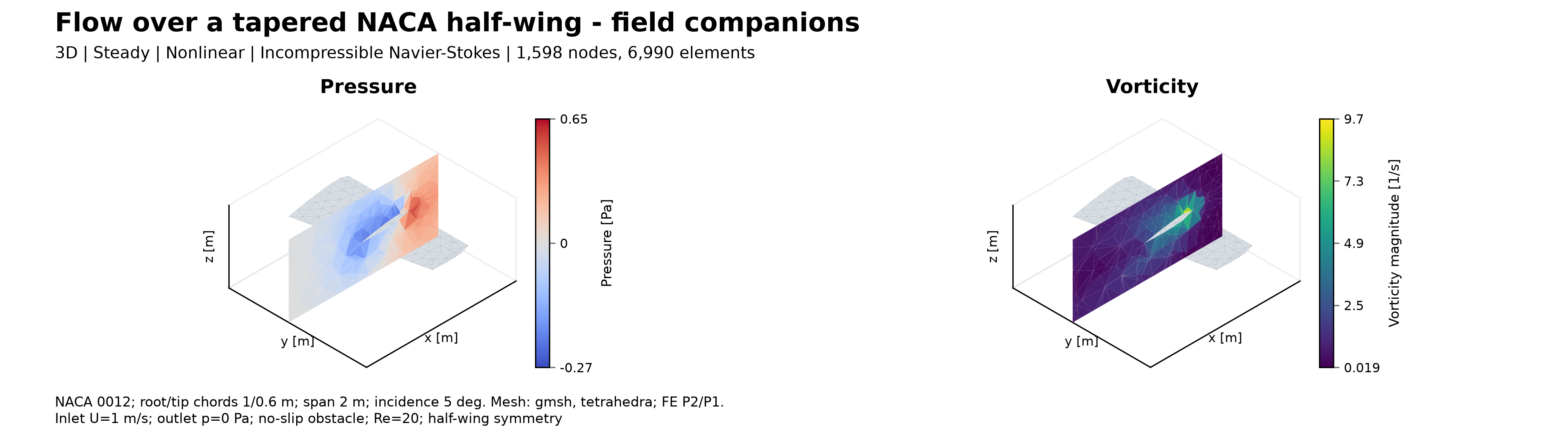}
  {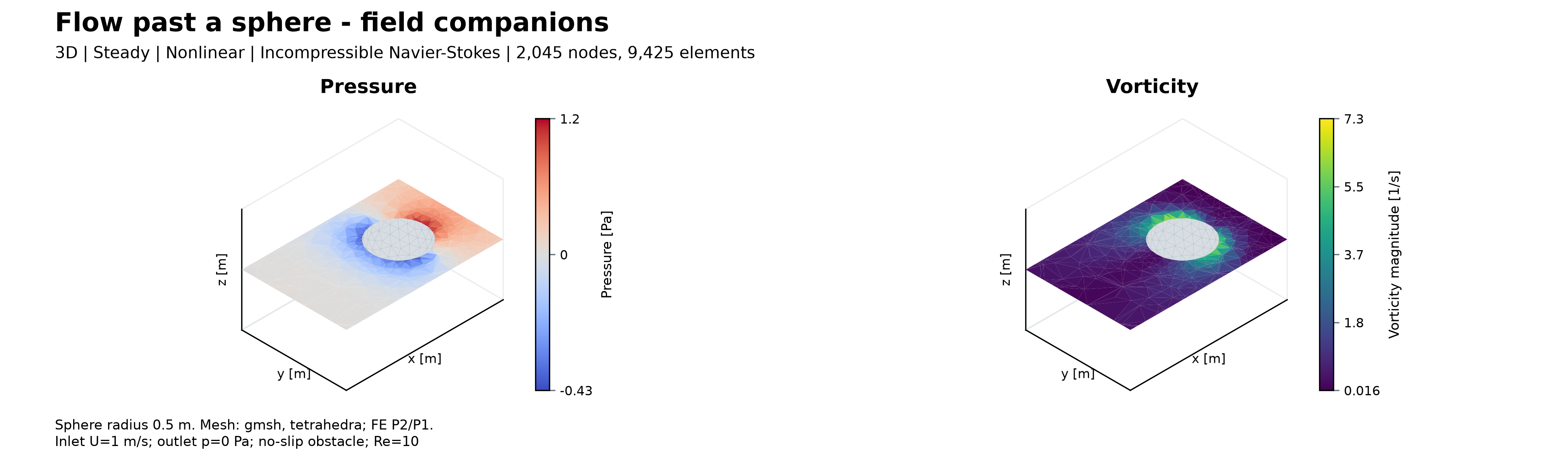}
  {}
  {fig:gallery-companion-3d-flow}

\gallerypair{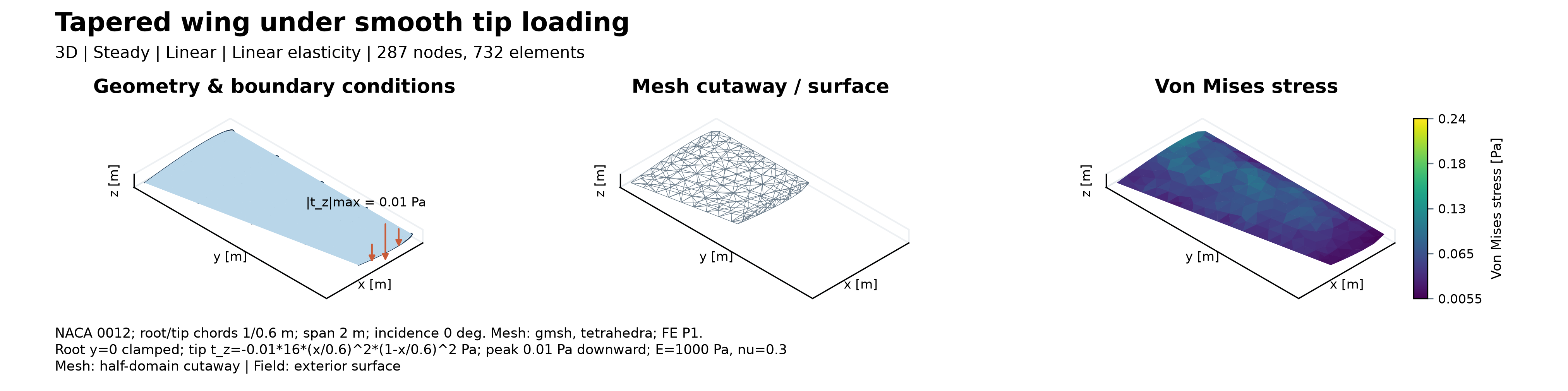}
  {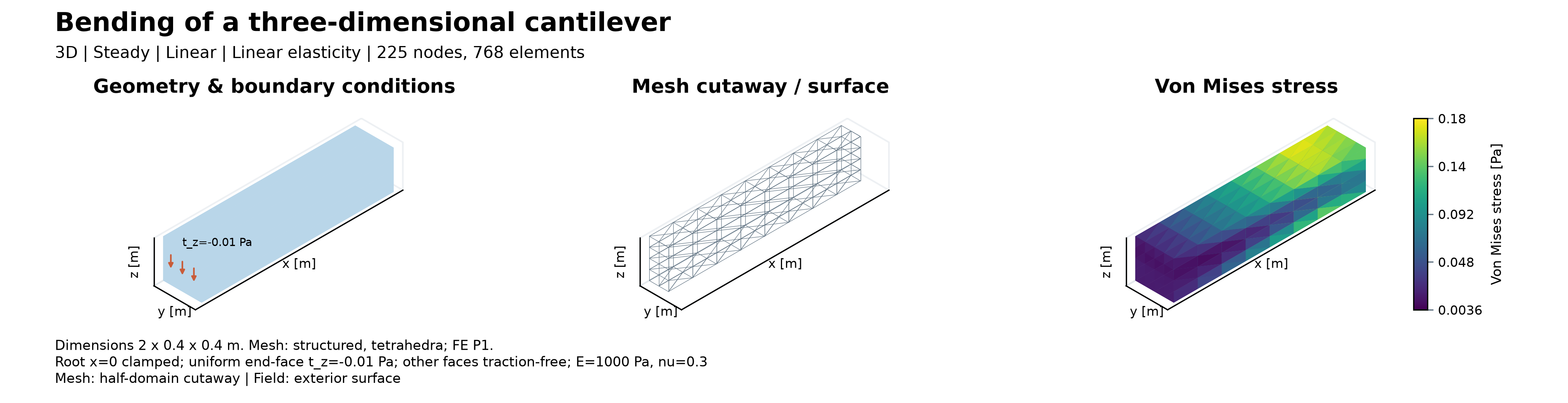}
  {}
  {fig:gallery-3d-elasticity}

\gallerypair{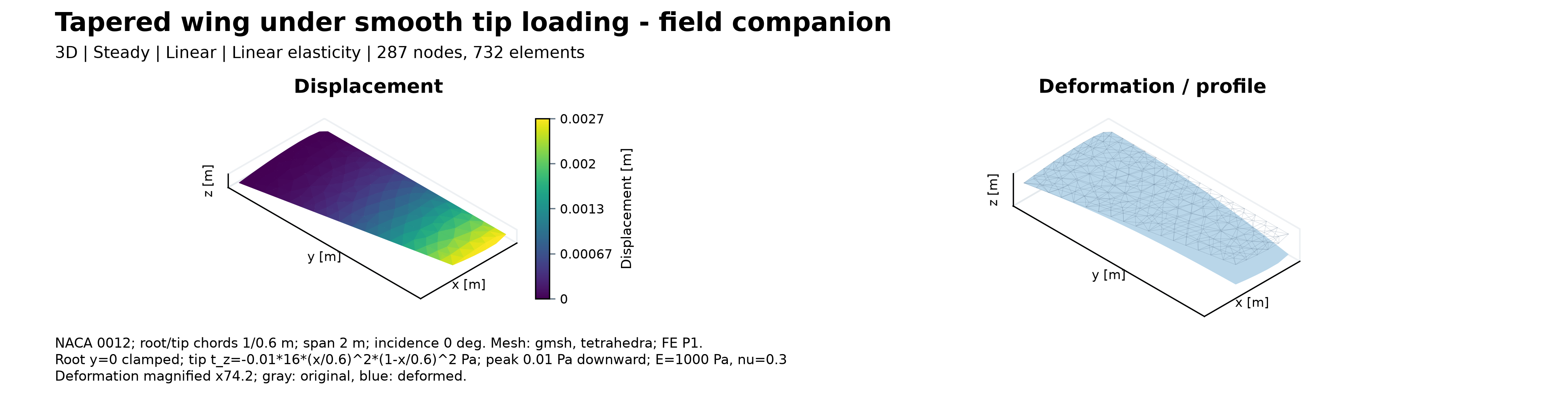}
  {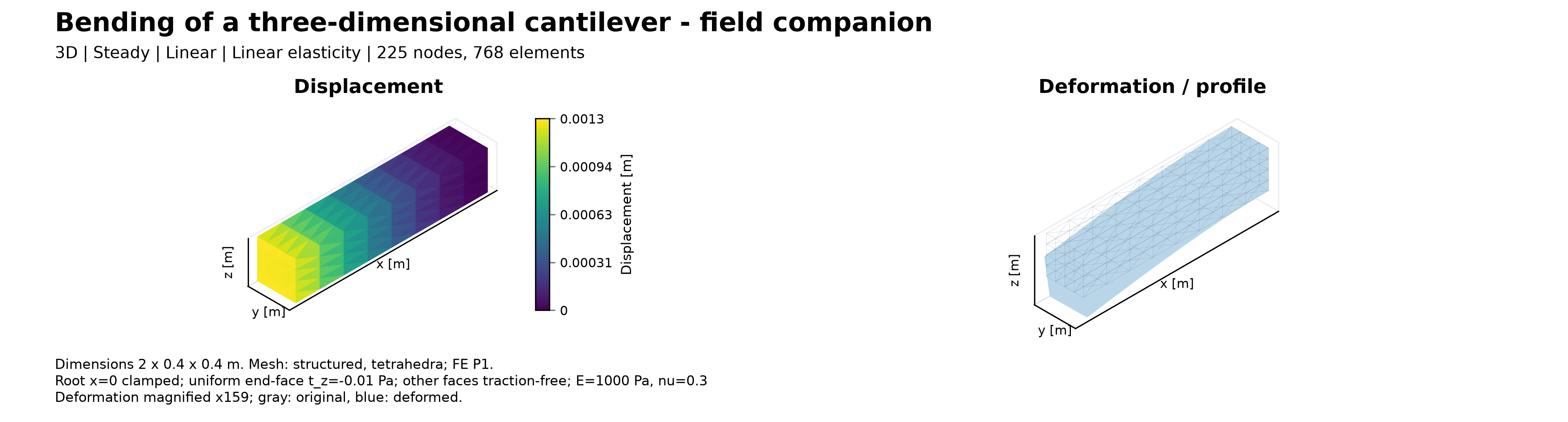}
  {}
  {fig:gallery-companion-3d-elasticity}

\clearpage
\endgroup


\end{document}